\documentclass[10pt,twocolumn,letterpaper]{article}

\usepackage[pagenumbers]{iccv} 

\usepackage{float}

\definecolor{iccvblue}{rgb}{0.21,0.49,0.74}
\usepackage[pagebackref,breaklinks,colorlinks,allcolors=iccvblue]{hyperref}

\def\paperID{*****} 
\def\confName{ICCV}
\def\confYear{2025}

\title{Injecting Image Guidance into Text-Conditioned Diffusion Models at Inference}

\author{%
  \textbf{Agata Żywot} \quad
  \textbf{Iason Skylitsis} \quad
  \textbf{Thijmen Nijdam} \quad
  \textbf{Zoe Tzifa-Kratira} \\
  \textbf{Derck Prinzhorn} \quad
  \textbf{Konrad Szewczyk} \quad
  \textbf{Aritra Bhowmik} \\
  University of Amsterdam, Netherlands \\
}

\begin{document} 
\maketitle
\begin{abstract}
Text-to-image diffusion models like Stable Diffusion generate high-quality images from text, but lack a way to inject visual guidance (e.g. sketches, styles) at inference without retraining. Existing methods either require computationally expensive fine-tuning or rely on style transfer techniques that risk semantic misalignment with textual prompts. We introduce Visual Concept Fusion (VCF), the first method offering dual conditioning on both an image and text prompt at inference time without any concept-specific training. VCF enables visual concept injection into Stable Diffusion by aligning CLIP image features with the text embedding space.
VCF consists of three components: (1) a lightweight aligner that maps image tokens to the text embedding manifold using InfoNCE and cross-attention reconstruction losses, (2) a fusion strategy that preserves both textual and visual semantics, and (3) an optional Prompt-Noise Optimization (PNO) module for test-time refinement. 
Our experiments demonstrate that VCF successfully transfers visual attributes including style, composition, and color palette from reference images while maintaining prompt adherence. 
Quantitative results show a trade-off between text alignment (CLIP score) and visual correspondence (LPIPS), with VCF outperforming baselines in reference fidelity.
\end{abstract}    

\section{Introduction}
\label{sec:intro}

Recent advancements in text-to-image diffusion models, such as Stable Diffusion \cite{stable_diffusion}, have enabled the creation of highly realistic and diverse images conditioned on natural language prompts. The samples generated by these models frequently exhibit rich textures and meaningful semantics, indicating a strong ability to capture information at both low (edges, textures) and high (semantics, composition) levels. However, guiding the models to represent users’ ideas faithfully often requires significant effort dedicated to precise prompt engineering \cite{diff_p_eng}.

\begin{figure}[h!]
\centering
\begin{minipage}[t]{0.25\linewidth}
\centering
\includegraphics[width=\linewidth]{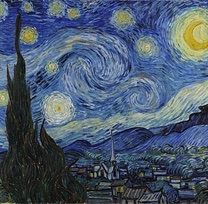} 
\scriptsize Reference Image \
\texttt{+\,a\,photo of\,a\,cat}
\end{minipage}
\hspace{0.03\linewidth}
\begin{minipage}[t]{0.25\linewidth}
\centering
\includegraphics[width=\linewidth]{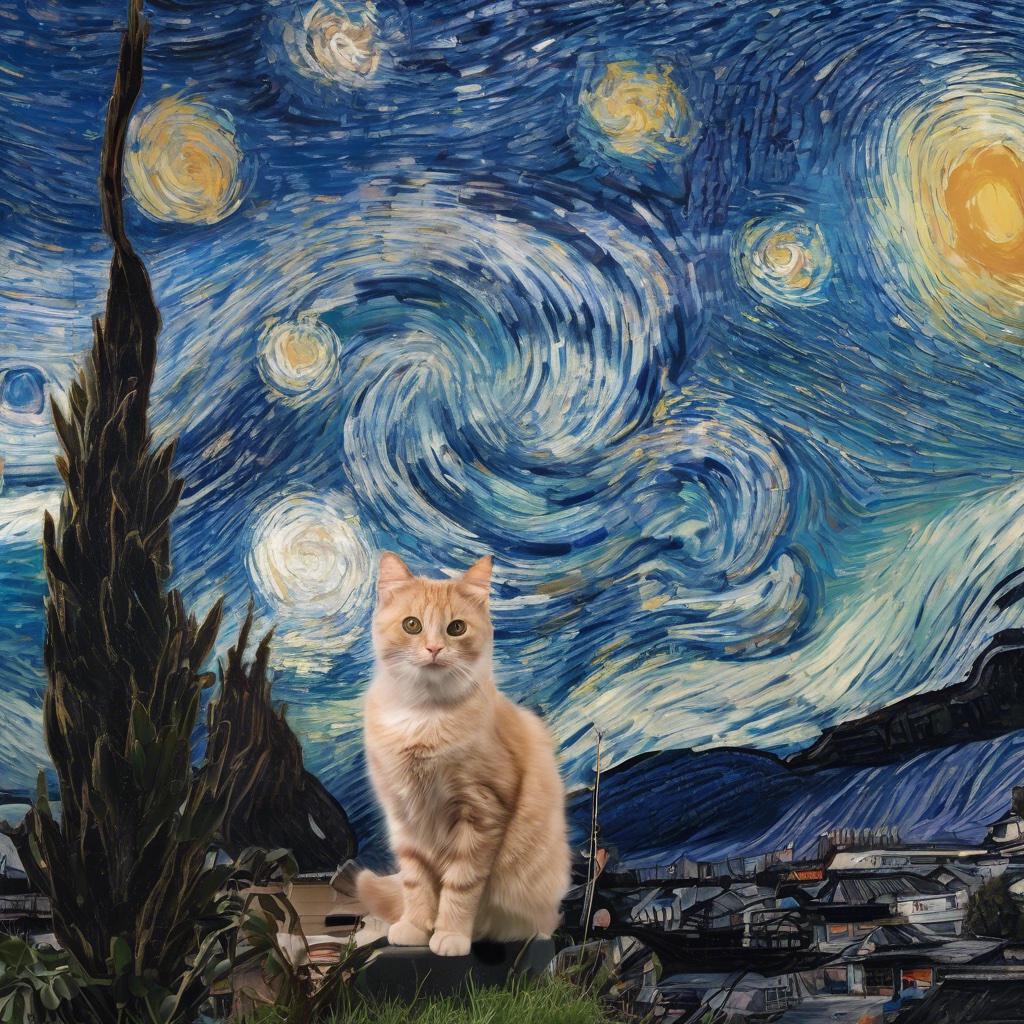} 
\scriptsize Trained Image-Text Model
\end{minipage}
\hspace{0.03\linewidth}
\begin{minipage}[t]{0.25\linewidth}
\centering
\includegraphics[width=\linewidth]{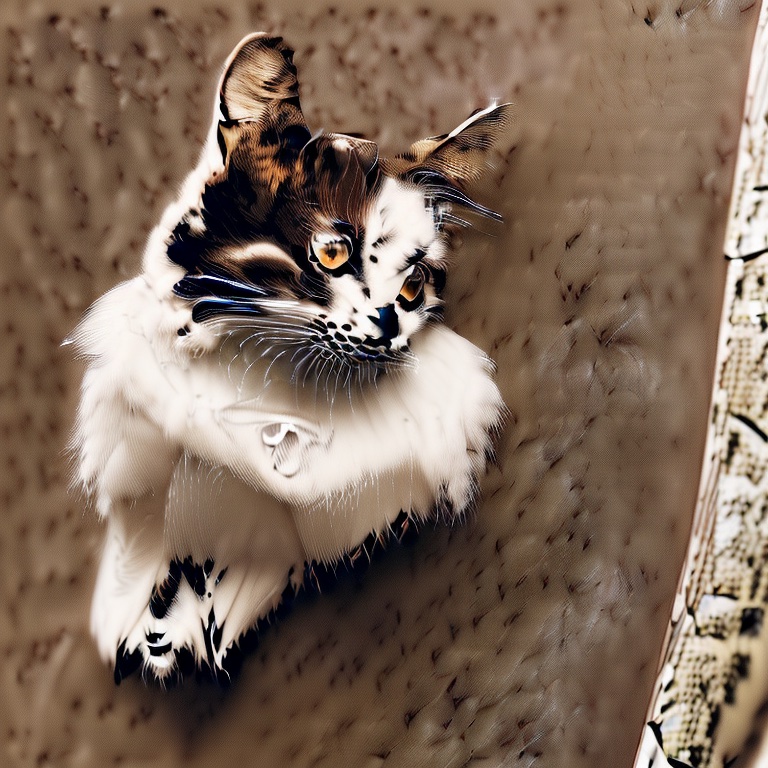} 
\scriptsize Naive Image Fusion (SD)
\end{minipage}
\caption{Illustration of challenges in visual guidance. Left: Reference image and text prompt. Middle: Output from a model trained for joint image-text conditioning \cite{dreambooth}, struggling with full style integration. Right: Output from a standard text-to-image model (SD) with naively blended image features, resulting in a distorted image.}
\label{fig:example_intro}
\end{figure}

To reduce reliance on precise prompting, an emerging solution is to incorporate visual references alongside text, such as sketches, style references, or exemplary images. While this method of conditioning can allow for more accurate and human-friendly guidance of the generation process, existing methods typically require additional fine-tuning \cite{T2I-adapter, ControlNet, dreambooth}. Such fine-tuning can be computationally expensive and necessitates access to additional datasets. Alternative approaches, such as style transfer (e.g., AdaIN \cite{AdaIN}), may risk semantic misalignment with the textual prompt. Furthermore, even models designed for joint conditioning on text and image can be prone to overlooking or inadequately integrating reference image cues. As shown in Figure~\ref{fig:example_intro}, such models may preserve a reference style (\textit{Starry Night}) but apply it inconsistently to the textual subject (e.g., \textit{a photo of a cat}). Effectively integrating such visual cues often demands further costly fine-tuning. Conversely, naively introducing image features into standard text-conditioned pipelines—such as directly adding image tokens through a weighted sum—presents an extrapolation problem, typically yielding poor-quality outputs. This highlights a critical gap: either the model must be retrained extensively for joint conditioning, or visual cues must be integrated in a more sophisticated, non-naive manner. This raises the question: \textit{Can we guide image generation using visual references at inference, without retraining the underlying diffusion model while simultaneously preserving full compatibility with text prompts
?}

\vspace{2mm}

In this paper, we explore the feasibility of injecting visual cues into text-to-image diffusion models at inference time without finetuning the generative model. Our key contribution is the first method that enables simultaneous dual conditioning on both image and text prompts at inference time without requiring any concept-specific training. 
Based on intuition stemming from previous works on adapter models \cite{T2I-adapter}, we posit that diffusion models can be efficiently controlled by adjusting the conditioning signal based on reference image features.
However, naive methods for blending textual and image features yield unsatisfactory results due to misalignment between the distribution of textual and image features. 

Therefore, we propose Visual Concept Fusion (VCF), an efficient approach for enabling style transfer capabilities in text-to-image diffusion models without the need for fine-tuning the diffusion model. Our method can be decomposed into three major components:
\begin{itemize}
    \item \textbf{Modality alignment:} We train a small feature aligner model to alleviate the distribution mismatch between image and textual features. The training requires only a small amount of image–caption data and does not involve the generative diffusion model.
    \item \textbf{Text–image fusion:} We experiment with three distinct fusion methods for blending image and text tokens: (1) Naive fusion, (2) Concatenation, and (3) Cross-attention fusion.
    \item \textbf{Prompt–Noise Optimisation (PNO):} An optional test-time optimisation loop designed to further enhance semantic alignment. It refines both the conditioning signal and the initial noise input to the diffusion process, aiming to maximise the similarity between the generated image and a target visual reference in CLIP’s embedding space.
\end{itemize}

In our work, we demonstrate that the images generated using VCF exhibit similarities in style, composition or colour palette with the reference images, while capturing the contents of the textual prompts. Moreover, we show empirically the impact that the choice of major components of our method (e.g. the aligner, PNO) has on the faithfulness and the quality of the generated samples. We will release our code, aligner weights, and example notebooks to facilitate reproducibility and future research.




\section{Related Work}
\label{sec:related}

\textbf{Deep generative image modeling.} The generation of\,novel images has been a long-studied area of computer vision and deep learning research. Early approaches include Variational Autoencoders (VAEs) \cite{VAE}, which learn an easy-to-sample latent space representation mapped to the image space with a trained decoder, and Generative Adversarial Networks (GANs) \cite{GAN}, which pit a generator against a discriminator during the training phase to produce increasingly realistic samples. While GANs in particular have been proven capable of achieving remarkable image quality \cite{StyleGAN, StyleGANbutBetter}, both of these models suffer from training instability and the risk of mode collapse.

More recently, Denoising Diffusion Probabilistic Models (DDPMs) \cite{DDPM} have emerged as a powerful class of image generative models, demonstrating state-of-the-art performance. At their core are two processes — a fixed forward (diffusion) process that gradually adds Gaussian noise to an input sample over a sequence of
$T$ steps, and a learned reverse (denoising) process that reconstructs a sample from the target data distribution by gradually removing noise, starting from pure Gaussian noise. A significant improvement in making diffusion models more efficient, particularly when working with high-resolution data, was a class of models known as Latent Diffusion Models (LDMs) \cite{stable_diffusion}. Instead of operating in the high-dimensional pixel space, these models perform diffusion and denoising in a lower-dimensional latent space, drastically reducing computational requirements.

Stable Diffusion \cite{stable_diffusion} is a prominent example of an LDM trained for the task of text-to-image generation. It uses CLIP \cite{CLIP} text embeddings as conditioning within the denoising model by injecting them via cross-attention mechanisms. This provided a significant breakthrough in highly realistic image synthesis; however, the conditioning signal is limited to text and introducing other conditioning modalities, such as reference images, poses a difficult challenge due to the features lying in misaligned data distributions.

\textbf{Fine-tuning and adapter-based conditioning.} A\,prominent line of work aiming to solve this problem involves augmenting or fine-tuning pre-trained diffusion models to accept additional image-based conditioning. DreamBooth \cite{dreambooth} enables the personalisation of models by fine-tuning them on a small set of subject images. However, DreamBooth requires computationally expensive fine-tuning of the entire model ($~860M$ parameters) for each new concept and struggles with overfitting when training on limited data. 
Similarly, textual inversion techniques \cite{TextInversion} learn a distribution of new pseudo-words to represent specific visual styles. While more parameter-efficient, textual inversion often struggles to capture complex styles within a few token embeddings and suffers from "concept bleeding," where the learned style overly influences unrelated parts of the prompt. Our method avoids this by aligning feature maps rather than learning discrete tokens, preserving the integrity of the original text prompt.

Other methods like CustomDiffusion \cite{CustomDiffusion} offer more efficient multi-concept customisation by fine-tuning only the key and value projection matrices in the cross-attention layers, requiring only about 75K trainable parameters per concept. However, this still necessitates separate training for each concept and limits scalability.  More recently, StyleDrop \cite{StyleDrop} demonstrated a method for capturing a specific style from a single reference image by fine-tuning a pretrained text-to-image model. While this fine-tuning approach yields impressive results, particularly with large-scale models like Imagen \cite{Imagen}, its effectiveness on publicly available diffusion models like Stable Diffusion can be less pronounced. A significant drawback is that this method requires iterative training of an adapter and fine-tuning of roughly 10M parameters for \textit{each new style}, which is computationally demanding and limits its scalability. Additionally, while effective at style transfer, StyleDrop is still limited to style conditioning only, without supporting simultaneous text and image conditioning.

Another family of approaches includes T2I-Adapter \cite{T2I-adapter} and ControlNet \cite{ControlNet}, which utilise lightweight, trainable modules that inject additional conditioning (e.g., based on visual cues from reference depth maps or sketches) into the frozen backbone of a pre-trained diffusion model. While enabling precise model steering based on various types of visual cues, these methods require training the adapter modules on large datasets of paired image–condition data. Although the core diffusion backbone remains frozen, the training process still demands computationally expensive image sampling at every training step. Our work diverges from these approaches by explicitly avoiding any training that would involve the denoising model directly, instead training a small, modality-aligning network completely separate from the diffusion process.

\textbf{Image prompt adapters.} Recent work has explored more direct approaches to image conditioning. IP-Adapter \cite{ye2023ip} presents a lightweight adapter (22M parameters) that uses decoupled cross-attention to enable image prompt capability in pretrained text-to-image diffusion models. While IP-Adapter successfully enables image prompting, it requires training on large image-caption datasets and primarily focuses on image-only conditioning, with limited exploration of simultaneous image-text conditioning. The decoupled cross-attention strategy separates processing of text and image features but still requires substantial training to align the modalities.

\begin{table}[h!]
\centering
\begin{tabular}{l p{1.5cm} p{3.3cm}}
\toprule
\textbf{Method}
  & \textbf{Trainable\newline Parameters}
  & \textbf{Training\newline Requirement} \\
\midrule
SDEdit              & 0                     & None (inference‑only) \\
CustomDiffusion     & \(\approx75\,\mathrm{k}\)  & Per‑concept fine‑tuning \\
StyleDrop           & \(\approx10\,\mathrm{M}\)  & Per‑style fine‑tuning \\
ControlNet          & \(\approx300\,\mathrm{M}\) & Large paired datasets \\
\textbf{VCF (Ours)} & \(\approx2.4\,\mathrm{M}\) & \textbf{Single model \newline for all images} \\
\bottomrule
\end{tabular}
\caption{Comparison of trainable parameters and training requirements for different generative methods with guidance.}
\label{tab:param_comparison}
\end{table}





\textbf{Training-free guidance.} Training-free diffusion guidance methods aim to steer the generation process at inference time, leveraging the knowledge already present within a pre-trained model. While prompt engineering \cite{TaxPromptEng} can be used to steer generation, it is often complex and time-consuming to achieve results that faithfully reflect the user's intent. As one of the first approaches enabling training-free injection of a visual reference, SDEdit \cite{SDEdit} and its application on models such as Stable Diffusion demonstrated that when a noisy version of a source image is denoised with a diffusion model, the result retains aspects of the source image while adhering to the original conditioning. However, this method is mostly limited to tasks in which the composition of the target image should resemble the reference image and, thus, does not work well for style transfer and similar problems.

Moreover, several techniques focus on manipulating the sampling process of pre-trained diffusion models. 
SkipInject \cite{schaerf2025training} leverages U-Net skip connections in Stable Diffusion for training-free style and content transfer by injecting features from specific skip connections (l=4 and l=5). While the method achieves impressive results for style transfer, it operates primarily on a single image and requires careful timestep scheduling, limiting its applicability to text-guided generation with visual references. 
Plug-and-Play Diffusion Features \cite{PnP_diff} allow for generation control by inverting the reference image using DDIM inversion \cite{DDIM} into the initial noise, which is then denoised using a text-conditioned pre-trained model. Similarly, Add-It \cite{Add-it} enables efficient object insertion into reference images by injecting additional information—provided by an external segmentation model \cite{SAM-2}—into the attention mechanism of the denoising model. However, both of those methods share the same problem as SDEdit in being limited to preserving spatial composition rather than transferring high-level concepts such as art style or semantic content. In contrast, our method is capable of transferring also the high-level concepts such as the art-style or content from the reference image.

\textbf{Limitations of existing approaches and our contribution.} As summarized in \autoref{tab:param_comparison}, existing methods face significant limitations: fine-tuning approaches require expensive per-concept training and substantial computational resources; adapter-based methods, while more efficient, still necessitate training on large paired datasets; and training-free methods are typically limited to spatial composition transfer rather than semantic concept injection. Critically, none of the prior methods simultaneously offer dual conditioning on both an image and text prompt at inference time without any concept-specific training.
Our method avoids these limitations by aligning feature maps rather than learning discrete tokens, preserving the integrity of the original text prompt while enabling flexible visual guidance. VCF represents the first approach to achieve simultaneous dual conditioning on both image and text prompts at inference time without requiring concept-specific training, offering a unique combination of efficiency, flexibility, and expressiveness.

\section{Method}
\label{sec:method}
\begin{figure}[htb]
\centering
\includegraphics[width=0.48\textwidth]{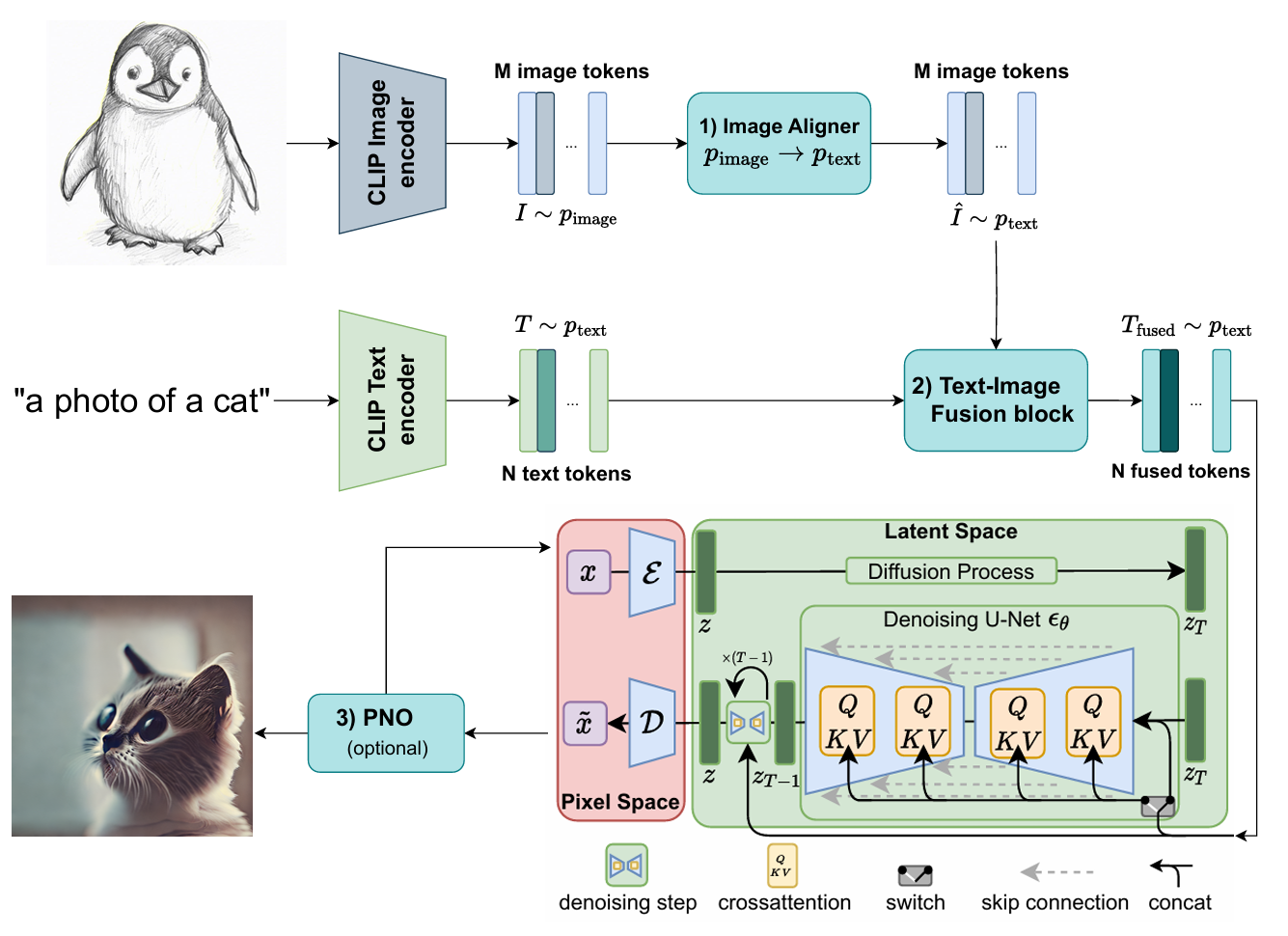}
\caption{VCF pipeline overview. The pipeline integrates image guidance into text-conditioned diffusion models via three key components: (1) the Image Aligner maps image tokens to the text embedding space; (2) the Text–Image Fusion module combines aligned image and text tokens into fused representations; and (3) PNO (optional) refines the fused conditioning and initial noise to enhance visual alignment in the final output.}
\label{fig:pipeline}
\end{figure}

We propose Visual Concept Fusion (VCF), a novel pipeline that integrates image guidance into text-conditioned diffusion models. As shown in \autoref{fig:pipeline}, VCF comprises three key components: (1) an Image Aligner that maps image tokens into the text embedding space for modality alignment; (2) a Text–Image Fusion block that merges aligned image and text features; and (3) an optional Prompt–Noise Optimisation (PNO) module that optimises the generation process at inference.

\subsection{Image-to-Text Alignment}  
Stable Diffusion~v2 (SDv2) conditions its denoising network on \emph{pre-projection} tokens from the CLIP text encoder.  
We denote these tokens by \(T \in \mathbb{R}^{n \times d_{\text{text}}}\), drawn from the distribution \(p_{\text{text}}(T)\).  
Pre-projection tokens are preferred because they preserve richer linguistic detail than the final projected text vector—a single \(1 \times d_{\text{proj}}\) embedding—used in CLIP’s final contrastive loss during training.

To inject visual guidance, we likewise extract pre-projection tokens from the CLIP image encoder, yielding  
\(I \in \mathbb{R}^{m \times d_{\text{image}}}\) with distribution \(p_{\text{image}}(I)\).  
Although the text and image branches are trained jointly, their alignment is enforced only \emph{after} the linear projection layers used for the contrastive loss.  
Consequently, the two pre-projection spaces are not yet aligned, so \(p_{\text{text}} \neq p_{\text{image}}\).  
Injecting \(I\) directly into a text-conditioned SDv2 model therefore creates a modality mismatch, which we quantify via the KL divergence

\[
\Delta_{\mathrm{KL}} 
= \mathrm{KL}\bigl(p_{\theta}(x_0 \mid I)\; \|\; p_{\theta}(x_0 \mid T)\bigr),
\]
where \(x_0\) denotes the final denoised sample.  
A large \(\Delta_{\mathrm{KL}}\) leads to unstable denoising and images that are neither faithful to the reference nor well aligned with the prompt.

\paragraph{Aligner architecture.}
To mitigate this mismatch, we introduce a lightweight aligner \(f_{\phi}\):  
a two-layer MLP with LayerNorm and ReLU activations.  
It is the only component in the VCF pipeline that is trained from scratch; the underlying SD model remains frozen.  
The aligner maps image tokens to an aligned representation  
\(\hat{I} = f_{\phi}(I) \in \mathbb{R}^{m \times d_{\text{text}}}\).

\paragraph{Global alignment objective.}
We encourage the distribution of the aligned tokens \(\hat{I}\) to match that of the text tokens \(T\) via an InfoNCE loss:
\[
\mathcal{L}_{\text{InfoNCE}}
 = -\log
   \frac{\exp\bigl(\cos(\mu_{\hat{I}}, \mu_{T})/\tau\bigr)}
        {\sum_{j} \exp\bigl(\cos(\mu_{\hat{I}}, \mu_{T_j})/\tau\bigr)},
\]
where \(\mu_{\hat{I}}\) and \(\mu_{T}\) are mean embeddings of the image and text tokens, respectively, and \(\tau\) is a learnable temperature.

\paragraph{Local alignment objective.}
To preserve token-level structure, we add a cross-attention reconstruction loss.  
Text tokens are reconstructed from the aligned image tokens:
\[
T' = \operatorname{Attn}(Q = \hat{I},\; K = T,\; V = T),
\quad
\mathcal{L}_{\text{attn}}
 = \|T' - T\|_2^{2}.
\]

\paragraph{Joint training.}
The aligner parameters \(\phi\) are learned with the combined loss:
\[
\mathcal{L}_{\text{align}}
  = \lambda\,\mathcal{L}_{\text{InfoNCE}}
  + \mathcal{L}_{\text{attn}}.
\]
We set \(\lambda = 0.2\).  
Minimising \(\mathcal{L}_{\text{align}}\) realigns the image-derived tokens with the text‐embedding manifold, thereby reducing \(\Delta_{\mathrm{KL}}\) and enabling SD to utilise reference images without sacrificing prompt fidelity.

\subsection{Text--Image Fusion}

After aligning the image tokens \(\hat{I}\) to the text embedding space, we fuse them with the original text tokens \(T\) so that both modalities can guide the diffusion process. We consider three fusion strategies.

\paragraph{Naive (mean) fusion.}
The simplest strategy injects the \emph{same} image-derived signal into every text token.
Given \(\hat{I}\in\mathbb{R}^{m\times d_{\text{text}}}\) and
\(T\in\mathbb{R}^{n\times d_{\text{text}}}\) with \(m\neq n\),
we first average the image tokens,
\[
\hat{I}_{\text{global}}\;=\;\frac{1}{m}\sum_{j=1}^{m}\hat{I}_j
\in\mathbb{R}^{d_{\text{text}}},
\]
and linearly blend this vector with each text token:
\[
T^{\text{fused}}_i
  \;=\;
  (1-\alpha)\,T_i
  + \alpha\,\hat{I}_{\text{global}},
  \qquad i=1,\dots,n,
\]
where \(\alpha\in[0,1]\) controls the influence of the image signal.
Although straightforward, this uniform perturbation often suppresses linguistic nuances in \(T\), leading to noisy and semantically inconsistent outputs; we therefore retain it only as a baseline and refer to it as \emph{naive fusion}.

\paragraph{Concatenation fusion (VCF).}
Our primary method simply concatenates the aligned image tokens to the end of the text sequence,
\([T;\hat{I}]\),
and feeds the combined tokens to Stable Diffusion unchanged.
This preserves the individual semantics of each modality and, empirically, yields the best balance between prompt fidelity and reference adherence.

\paragraph{Cross-attention fusion.}
A third variant allows the text tokens to attend to the image tokens, producing a cross-attended representation that is re-scaled and blended back into the text at every denoising step.
While this approach alleviates some artifacts of naive fusion, it does not match the performance of concatenation fusion in our experiments.
Implementation details and qualitative examples appear in~\autoref{app:cross_attention_fusion}.

\subsection{Prompt-Noise Optimisation}
\label{sec:pno_method}
The final component in our VCF pipeline is Prompt–Noise Optimisation (PNO), an optional, test-time procedure that can be applied to further refine the generation process. Inspired by the original PNO work \cite{pno}, which aimed to mitigate undesirable toxicity, we adapt the framework to enhance visual alignment with a reference image. Specifically, PNO jointly optimises the conditioning tokens \(T_{\text{final}}\) and the initial diffusion noise \(x_T\) to maximise the CLIP similarity between the final generated image and a user-provided visual guide. This process steers the generation towards the reference style or content without compromising the overall image quality. A detailed description of the PNO framework and its mathematical formulation is provided in \textbf{\autoref{app:pno}}.




\section{Results}
\label{sec:results}
We evaluate the effectiveness of our VCF pipeline on the task of guided image generation, where both a reference image and a textual prompt jointly influence the output. We first describe the experimental setup and evaluation metrics, followed by an qualitative and quantitative analysis of the results. All experiments were conducted using our open-source implementation, which will be made publicly available.

\subsection{Experimental Setup}
\label{sec:experimental-setup}

All experiments are conducted using the publicly available Stable Diffusion v2 model\footnote{\url{https://github.com/Stability-AI/stablediffusion}} (768-ema-pruned variant), with DDIM sampling over 50 steps at a resolution of $768 \times 768$ pixels. Our aligner is trained on a 10\% subset of the COCO Captions dataset\footnote{\url{https://huggingface.co/datasets/sentence-transformers/coco-captions}}, consisting of approximately 60,000 randomly selected image–caption pairs. We use an 80/10/10 split for training, validation, and testing, respectively. The training objective combines InfoNCE with a cross-attention reconstruction loss, as described in \autoref{sec:method}. Training the aligner is computationally lightweight and completes in under two hours on a single A100 GPU.

\paragraph{Dataset.}
COCO Captions~\cite{chen2015microsoft} is a large-scale image–caption dataset comprising over 120{,}000 images, each annotated with five human-written descriptions. The captions exhibit a high degree of linguistic diversity, often including compositional and stylistic elements, making the dataset well suited for learning rich text–image alignments. During training, we randomly sample one of the five captions for each image in every epoch to encourage robustness to paraphrasing.

\paragraph{Hyperparameters.}
We adopt standard diffusion settings and introduce additional parameters for the aligner and Prompt–Noise Optimisation (PNO). The InfoNCE loss uses a learnable temperature parameter~\(\tau\), and we balance it with the cross-attention reconstruction loss using a fixed weight of \(\lambda_{\text{align}} = 0.2\). We use fusion strength \(\alpha=0.3\), and apply PNO as an optional test-time refinement. Full hyperparameter details, grouped by component, are provided in \autoref{tab:exp-hparams}.


\begin{table}[H]
\centering
\small 
\caption{Hyperparameters used in all experiments, grouped by component.}
\label{tab:exp-hparams}
\begin{tabular}{@{}p{0.42\linewidth}p{0.48\linewidth}@{}}
\toprule
\multicolumn{2}{l}{\textbf{Diffusion Parameters}} \\
\midrule
Base model & Stable Diffusion v2.1 (768-ema-pruned) \\
Image resolution & $768 \times 768$ \\
Sampling method & DDIM \\
DDIM steps & 50 \\

\midrule
\multicolumn{2}{l}{\textbf{Aligner Parameters}} \\
\midrule
Training dataset & COCO Captions (10\%) \\
Loss function & InfoNCE + Cross-Attention Reconstruction \\
Loss weighting $\lambda_{\text{InfoNCE}}$ & 0.2 \\
InfoNCE temperature $\tau$ & Learnable \\
Training epochs & $10$ \\
\midrule
\multicolumn{2}{l}{\textbf{PNO Parameters}} \\
\midrule
PNO steps & 10–50 \\
Learning rate (PNO) & $1 \times 10^{-2}$ \\
Noise regularisation $\lambda_{\text{reg}}$ & 0.1 \\
Gradient clipping & 1.0 \\
\bottomrule
\end{tabular}
\end{table}


\subsection{Evaluation Metrics}
\noindent
To evaluate the quality of generated images, we consider two main criteria: alignment with the input text prompt, and correspondence to the visual reference. The following metrics are used:

\paragraph{CLIP Score (Text Alignment).}
We quantify semantic alignment between the generated image and the text prompt using the CLIP similarity score. Specifically, we compute the cosine similarity between their embeddings in the CLIP space:
\[
\text{CLIP}(x, t) = \frac{f_{\text{CLIP}}(x) \cdot f_{\text{CLIP}}(t)}{\|f_{\text{CLIP}}(x)\| \, \|f_{\text{CLIP}}(t)\|}
\]
where \( f_{\text{CLIP}}(\cdot) \) denotes the CLIP encoder applied to images and text, respectively. Higher values indicate stronger alignment.

\paragraph{LPIPS (Reference Image Correspondence).}
The Learned Perceptual Image Patch Similarity (LPIPS) \cite{LPIPS} metric measures perceptual similarity between the generated image \( \hat{x} \) and the reference image \( x_{\text{ref}} \). It is defined as:
\[
\text{LPIPS}(x_{\text{ref}}, \hat{x}) = \sum_l w_l \left\| \phi_l(x_{\text{ref}}) - \phi_l(\hat{x}) \right\|_2^2
\]
where \( \phi_l \) are features extracted from layer \( l \) of a pretrained VGG network~\cite{vgg}, and \( w_l \) are learned weights.  
In our setup, we do not learn custom weights and instead fix \( w_l = 1 \) across all layers. Lower LPIPS scores indicate greater perceptual similarity to the reference image.

\subsection{Qualitative Results}
\noindent
We present an overview of qualitative results in \autoref{fig:qualitative-results}, comparing three generation modes: (i) text-only generation using SDv2, (ii) naive fusion, and (iii) our proposed VCF pipeline. All outputs are conditioned on the same prompt—``A photo of a cat''—with only the reference image varying across samples to isolate its influence on the output. Additional examples are provided in \autoref{app:more_qualitative_main}.

As expected, naive fusion does not reliably integrate information from the reference image. While the generated images depict cats, they often appear less realistic and exhibit elevated visual noise. In many instances, these outputs closely resemble those produced by the text-only baseline, indicating that naive fusion fails to meaningfully modulate generation based on the visual reference.

By contrast, generations produced by our VCF method exhibit a much stronger correspondence with the reference image. The transferred features span both high-level semantics (e.g., artistic style, presence of background objects) and low-level visual cues (e.g., colour distribution, shading, depth). For example, when a dog is used as the reference, the output often resembles a hybrid ``cat–dog'' entity that blends shape and colour characteristics from both the text prompt and the image. Moreover, the level of realism in the generated outputs tends to reflect the style of the reference: photorealistic inputs yield realistic generations, while stylised references—such as paintings or prints—result in outputs with matching stylistic attributes.

\newlength{\imgheight}
\setlength{\imgheight}{0.90in}   

\newcommand{\resultrowfour}[4]{%
  \begin{minipage}[t]{0.23\linewidth}\centering
    \includegraphics[width=\linewidth,height=\imgheight,keepaspectratio]{#1}\par
    \scriptsize Reference image
  \end{minipage}\hspace{0.04\linewidth}%
  \begin{minipage}[t]{0.23\linewidth}\centering
    \includegraphics[width=\linewidth,height=\imgheight,keepaspectratio]{#2}\par
    \scriptsize SDv2 (text-only)
  \end{minipage}\hspace{0.01\linewidth}%
  \begin{minipage}[t]{0.23\linewidth}\centering
    \includegraphics[width=\linewidth,height=\imgheight,keepaspectratio]{#3}\par
    \scriptsize Naive fusion
  \end{minipage}\hspace{0.01\linewidth}%
  \begin{minipage}[t]{0.23\linewidth}\centering
    \includegraphics[width=\linewidth,height=\imgheight,keepaspectratio]{#4}\par
    \scriptsize VCF (Ours)
  \end{minipage}\par\vspace{0.3ex} 
}

\begin{figure}[htbp]
  \centering
  \resultrowfour{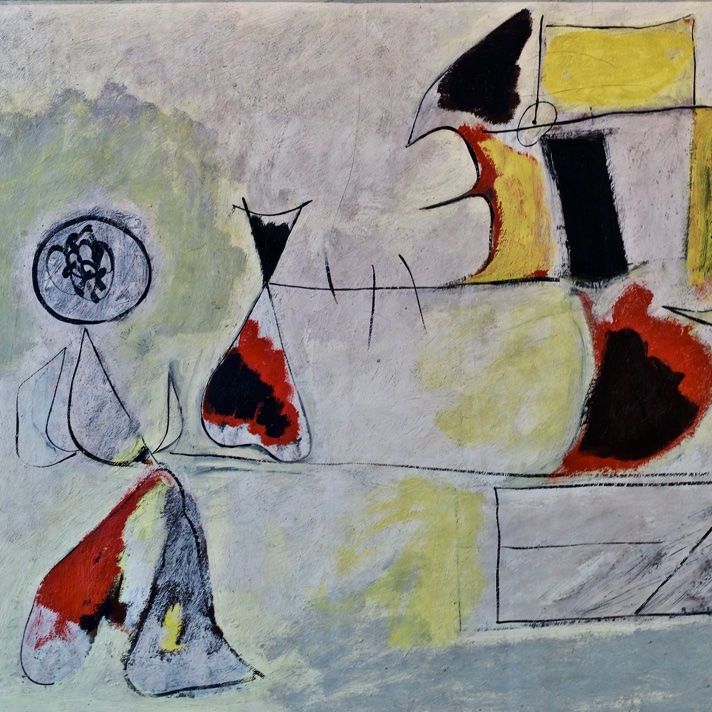}
               {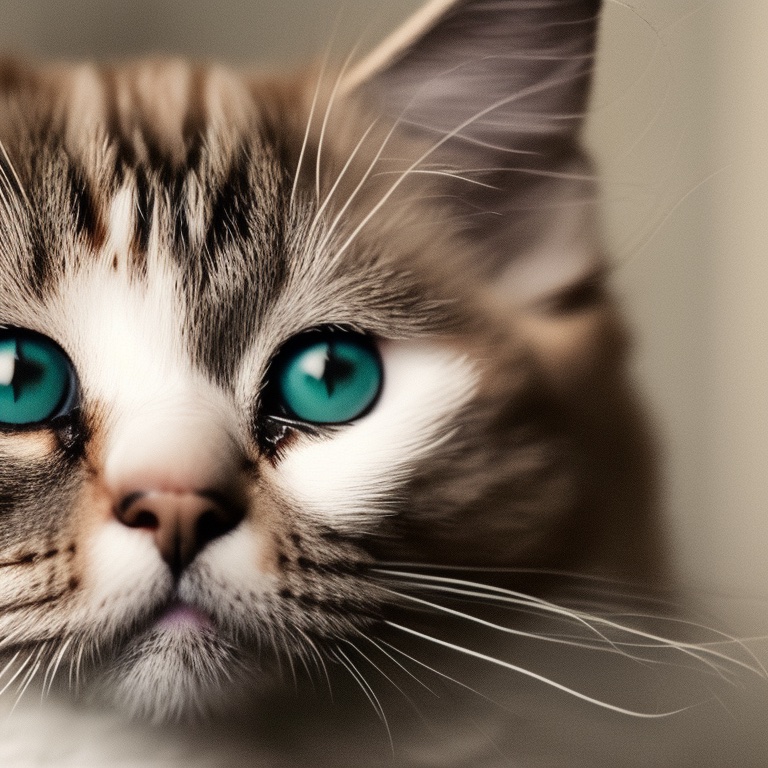}
               {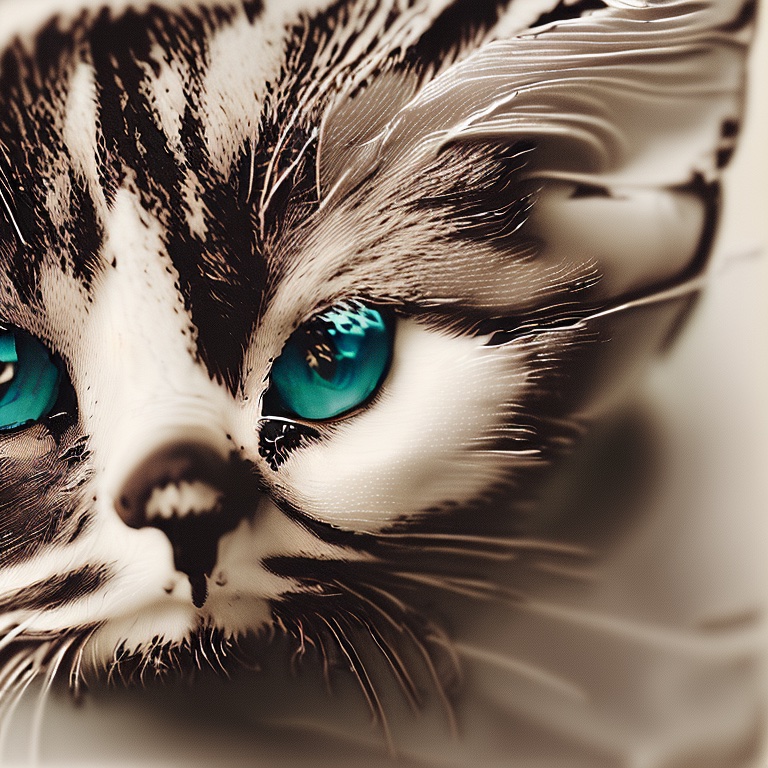}
               {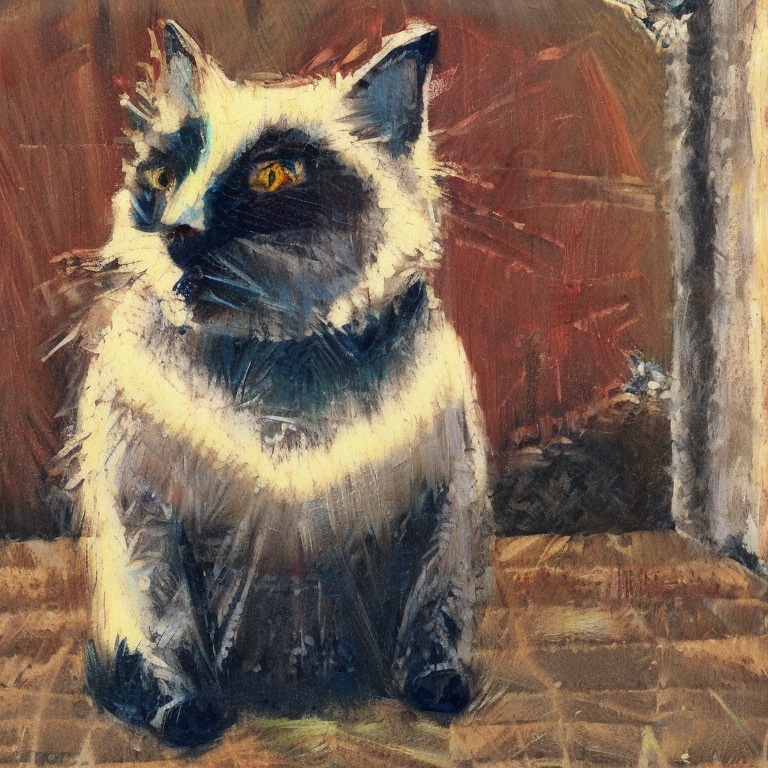}

  \resultrowfour{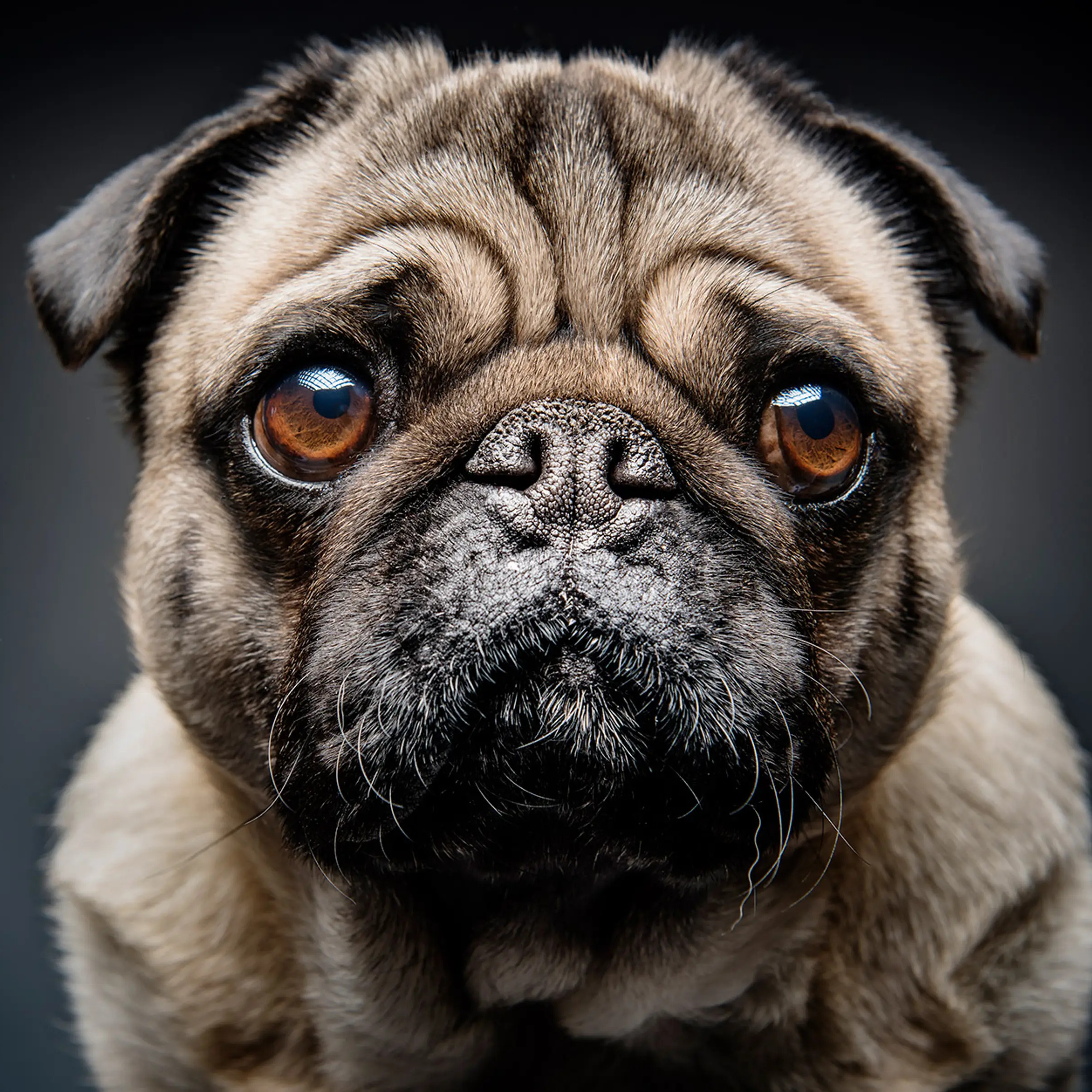}
               {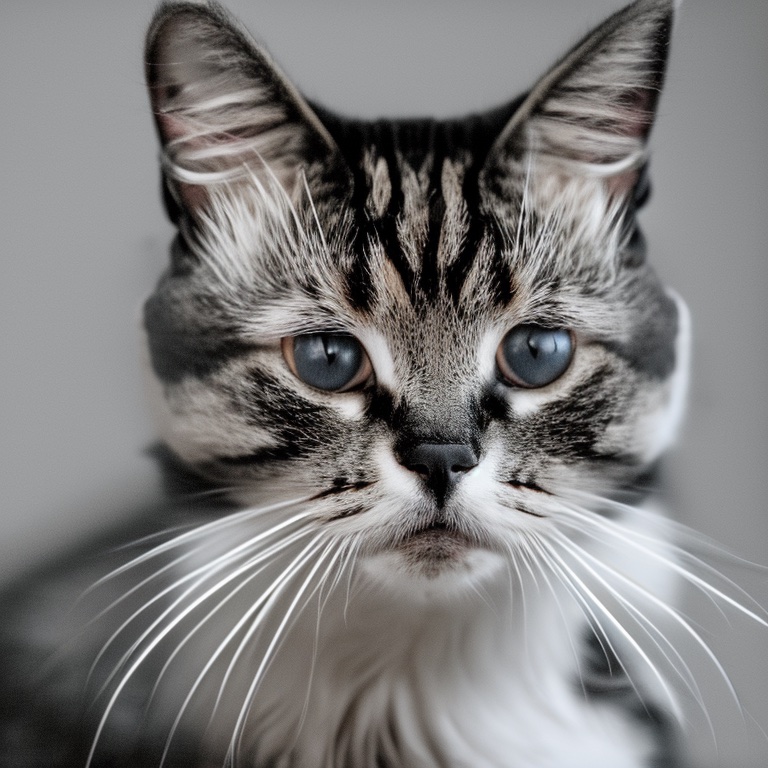}
               {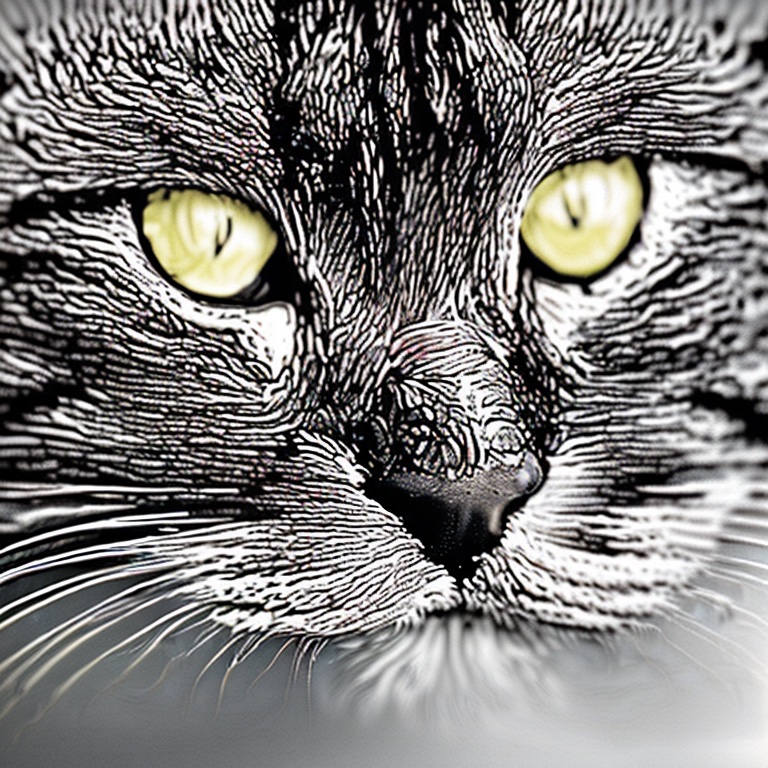}
               {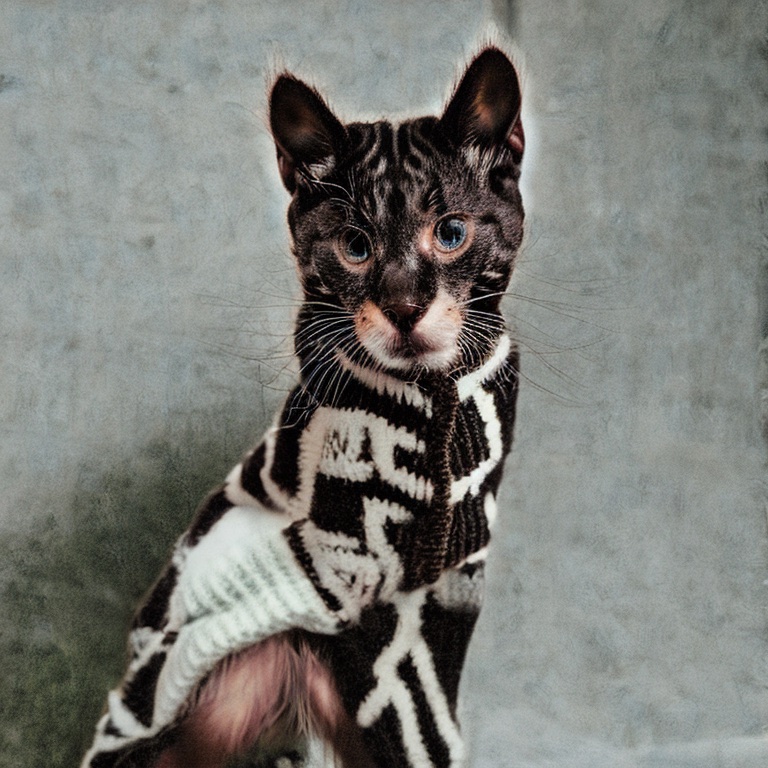}

  \resultrowfour{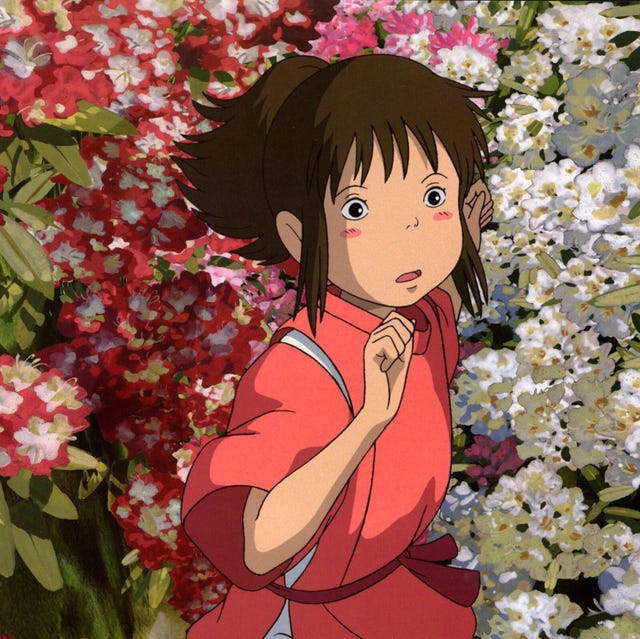}
               {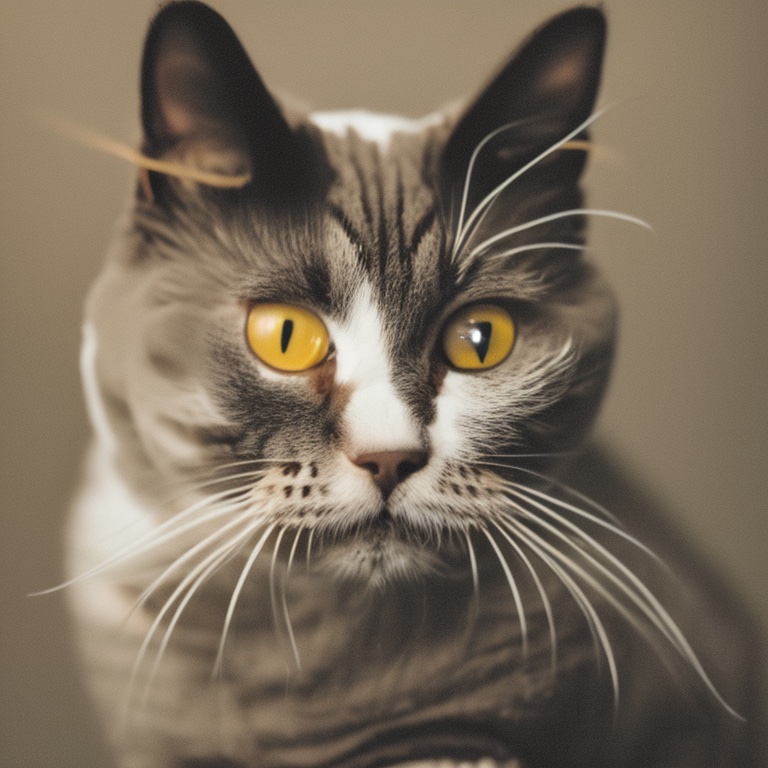}
               {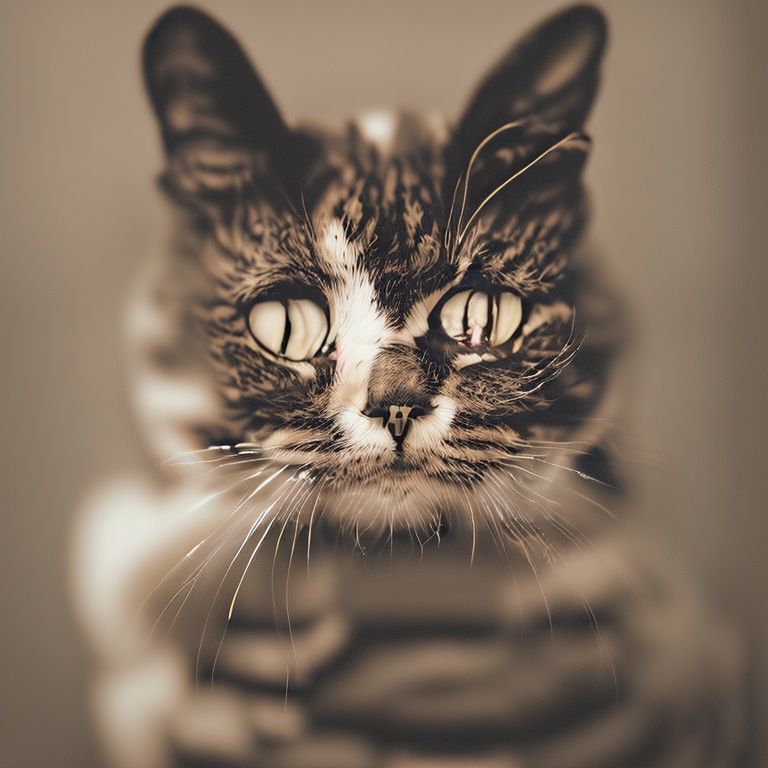}
               {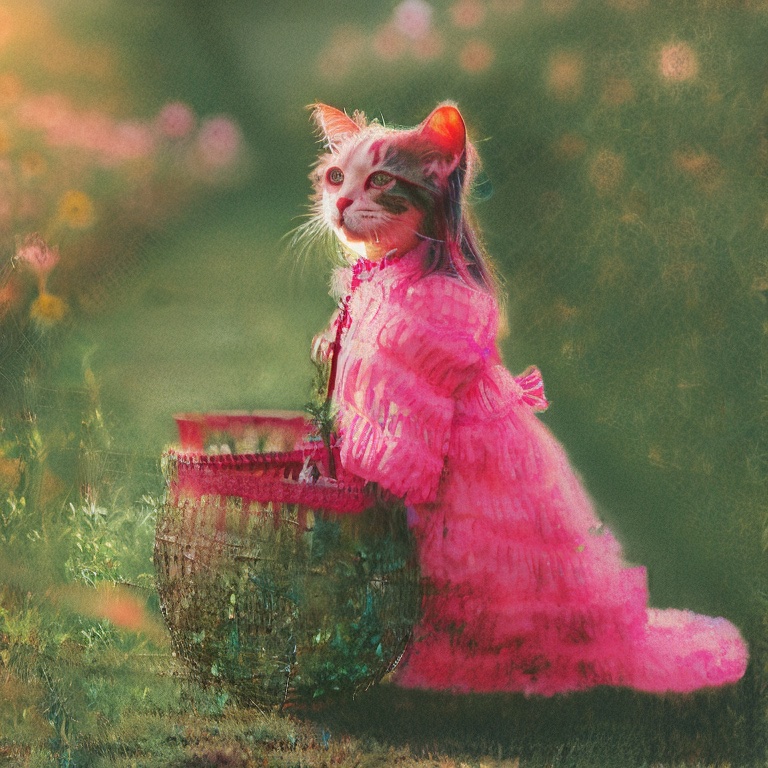}

  \resultrowfour{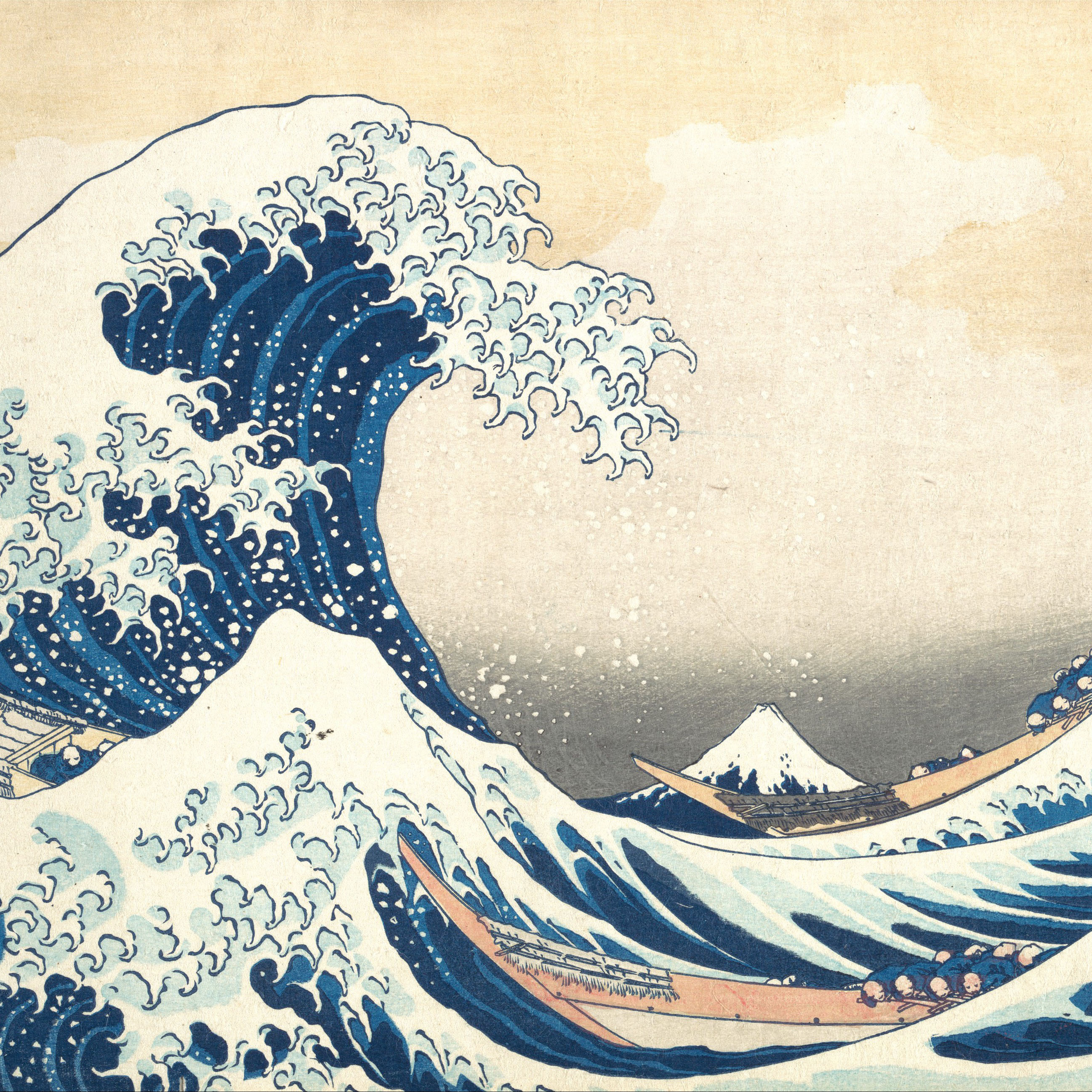}
               {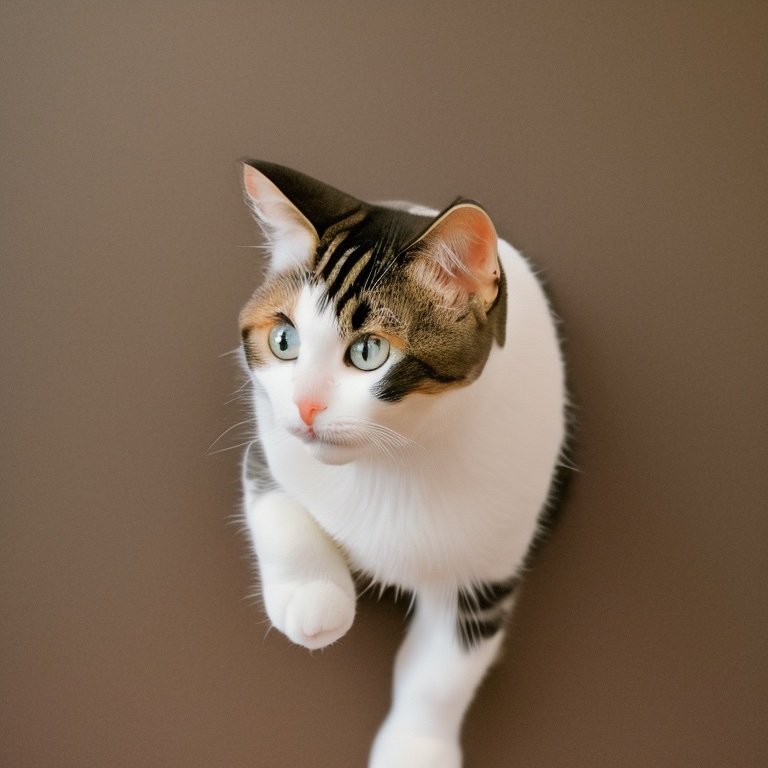}
               {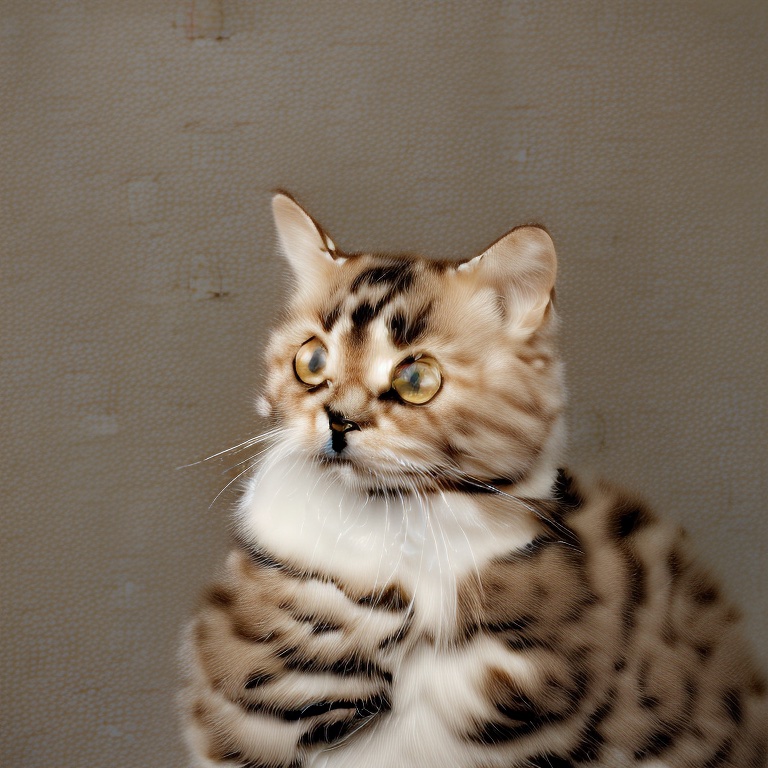}
               {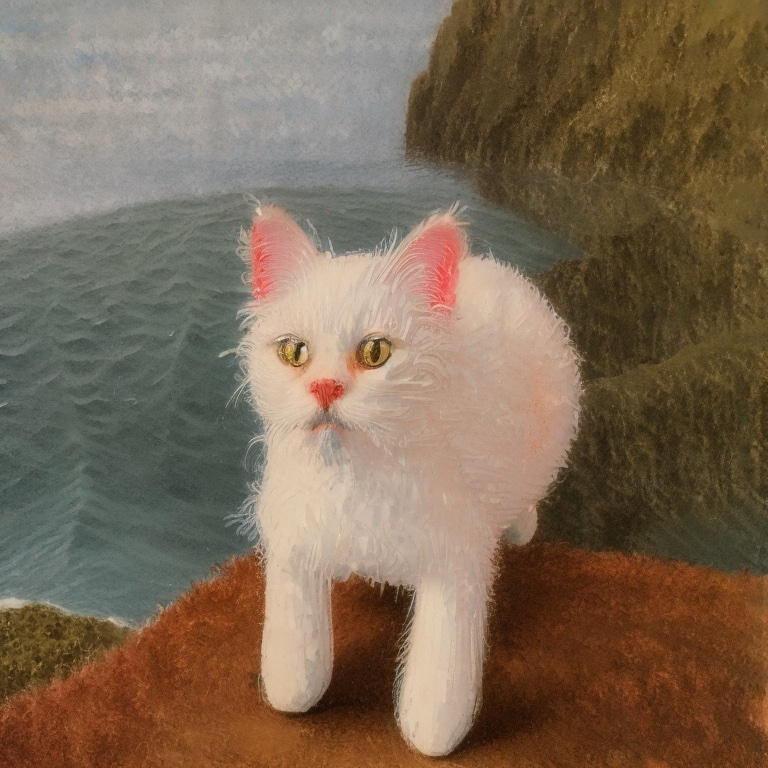}

  \caption{Qualitative comparison of generation methods. Each row shows (left~→~right): the reference image, baseline text-only SDv2 output, naive fusion, and our proposed VCF.}
  \label{fig:qualitative-results}
\end{figure}

\subsection{Quantitative Results}

\noindent
\autoref{tab:main-results} reports performance across two metrics: CLIP score, which measures alignment with the text prompt, and LPIPS, which quantifies perceptual similarity between the generated image and the visual reference.

As expected, the text-only SDv2 model achieves the highest CLIP score, reflecting strong semantic adherence to the prompt. Naive fusion yields a slightly reduced CLIP score, likely due to the noisier and less coherent outputs. Our VCF method shows a further reduction in CLIP score, which is anticipated given its increased reliance on visual guidance. This trade-off is evident in cases such as the ``cat–dog'' hybrid or stylised cat generations shown in \autoref{fig:qualitative-results}, where visual fidelity to the reference image overrides strict prompt literalism.

In contrast, VCF achieves the lowest LPIPS score, indicating the greatest perceptual similarity to the reference images. This result confirms that our method more effectively integrates visual features from the reference. Naive fusion, by comparison, obtains the highest LPIPS score, consistent with its limited capacity to meaningfully condition on the reference and its tendency to revert toward the text-only baseline.

\begin{table}[H]
\centering
\caption{
Quantitative comparison of generation methods. 
CLIP indicates alignment with the text prompt (higher is better), and LPIPS measures perceptual similarity to the reference image (lower is better). Best results per metric are shown in bold.
}
\setlength{\tabcolsep}{4pt}
\renewcommand{\arraystretch}{1.1}
\begin{tabular}{lcc}
\toprule
\textbf{Method} & \textbf{CLIP} $\uparrow$ & \textbf{LPIPS} $\downarrow$ \\
\midrule
SDv2 (text-only) & \textbf{0.29}  & 0.78 \\
Naive fusion              & 0.28           & 0.77 \\
VCF (Ours)     & 0.27           & \textbf{0.76} \\
\bottomrule
\end{tabular}
\label{tab:main-results}
\end{table}

\section{Ablations}

\noindent
To further assess the contributions of individual components in the VCF pipeline, we conduct a series of ablation experiments. All generations are conditioned on the same prompt---``A photo of a cat''---as in previous evaluations. Our main ablation on the aligner loss function is presented below. An additional ablation study on the effect of the optional PNO module can be found in \textbf{\autoref{app:pno_ablation}}.

\subsection{Effect of the Aligner Loss Function}

\noindent
The VCF aligner is trained using a combined objective comprising an InfoNCE loss and a cross-attention reconstruction loss. To understand the role of each term, we retrain the aligner under two ablated configurations: (i) InfoNCE-only, and (ii) cross-attention-only. Qualitative results are shown in \autoref{fig:aligner-ablation}.

With the InfoNCE-only aligner, generated images display little or no visual resemblance to the reference image, although the overall image quality remains comparable to SDv2. This suggests that global distribution alignment alone is insufficient to guide the cross-attention mechanism in Stable Diffusion.

In contrast, using only the cross-attention loss produces outputs that closely follow the reference image, often at the expense of prompt fidelity. For instance, when given a dog as reference, the model generates an image of a dog—even though the prompt specifies a cat. Similarly, a reference depicting a girl in a floral setting yields an output of a girl surrounded by flowers.

Combining both losses achieves a more desirable balance. The InfoNCE term regularises the embedding space globally, while the cross-attention term injects local structure and fine-grained visual cues. This combination enables VCF to produce outputs that respect both the semantics of the prompt and the salient features of the reference.

\newlength{\ablimgheight}
\setlength{\ablimgheight}{0.90in}  

\newcommand{\alignerrow}[4]{%
  \begin{minipage}[t]{0.23\linewidth}\centering
    \includegraphics[width=\linewidth,height=\ablimgheight,keepaspectratio]{#1}\par
    \scriptsize Reference image
  \end{minipage}\hspace{0.04\linewidth}%
  \begin{minipage}[t]{0.23\linewidth}\centering
    \includegraphics[width=\linewidth,height=\ablimgheight,keepaspectratio]{#2}\par
    \scriptsize InfoNCE
  \end{minipage}\hspace{0.01\linewidth}%
  \begin{minipage}[t]{0.23\linewidth}\centering
    \includegraphics[width=\linewidth,height=\ablimgheight,keepaspectratio]{#3}\par
    \scriptsize Cross-Attention
  \end{minipage}\hspace{0.01\linewidth}%
  \begin{minipage}[t]{0.23\linewidth}\centering
      \includegraphics[width=\linewidth,height=\ablimgheight,keepaspectratio]{#4}\par
      \scriptsize 
      \scalebox{0.83}{Both ($\lambda_{\text{InfoNCE}}=0.2$)}
  \end{minipage}\par\vspace{0.3ex}%
}

\begin{figure}[htbp]
  \centering
  \alignerrow{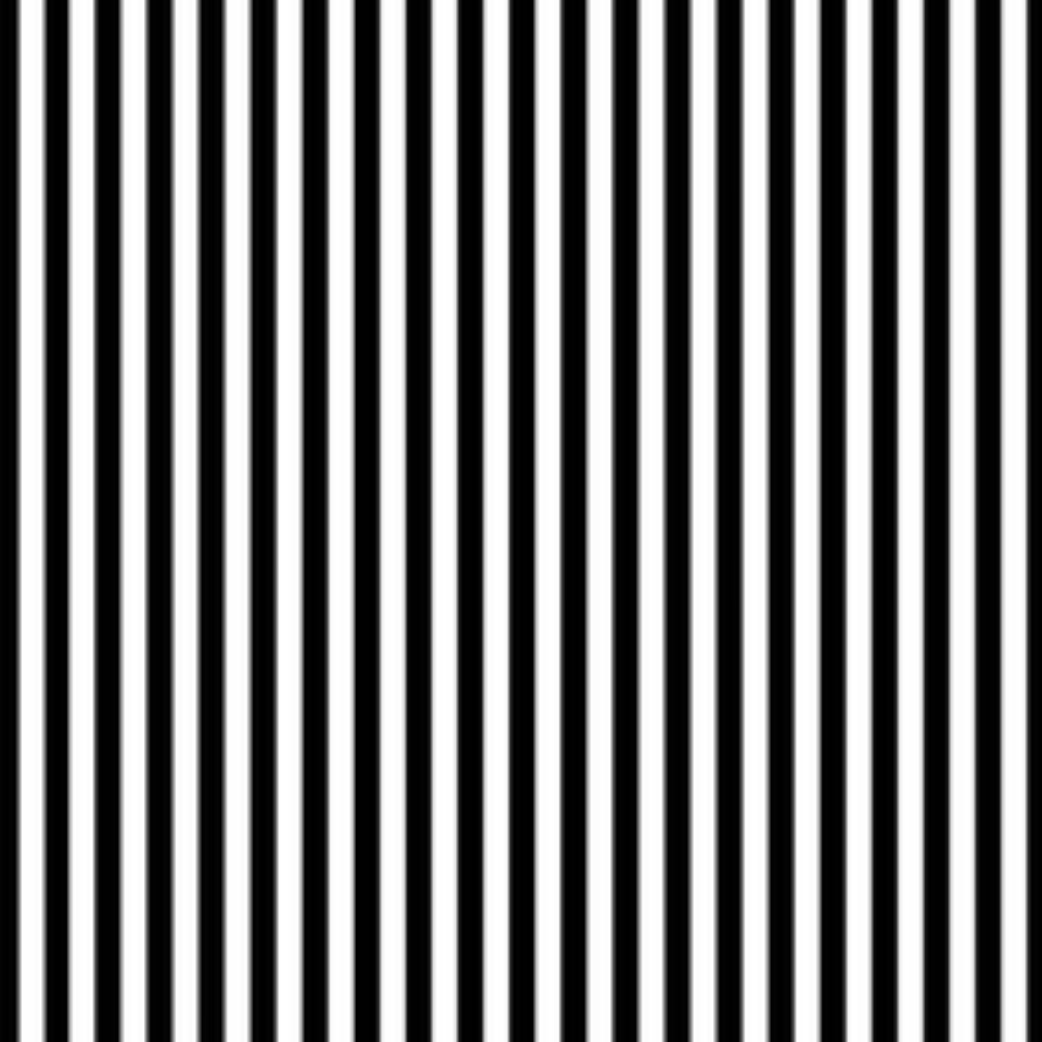}
             {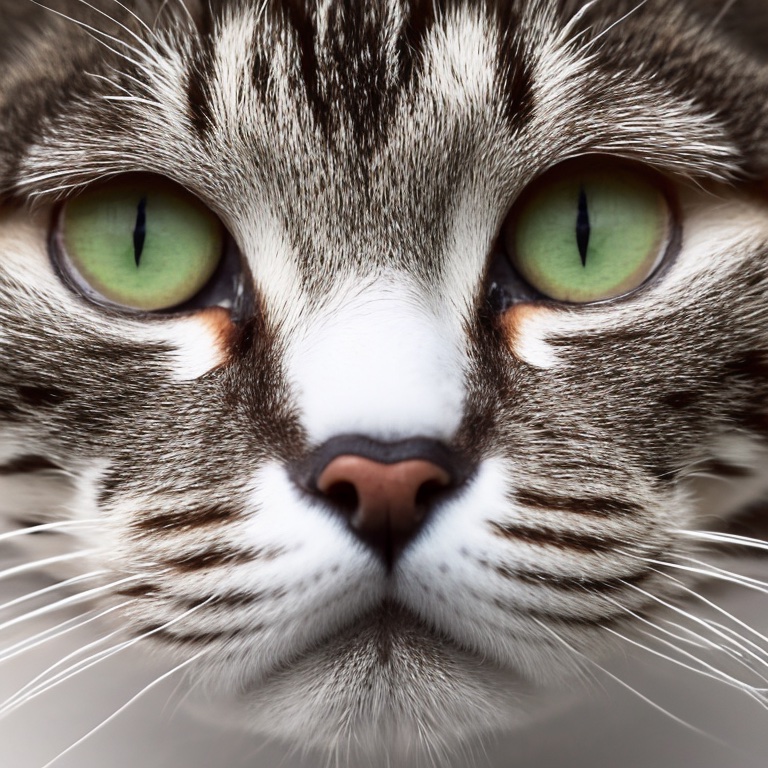}
             {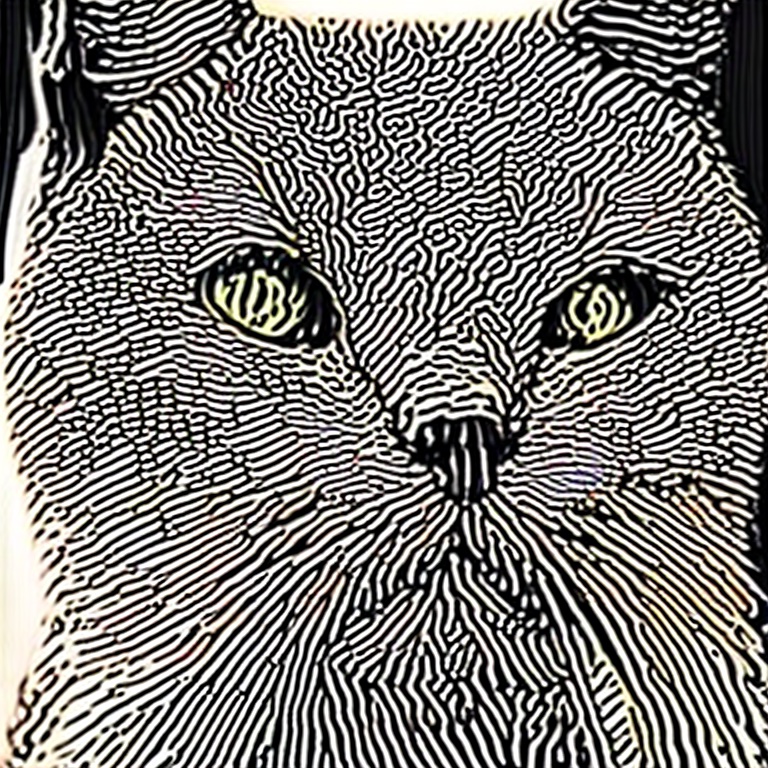}
             {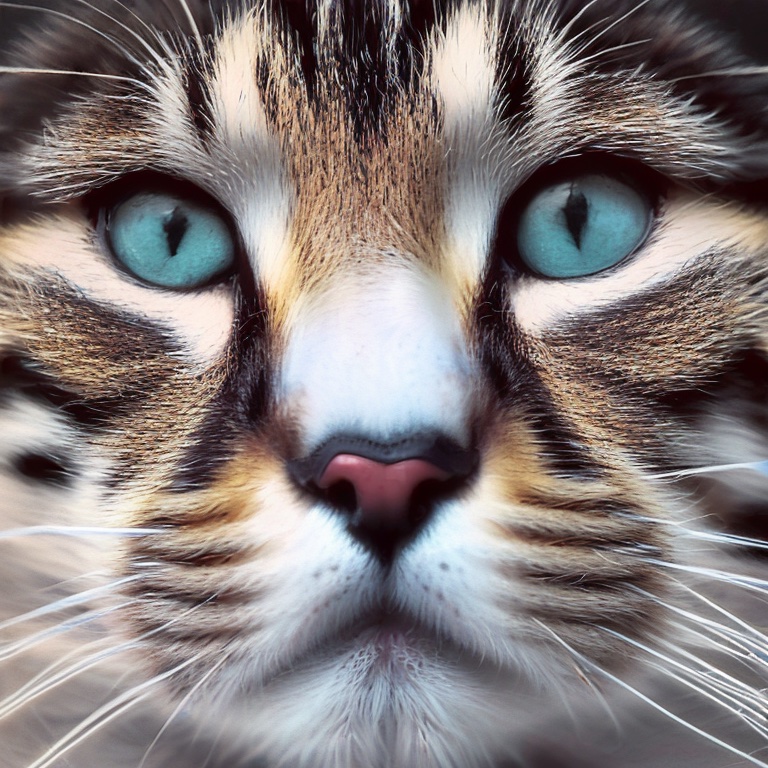}

  \alignerrow{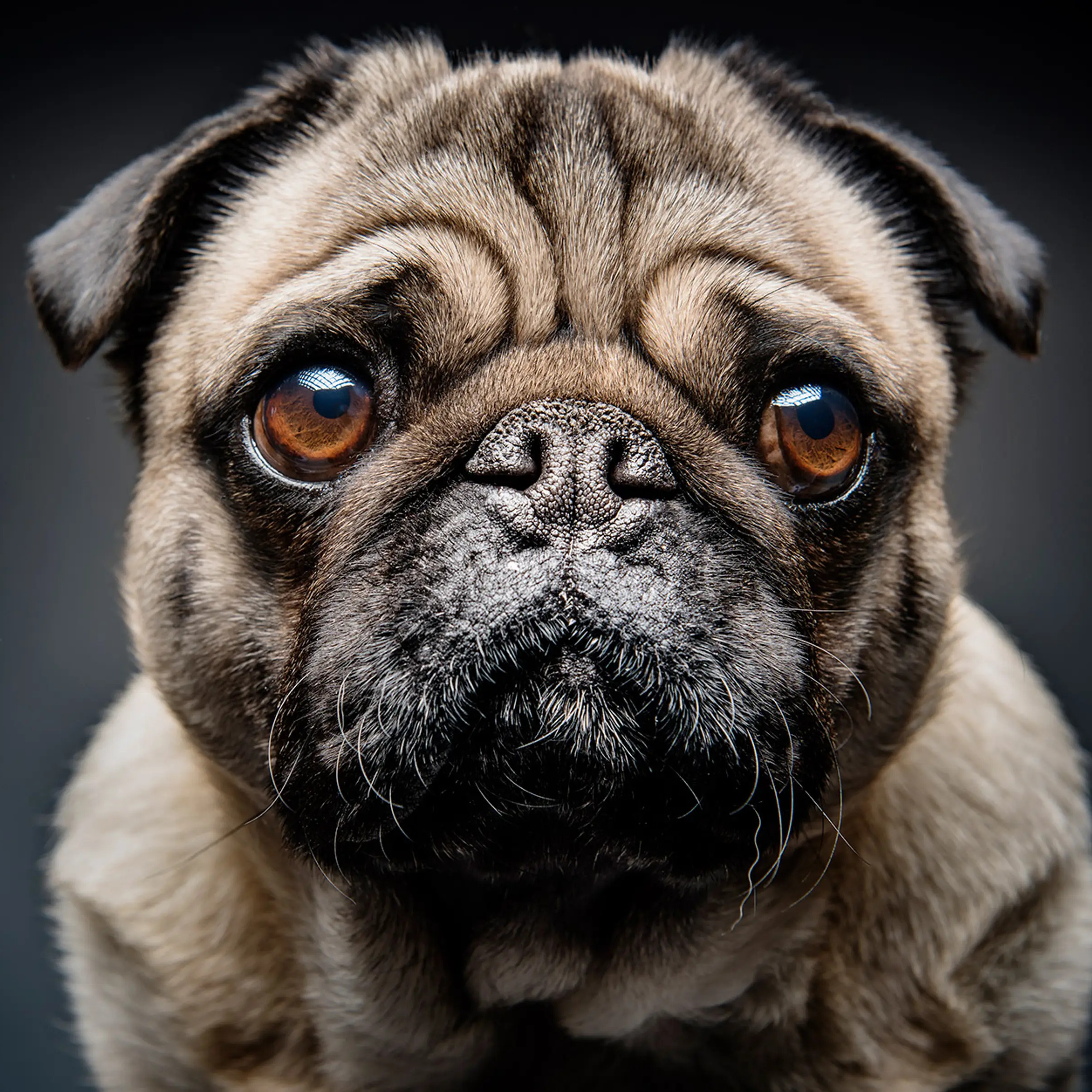}
             {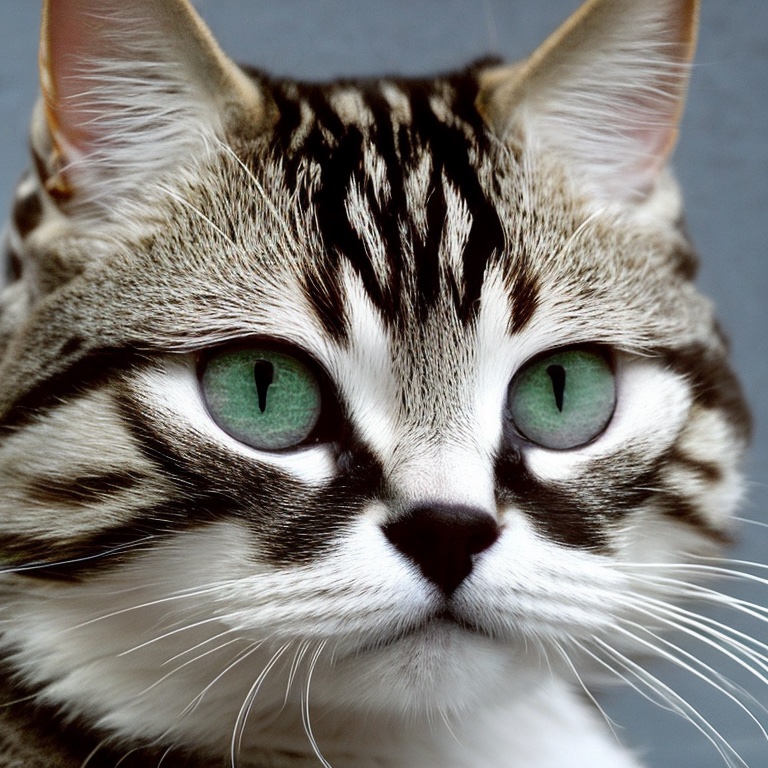}
             {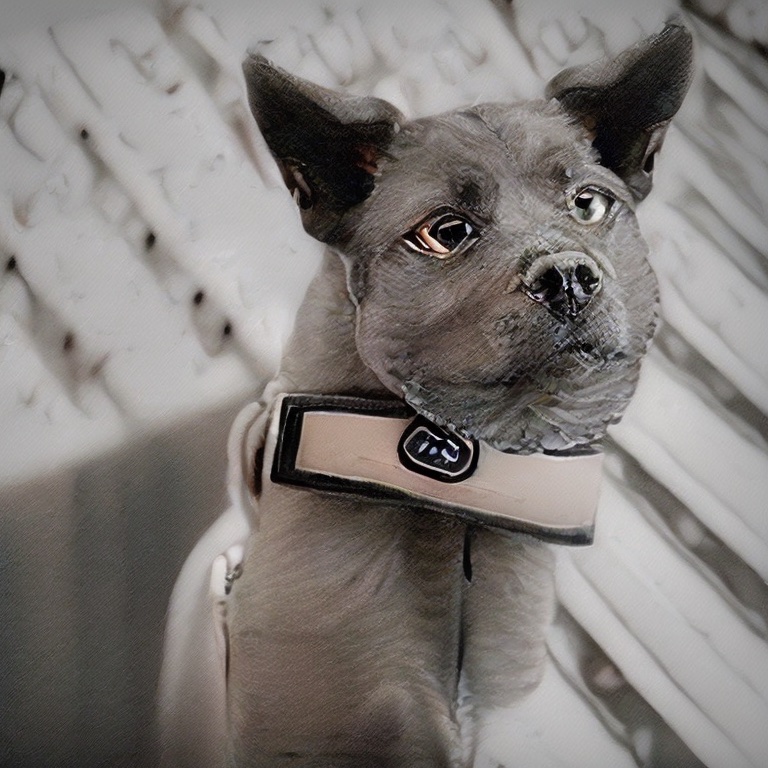}
             {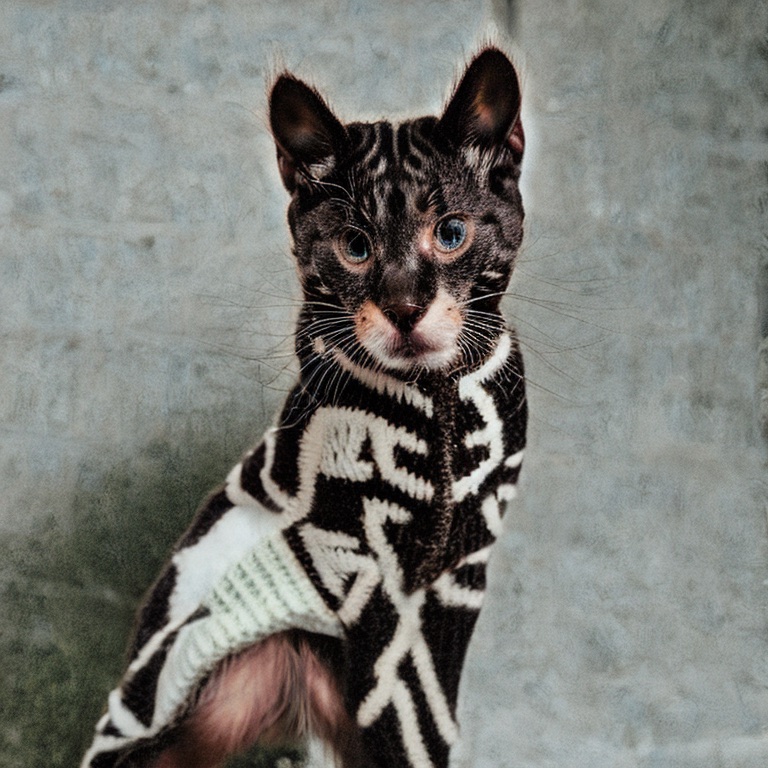}

  \alignerrow{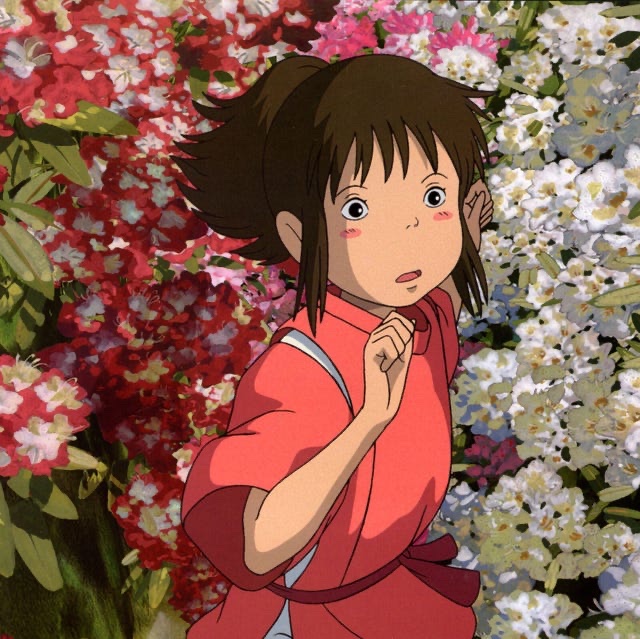}
             {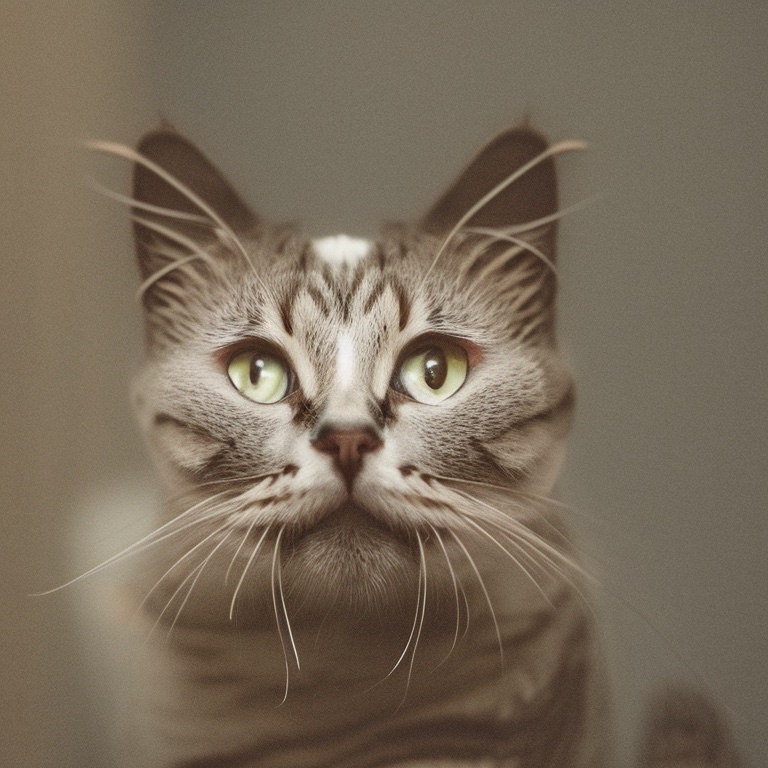}
             {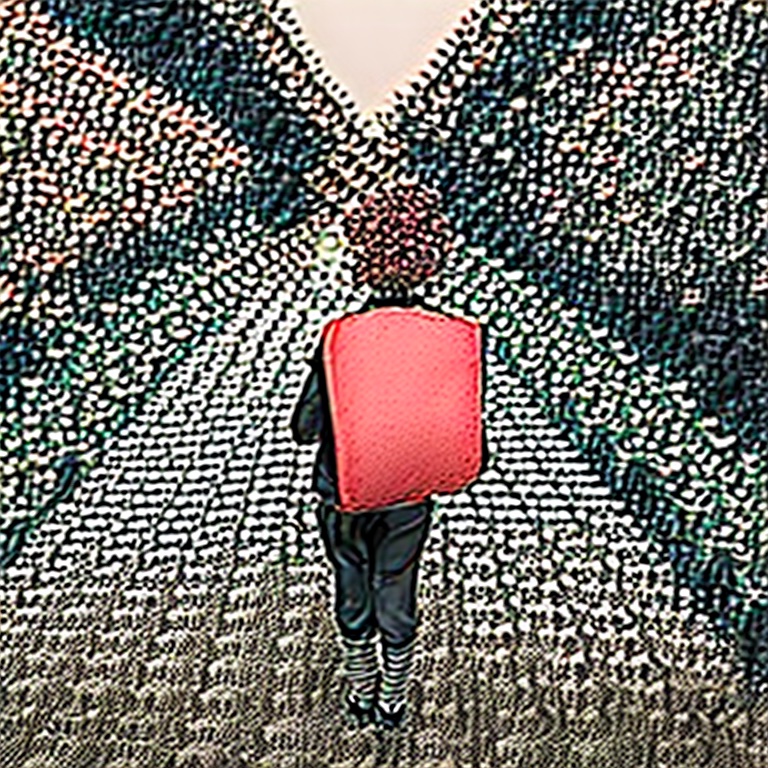}
             {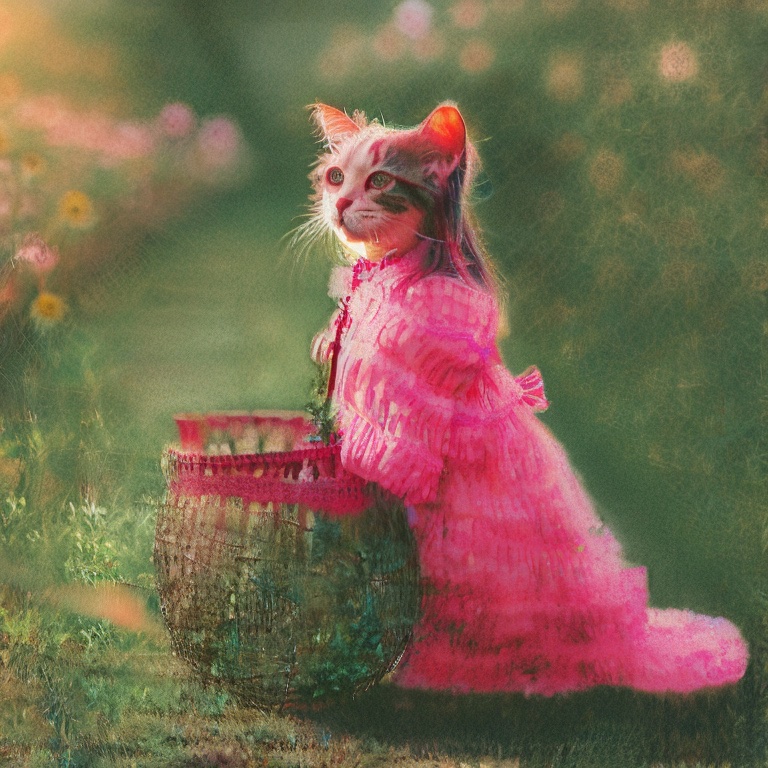}

  \caption{%
  Ablation on aligner loss functions. Each row presents the reference image (left) followed by generations using an InfoNCE loss, a cross-attention reconstruction loss, and their combination with~$\lambda_{\text{InfoNCE}}=0.2$.%
  }
  \label{fig:aligner-ablation}
\end{figure}

\section{Discussion}
\label{sec:discussion}

Our experiments demonstrate that Visual Concept Fusion (VCF) provides an effective framework for integrating reference images into text-conditioned diffusion models. 
Notably, VCF shows particularly significant performance improvements when working with abstract or vague text prompts. As demonstrated in \autoref{fig:appendix3-main-results}, when given ambiguous prompts like "A charming character emerging from the scene," the default Stable Diffusion model struggles to generate coherent and meaningful content. However, introducing reference image conditioning through VCF dramatically improves the quality, detail, and semantic coherence of the generated outputs, transforming vague textual descriptions into visually compelling and meaningful images.
This capability is especially valuable for creative workflows where users may have a clear visual concept in mind but struggle to articulate it precisely through text. The significant boost in generation quality for abstract prompts highlights VCF's unique ability to bridge the semantic gap between imprecise language and precise visual intent. This represents an advancement over text-only generation, where users often resort to complex prompt engineering to achieve desired results.
The results show that naive fusion fails to meaningfully steer generation, whereas VCF consistently produces outputs that reflect both the prompt and the reference. These outputs capture a range of visual attributes, including style, shape, and texture, and adapt to the realism or abstraction of the reference image. The ablations confirm that both components of our method—the aligner and the Prompt–Noise Optimisation—contribute to this improved control.

\paragraph{Limitations} While promising, our work also has several limitations. First, there is no mechanism to control which visual features of the reference image are incorporated into the final output, which may result in unpredictable or overly dominant influence. Second, our ablation studies on aligner training are limited: we only compare loss functions (InfoNCE, cross-attention, or both), using a single dataset (COCO) and one randomly sampled caption per image. Exploring different datasets (e.g., Flickr30K \cite{flicker30k}) or caption strategies may further improve alignment. Lastly, due to time constraints, we were unable to benchmark VCF against existing reference-guided baselines such as SDEdit \cite{SDEdit}, limiting direct comparison with prior work.

Future research could address current limitations by introducing finer control over transferred features, extending training regimes, and evaluating VCF in broader comparative settings. A particularly promising direction would be developing the "steerability" capabilities of VCF by combining its semantic conditioning with spatial control mechanisms inspired by the work of \cite{schaerf2025training} . Such an approach could provide orthogonal control over semantic aspects (through VCF's aligned image features) and structural aspects (through targeted skip connection manipulation), enabling users to independently control content semantics, spatial composition, and temporal scheduling of different visual influences during the diffusion process. This would represent the first comprehensive framework for fine-grained, training-free control over both semantic and spatial dimensions of image generation. Finally, the potential for extending VCF to support multiple reference images simultaneously also presents an interesting avenue for future exploration.

{
    \small
    \bibliographystyle{ieeenat_fullname}
    \bibliography{main}

\begin{thebibliography}{29}
\providecommand{\natexlab}[1]{#1}
\providecommand{\url}[1]{\texttt{#1}}
\expandafter\ifx\csname urlstyle\endcsname\relax
  \providecommand{\doi}[1]{doi: #1}\else
  \providecommand{\doi}{doi: \begingroup \urlstyle{rm}\Url}\fi

\bibitem[Chen et~al.(2015)Chen, Fang, Lin, Vedantam, Gupta, Doll{\'a}r, and Zitnick]{chen2015microsoft}
Xinlei Chen, Hao Fang, Tsung-Yi Lin, Ramakrishna Vedantam, Saurabh Gupta, Piotr Doll{\'a}r, and C~Lawrence Zitnick.
\newblock Microsoft coco captions: Data collection and evaluation server.
\newblock \emph{arXiv preprint arXiv:1504.00325}, 2015.

\bibitem[Gal et~al.(2022)Gal, Alaluf, Atzmon, Patashnik, Bermano, Chechik, and Cohen-Or]{TextInversion}
Rinon Gal, Yuval Alaluf, Yuval Atzmon, Or Patashnik, Amit~H. Bermano, Gal Chechik, and Daniel Cohen-Or.
\newblock An image is worth one word: Personalizing text-to-image generation using textual inversion, 2022.

\bibitem[Ghazanfari et~al.(2023)Ghazanfari, Garg, Krishnamurthy, Khorrami, and Araujo]{LPIPS}
Sara Ghazanfari, Siddharth Garg, Prashanth Krishnamurthy, Farshad Khorrami, and Alexandre Araujo.
\newblock R-lpips: An adversarially robust perceptual similarity metric.
\newblock \emph{arXiv preprint arXiv:2307.15157}, 2023.

\bibitem[Goodfellow et~al.(2014)Goodfellow, Pouget-Abadie, Mirza, Xu, Warde-Farley, Ozair, Courville, and Bengio]{GAN}
Ian~J. Goodfellow, Jean Pouget-Abadie, Mehdi Mirza, Bing Xu, David Warde-Farley, Sherjil Ozair, Aaron Courville, and Yoshua Bengio.
\newblock Generative adversarial networks, 2014.

\bibitem[Ho et~al.(2020)Ho, Jain, and Abbeel]{DDPM}
Jonathan Ho, Ajay Jain, and Pieter Abbeel.
\newblock Denoising diffusion probabilistic models, 2020.

\bibitem[Huang and Belongie(2017)]{AdaIN}
Xun Huang and Serge Belongie.
\newblock Arbitrary style transfer in real-time with adaptive instance normalization, 2017.

\bibitem[Karras et~al.(2019)Karras, Laine, and Aila]{StyleGAN}
Tero Karras, Samuli Laine, and Timo Aila.
\newblock A style-based generator architecture for generative adversarial networks, 2019.

\bibitem[Karras et~al.(2020)Karras, Laine, Aittala, Hellsten, Lehtinen, and Aila]{StyleGANbutBetter}
Tero Karras, Samuli Laine, Miika Aittala, Janne Hellsten, Jaakko Lehtinen, and Timo Aila.
\newblock Analyzing and improving the image quality of stylegan, 2020.

\bibitem[Kingma and Welling(2013)]{VAE}
Diederik~P Kingma and Max Welling.
\newblock Auto-encoding variational bayes, 2013.

\bibitem[Kumari et~al.(2023)Kumari, Zhang, Zhang, Shechtman, and Zhu]{CustomDiffusion}
Nupur Kumari, Bingliang Zhang, Richard Zhang, Eli Shechtman, and Jun-Yan Zhu.
\newblock Multi-concept customization of text-to-image diffusion, 2023.

\bibitem[Liu and Chilton(2023)]{diff_p_eng}
Vivian Liu and Lydia~B. Chilton.
\newblock Design guidelines for prompt engineering text-to-image generative models, 2023.

\bibitem[Meng et~al.(2022)Meng, He, Song, Song, Wu, Zhu, and Ermon]{SDEdit}
Chenlin Meng, Yutong He, Yang Song, Jiaming Song, Jiajun Wu, Jun-Yan Zhu, and Stefano Ermon.
\newblock Sdedit: Guided image synthesis and editing with stochastic differential equations, 2022.

\bibitem[Mou et~al.(2023)Mou, Wang, Xie, Wu, Zhang, Qi, Shan, and Qie]{T2I-adapter}
Chong Mou, Xintao Wang, Liangbin Xie, Yanze Wu, Jian Zhang, Zhongang Qi, Ying Shan, and Xiaohu Qie.
\newblock T2i-adapter: Learning adapters to dig out more controllable ability for text-to-image diffusion models, 2023.

\bibitem[Oppenlaender(2023)]{TaxPromptEng}
Jonas Oppenlaender.
\newblock A taxonomy of prompt modifiers for text-to-image generation.
\newblock \emph{Behaviour \& Information Technology}, 43\penalty0 (15):\penalty0 3763–3776, 2023.

\bibitem[Peng et~al.(2024)Peng, Tang, Liu, Fleming, and Hong]{pno}
Jiangweizhi Peng, Zhiwei Tang, Gaowen Liu, Charles Fleming, and Mingyi Hong.
\newblock Safeguarding text-to-image generation via inference-time prompt-noise optimization, 2024.

\bibitem[Plummer et~al.(2015)Plummer, Wang, Cervantes, Caicedo, Hockenmaier, and Lazebnik]{flicker30k}
Bryan~A Plummer, Liwei Wang, Chris~M Cervantes, Juan~C Caicedo, Julia Hockenmaier, and Svetlana Lazebnik.
\newblock Flickr30k entities: Collecting region-to-phrase correspondences for richer image-to-sentence models.
\newblock In \emph{Proceedings of the IEEE international conference on computer vision}, pages 2641--2649, 2015.

\bibitem[Radford et~al.(2021)Radford, Kim, Hallacy, Ramesh, Goh, Agarwal, Sastry, Askell, Mishkin, Clark, Krueger, and Sutskever]{CLIP}
Alec Radford, Jong~Wook Kim, Chris Hallacy, Aditya Ramesh, Gabriel Goh, Sandhini Agarwal, Girish Sastry, Amanda Askell, Pamela Mishkin, Jack Clark, Gretchen Krueger, and Ilya Sutskever.
\newblock Learning transferable visual models from natural language supervision, 2021.

\bibitem[Ravi et~al.(2024)Ravi, Gabeur, Hu, Hu, Ryali, Ma, Khedr, Rädle, Rolland, Gustafson, Mintun, Pan, Alwala, Carion, Wu, Girshick, Dollár, and Feichtenhofer]{SAM-2}
Nikhila Ravi, Valentin Gabeur, Yuan-Ting Hu, Ronghang Hu, Chaitanya Ryali, Tengyu Ma, Haitham Khedr, Roman Rädle, Chloe Rolland, Laura Gustafson, Eric Mintun, Junting Pan, Kalyan~Vasudev Alwala, Nicolas Carion, Chao-Yuan Wu, Ross Girshick, Piotr Dollár, and Christoph Feichtenhofer.
\newblock Sam 2: Segment anything in images and videos, 2024.

\bibitem[Rombach et~al.(2021)Rombach, Blattmann, Lorenz, Esser, and Ommer]{stable_diffusion}
Robin Rombach, Andreas Blattmann, Dominik Lorenz, Patrick Esser, and Björn Ommer.
\newblock High-resolution image synthesis with latent diffusion models, 2021.

\bibitem[Ruiz et~al.(2023)Ruiz, Li, Jampani, Pritch, Rubinstein, and Aberman]{dreambooth}
Nataniel Ruiz, Yuanzhen Li, Varun Jampani, Yael Pritch, Michael Rubinstein, and Kfir Aberman.
\newblock Dreambooth: Fine tuning text-to-image diffusion models for subject-driven generation, 2023.

\bibitem[Saharia et~al.(2022)Saharia, Chan, Saxena, Li, Whang, Denton, Ghasemipour, Ayan, Mahdavi, Lopes, Salimans, Ho, Fleet, and Norouzi]{Imagen}
Chitwan Saharia, William Chan, Saurabh Saxena, Lala Li, Jay Whang, Emily Denton, Seyed Kamyar~Seyed Ghasemipour, Burcu~Karagol Ayan, S.~Sara Mahdavi, Rapha~Gontijo Lopes, Tim Salimans, Jonathan Ho, David~J Fleet, and Mohammad Norouzi.
\newblock Photorealistic text-to-image diffusion models with deep language understanding, 2022.

\bibitem[Schaerf et~al.(2025)Schaerf, Alfarano, Silvestri, and Impett]{schaerf2025training}
Ludovica Schaerf, Andrea Alfarano, Fabrizio Silvestri, and Leonardo Impett.
\newblock Training-free style and content transfer by leveraging u-net skip connections in stable diffusion 2.
\newblock \emph{arXiv preprint arXiv:2501.14524}, 2025.

\bibitem[Simonyan and Zisserman(2015)]{vgg}
Karen Simonyan and Andrew Zisserman.
\newblock Very deep convolutional networks for large-scale image recognition, 2015.

\bibitem[Sohn et~al.(2023)Sohn, Ruiz, Lee, Chin, Blok, Chang, Barber, Jiang, Entis, Li, Hao, Essa, Rubinstein, and Krishnan]{StyleDrop}
Kihyuk Sohn, Nataniel Ruiz, Kimin Lee, Daniel~Castro Chin, Irina Blok, Huiwen Chang, Jarred Barber, Lu Jiang, Glenn Entis, Yuanzhen Li, Yuan Hao, Irfan Essa, Michael Rubinstein, and Dilip Krishnan.
\newblock Styledrop: Text-to-image generation in any style, 2023.

\bibitem[Song et~al.(2022)Song, Meng, and Ermon]{DDIM}
Jiaming Song, Chenlin Meng, and Stefano Ermon.
\newblock Denoising diffusion implicit models, 2022.

\bibitem[Tewel et~al.(2024)Tewel, Gal, Samuel, Atzmon, Wolf, and Chechik]{Add-it}
Yoad Tewel, Rinon Gal, Dvir Samuel, Yuval Atzmon, Lior Wolf, and Gal Chechik.
\newblock Add-it: Training-free object insertion in images with pretrained diffusion models, 2024.

\bibitem[Tumanyan et~al.(2022)Tumanyan, Geyer, Bagon, and Dekel]{PnP_diff}
Narek Tumanyan, Michal Geyer, Shai Bagon, and Tali Dekel.
\newblock Plug-and-play diffusion features for text-driven image-to-image translation, 2022.

\bibitem[Ye et~al.(2023)Ye, Zhang, Liu, Han, and Yang]{ye2023ip}
Hu Ye, Jun Zhang, Sibo Liu, Xiao Han, and Wei Yang.
\newblock Ip-adapter: Text compatible image prompt adapter for text-to-image diffusion models.
\newblock \emph{arXiv preprint arXiv:2308.06721}, 2023.

\bibitem[Zhang et~al.(2023)Zhang, Rao, and Agrawala]{ControlNet}
Lvmin Zhang, Anyi Rao, and Maneesh Agrawala.
\newblock Adding conditional control to text-to-image diffusion models.
\newblock \emph{2023 IEEE/CVF International Conference on Computer Vision (ICCV)}, pages 3813--3824, 2023.

\end{thebibliography}
}
\appendix
\begin{center}\LARGE\textsc{Appendix}\end{center}
\label{sec:appendix}

\section{Prompt-Noise Optimisation (PNO) Details}
\label{app:pno}

As introduced in \autoref{sec:pno_method}, Prompt–Noise Optimisation (PNO) is an optional, test-time procedure that refines both the conditioning tokens \(T_{\text{final}}\) and the initial diffusion noise \(x_T\) before commencing the reverse sampling process. Our approach is inspired by the original Prompt–Noise Optimisation work \cite{pno}, which aimed to mitigate undesirable toxicity in generated images by optimising prompt embeddings and the noise trajectory. We adapt this framework by modifying the optimisation objective: instead of minimising a toxicity score, our PNO seeks to maximise the similarity (i.e., minimise the negative similarity) between the eventually generated image \(x_0\) and a user-provided visual reference \(x_{\text{guide}}\).

This optimisation leverages the CLIP model's embedding space. Specifically, we jointly optimise \(T_{\text{final}}\) and \(x_T\) to improve the cosine similarity between the CLIP embedding of the generated image \(x_0\) and that of the reference image \(x_{\text{guide}}\), while applying a regularisation term to the initial noise \(x_T\):
\begin{align*}
\mathcal{L}_{\text{PNO}} =\ & \lambda_{\text{reg}} \mathcal{L}_{\text{reg}}(x_T) \\
& - \cos\left(\text{CLIP}(f(x_T, T_{\text{final}})),\; \text{CLIP}(x_{\text{guide}})\right)
\end{align*}

Here, \(f(x_T, T_{\text{final}})\) represents the diffusion model's generation process that yields \(x_0\) from \(x_T\) and \(T_{\text{final}}\). \(x_{\text{guide}}\) is the reference image capturing the desired visual concept. \(\mathcal{L}_{\text{reg}}\) is a noise trajectory regularisation loss designed to prevent degenerate solutions and maintain a plausible noise structure for \(x_T\). \(\lambda_{\text{reg}}\) is a weighting factor balancing the two terms (set to 0.1 by default). In our experiments, this optimisation is performed for a small number of gradient steps (10–50) prior to initiating the full DDIM sampling.

It is important to note the role of \(x_T\) optimisation in this context. While the original PNO paper \cite{pno} discussed the concept of optimising the entire noise trajectory, which controls detailed image features, we operate within the framework of a deterministic DDIM sampler. For DDIM, the entire denoising trajectory—and consequently the final generated image \(x_0\)—is uniquely determined by the initial noise \(x_T\) (given fixed conditioning \(T_{\text{final}}\) and model parameters). Therefore, in our PNO implementation, optimising the \textit{noise trajectory} effectively translates to optimising this initial noise \(x_T\). Modifying \(x_T\) allows us to steer the generation towards better alignment with \(x_{\text{guide}}\) without compromising image quality, as significant deviations from a standard Gaussian distribution for intermediate noise steps could degrade generation quality.

\section{Ablation: Effect of PNO}
\label{app:pno_ablation}

\noindent
We investigate the impact of the optional PNO module on the final image generation. PNO is applied at test time to refine both the conditioning signal and the initial noise, aiming to enhance alignment with the reference image.

\begin{figure}[htb]
\centering
\newcommand{\resultspnointernalnew}[3]{
\begin{minipage}[t]{0.3\linewidth}
  \centering
  \includegraphics[width=\linewidth]{#1}
  \scriptsize Reference image
\end{minipage}
\hspace{1.5mm}
\begin{minipage}[t]{0.3\linewidth}
  \centering
  \includegraphics[width=\linewidth]{#2}
  \scriptsize SDv2 (text-only)
\end{minipage}
\hspace{1.5mm}
\begin{minipage}[t]{0.3\linewidth}
  \centering
  \includegraphics[width=\linewidth]{#3}
  \scriptsize SDv2 with PNO
\end{minipage}
\vspace{2mm}
}

\resultspnointernalnew
  {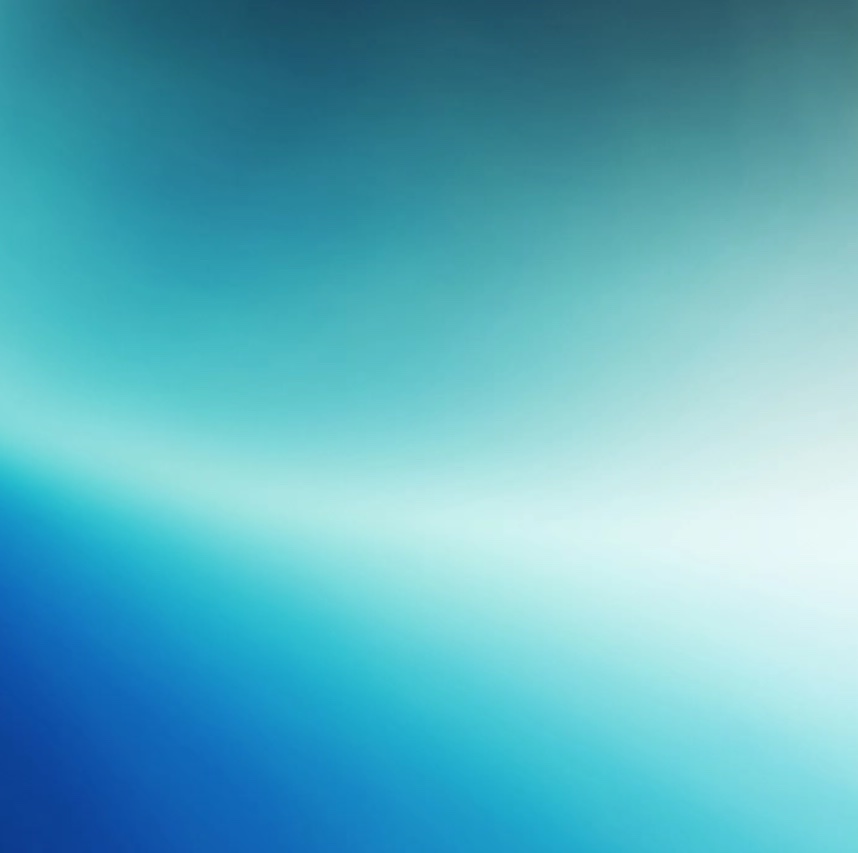}
  {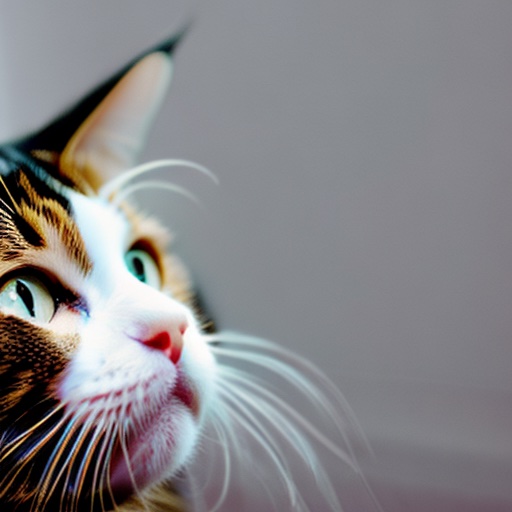}
  {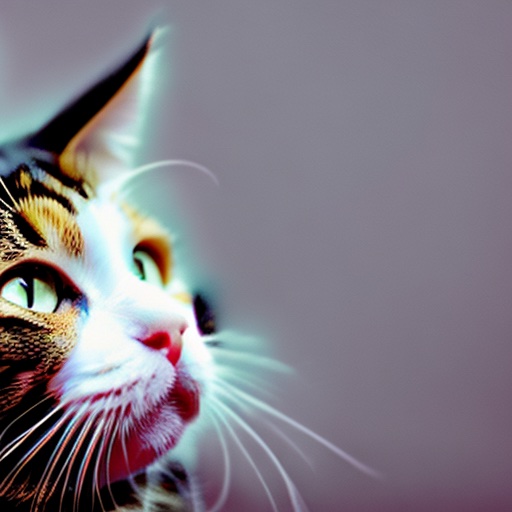}
\resultspnointernalnew
  {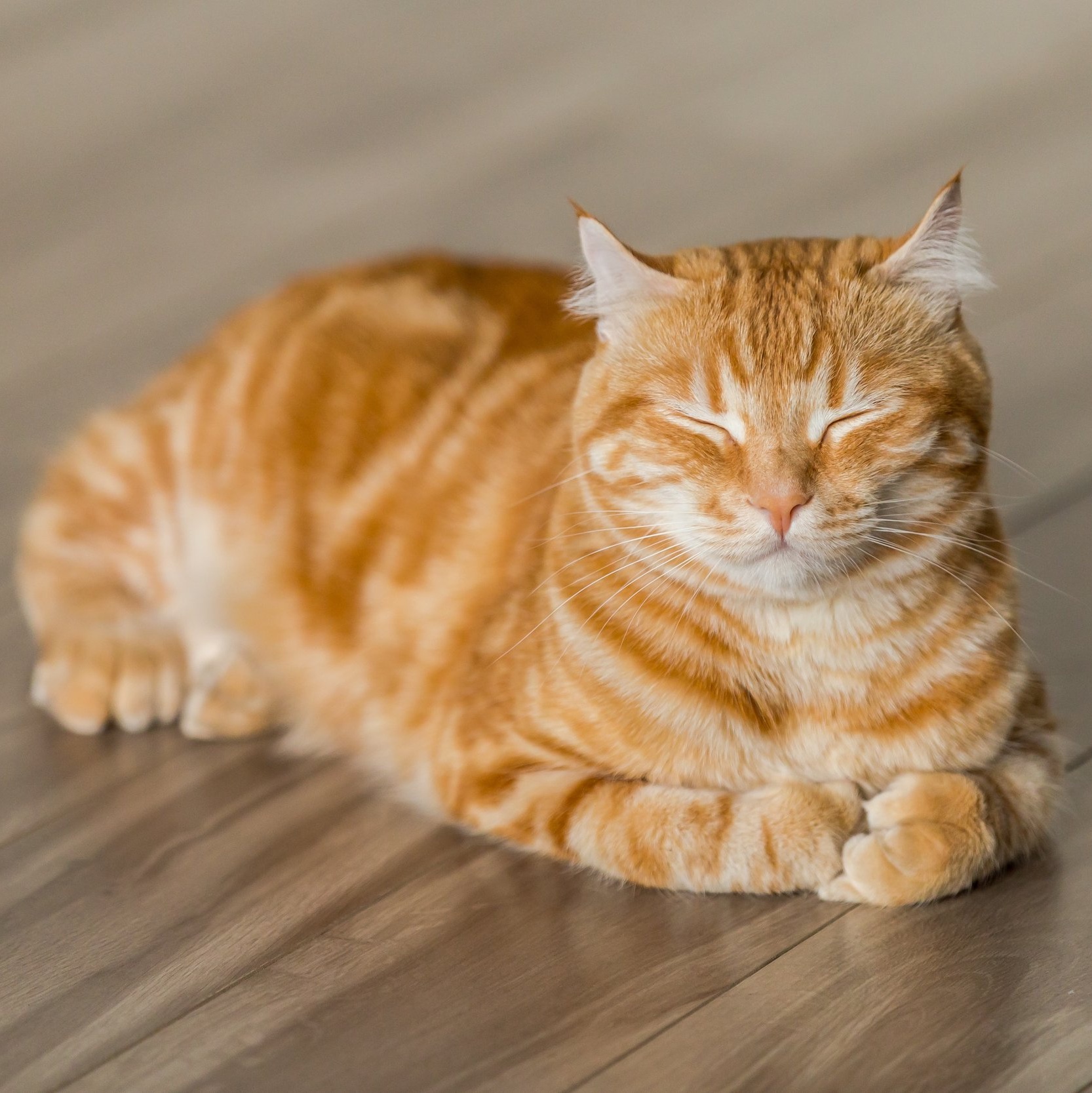}
  {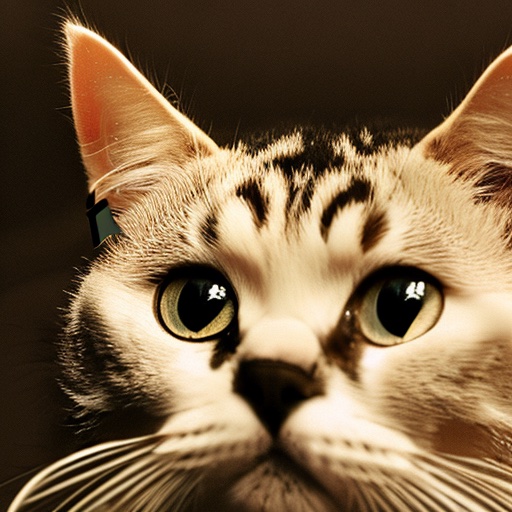}
  {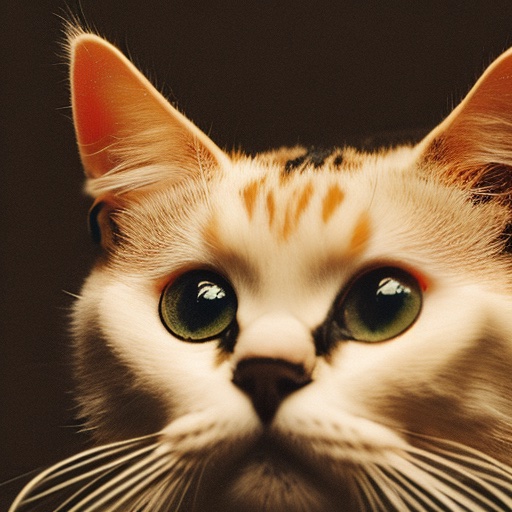}

\caption{Effect of PNO on text-only SDv2. Each row shows: the reference image (left), generation using only the text prompt (middle), and the result after applying PNO (right). PNO improves alignment with reference image features.}

\label{fig:pno_only_ablation}
\end{figure}

\begin{figure}[htb]
\centering
\newcommand{\resultspnointernal}[4]{
\begin{minipage}[t]{0.23\linewidth}
  \centering
  \includegraphics[width=\linewidth]{#1}
  \scriptsize Reference image
\end{minipage}
\hspace{2mm} 
\begin{minipage}[t]{0.23\linewidth}
  \centering
  \includegraphics[width=\linewidth]{#2}
  \scriptsize SDv2 (text-only)
\end{minipage}
\hspace{-1mm}
\begin{minipage}[t]{0.23\linewidth}
  \centering
  \includegraphics[width=\linewidth]{#3}
  \scriptsize VCF w/o PNO
\end{minipage}
\hspace{-1mm}
\begin{minipage}[t]{0.23\linewidth}
  \centering
  \includegraphics[width=\linewidth]{#4}
  \scriptsize VCF with PNO
\end{minipage}
\vspace{2mm}
}

\resultspnointernal
  {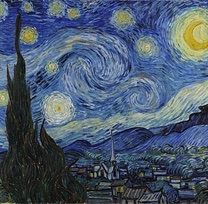}
  {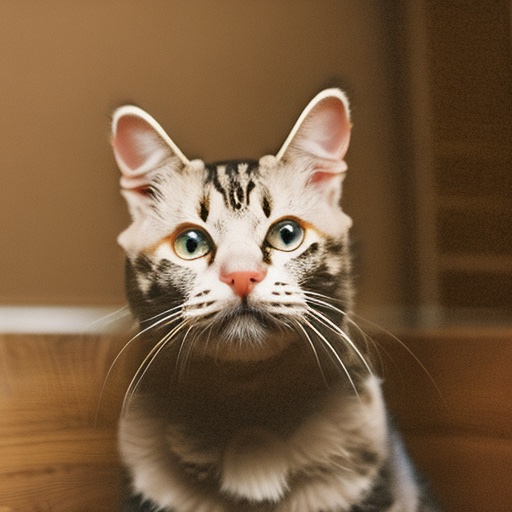}
  {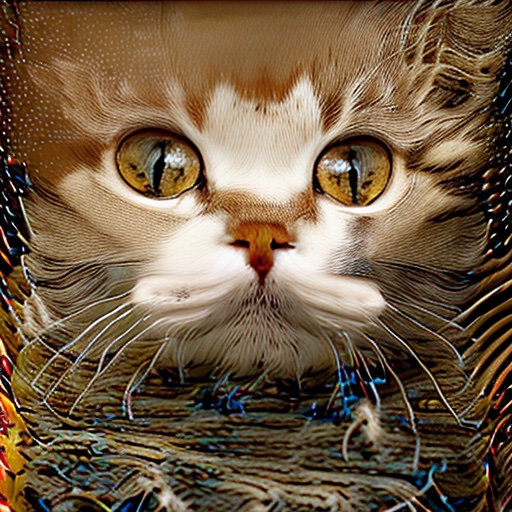}
  {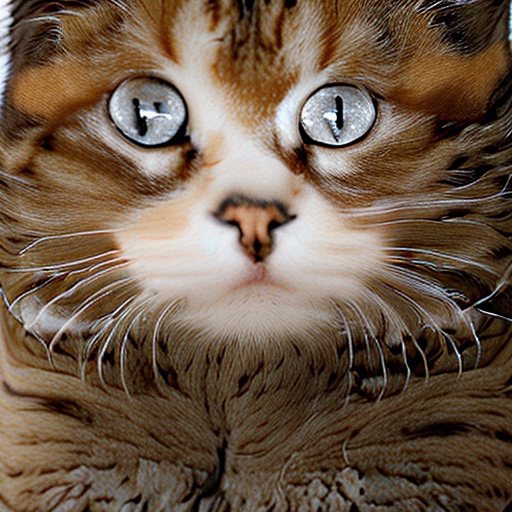}
\resultspnointernal
  {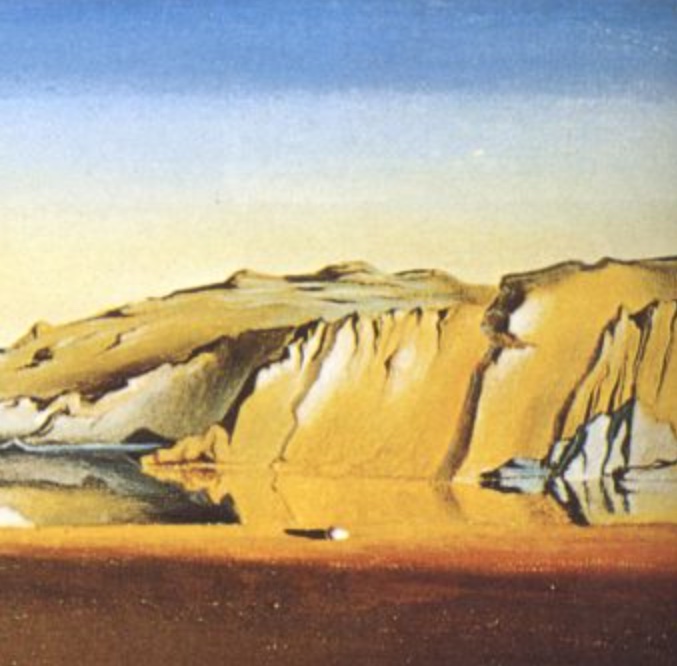}
  {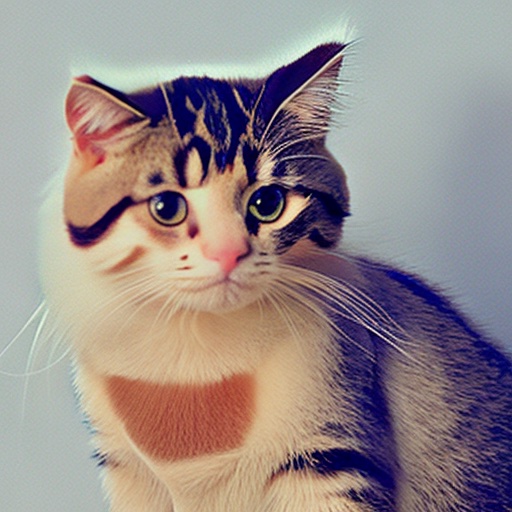}
  {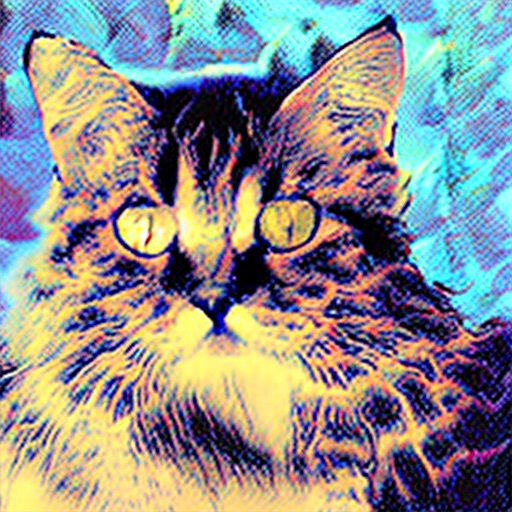}
  {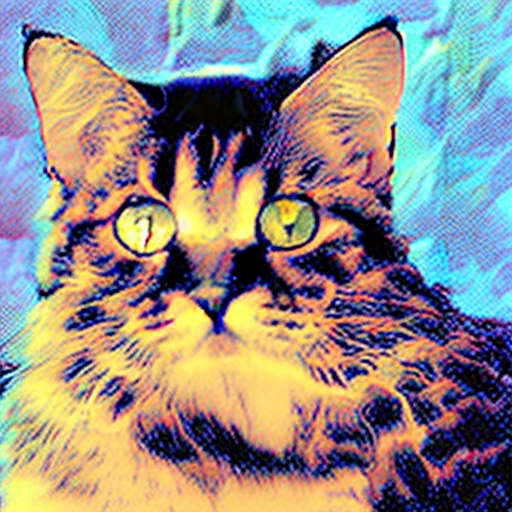}
\caption{Qualitative results of Prompt-Noise Optimization (PNO). Each row shows a reference image, the text-only generation from Stable Diffusion v2, our VCF pipeline without PNO, and our VCF pipeline with PNO. PNO can reduce noise (top row) and improve adherence to reference image details like color patterns (bottom row, more orange stripes).}
\label{fig:pno_ablations}
\end{figure}

\autoref{fig:pno_only_ablation} illustrates the effect of applying PNO to the text-only SDv2 model. Even without fusion-based guidance, incorporating a reference image during the optimisation process leads to outputs that exhibit improved structure and visual similarity to the reference. \autoref{fig:pno_ablations} shows the qualitative effect of PNO when applied to generations produced using the cross-attention fusion method, with an image guidance strength of \(\alpha = 0.3\). The number of PNO steps is fixed at 50.

\noindent
These results highlight the value of PNO as a refinement mechanism. In \autoref{fig:pno_only_ablation}, PNO improves both structural alignment and fidelity to the reference image, even in the absence of explicit fusion. When used in conjunction with cross-attention fusion, PNO helps suppress visual noise and artefacts introduced during fusion (e.g., \autoref{fig:pno_ablations}, top row), and can further steer the output toward reference-specific details (e.g., \autoref{fig:pno_ablations}, bottom row). For instance, PNO enhances colour fidelity by amplifying characteristic features such as the orange stripes on the cat. Overall, these qualitative examples suggest that PNO consistently improves both perceptual alignment with the reference image and the visual quality of the generated output.

\section{Additional Qualitative Examples of Main Results}
An interesting observation is that image guidance becomes particularly crucial when the text prompt is somewhat vague or abstract. This is exemplified clearly in Figure \ref{fig:appendix2-main-results}, where the default Stable Diffusion model (SDv2)—conditioned solely on text—struggles to generate coherent and meaningful characters from the prompt "A charming character emerging from the scene". However, introducing reference image conditioning significantly improves the quality, detail, and coherence of the generated characters, making them visually captivating and semantically meaningful. Additionally, the continued poor performance of naive fusion further emphasizes the complexity of effectively integrating visual and textual modalities. This highlights the challenging nature of the problem and demonstrates the effectiveness of our proposed fusion method, which significantly improves visual coherence and semantic alignment.
\label{app:more_qualitative_main}
\newlength{\appimgheight}
\setlength{\appimgheight}{1.6in}   

\newcommand{\appendixrow}[4]{%
  \begin{minipage}[t]{0.23\linewidth}\centering
    \includegraphics[width=\linewidth,height=\appimgheight,keepaspectratio]{#1}\par
    \scriptsize Reference image
  \end{minipage}\hspace{0.04\linewidth}%
  \begin{minipage}[t]{0.23\linewidth}\centering
    \includegraphics[width=\linewidth,height=\appimgheight,keepaspectratio]{#2}\par
    \scriptsize SDv2 (text-only)
  \end{minipage}\hspace{0.01\linewidth}%
  \begin{minipage}[t]{0.23\linewidth}\centering
    \includegraphics[width=\linewidth,height=\appimgheight,keepaspectratio]{#3}\par
    \scriptsize naive fusion
  \end{minipage}\hspace{0.01\linewidth}%
  \begin{minipage}[t]{0.23\linewidth}\centering
    \includegraphics[width=\linewidth,height=\appimgheight,keepaspectratio]{#4}\par
    \scriptsize VCF (Ours)
  \end{minipage}\par\vspace{0.9ex}%
}

\begin{figure}[H]  
  \centering
  \appendixrow{figures/main_results/1_1.jpg}
               {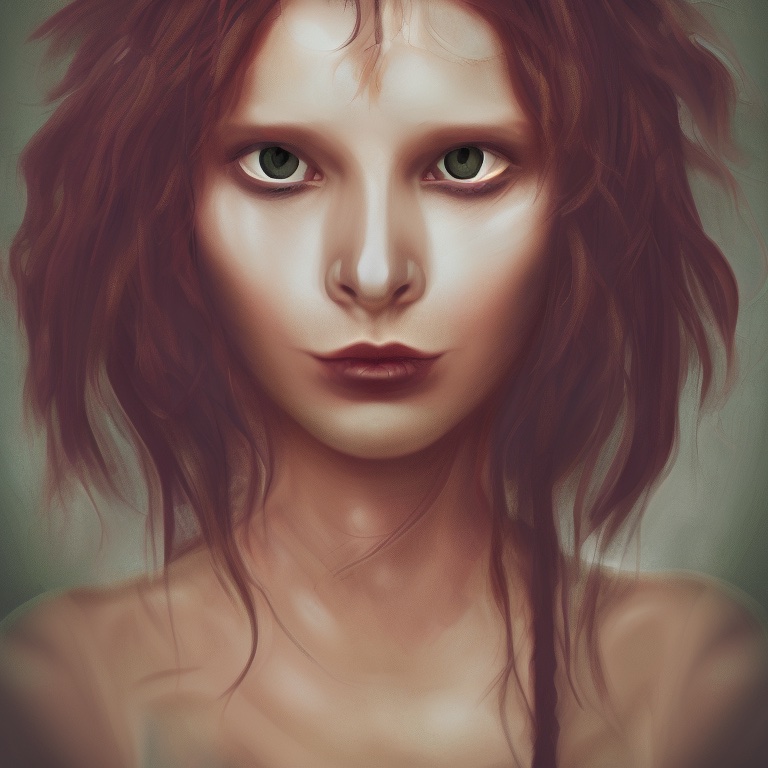}
               {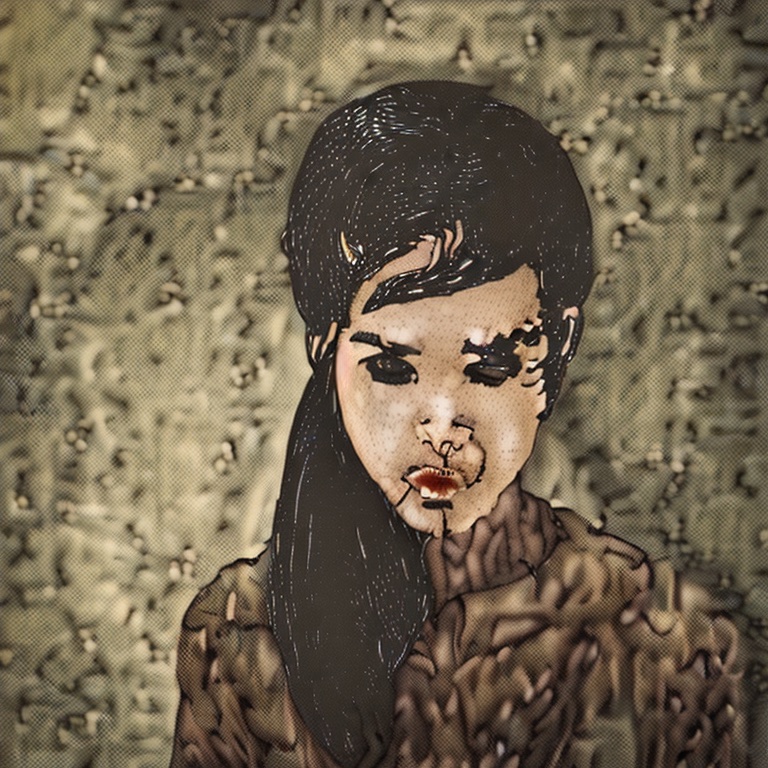}
               {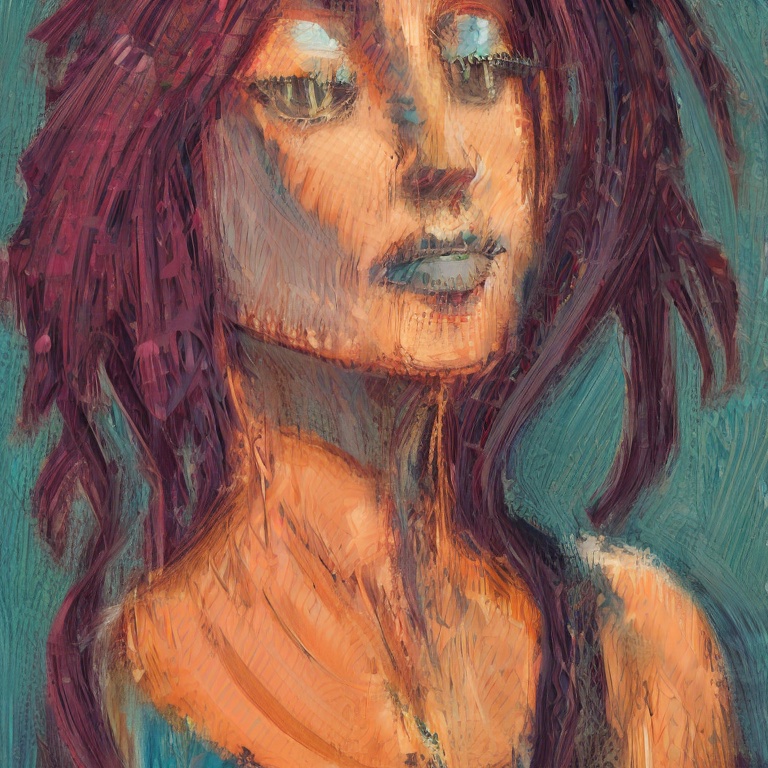}

  \appendixrow{figures/main_results/2_1.jpg}
               {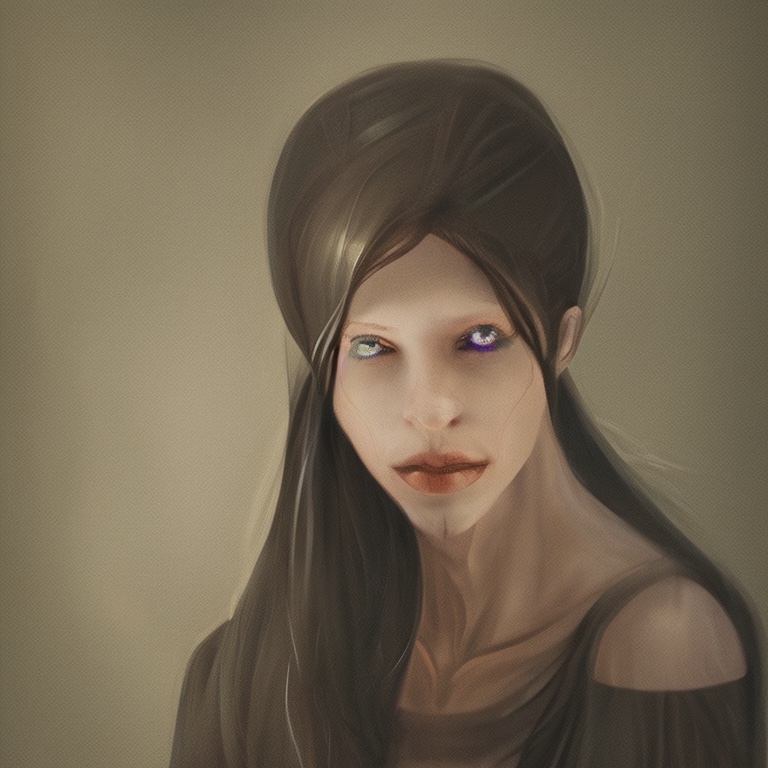}
               {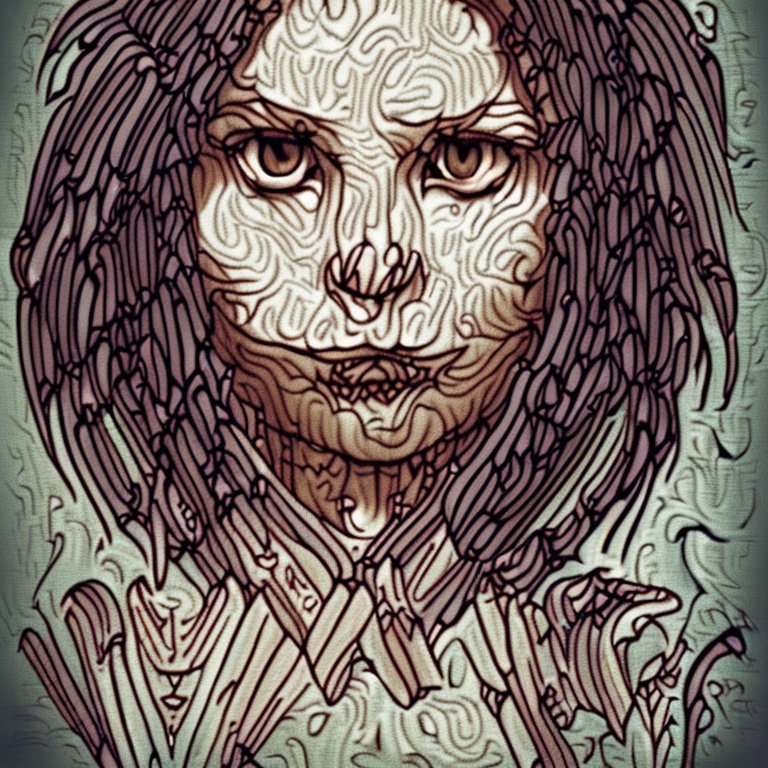}
               {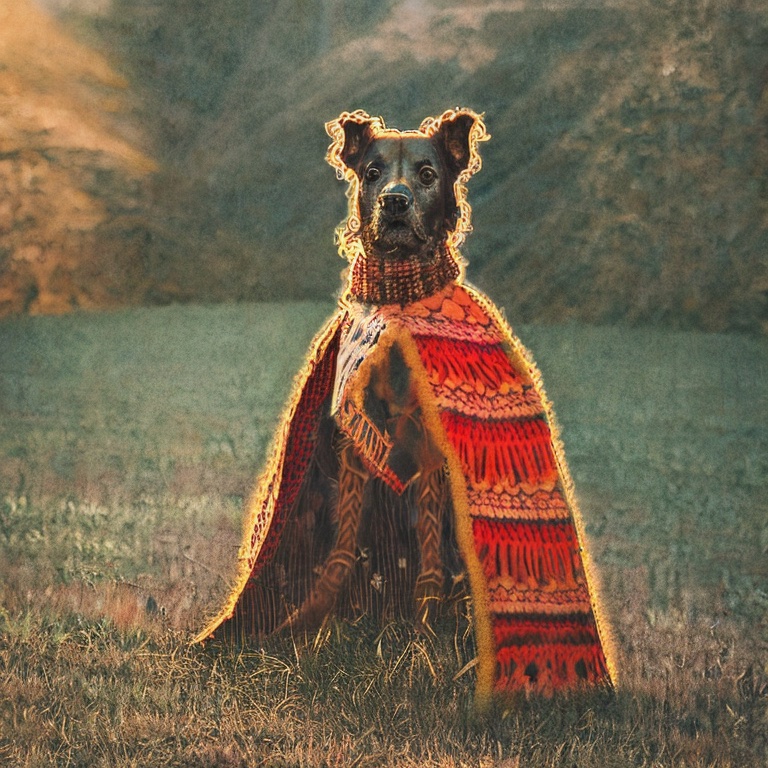}

  \appendixrow{figures/main_results/3_1.jpg}
               {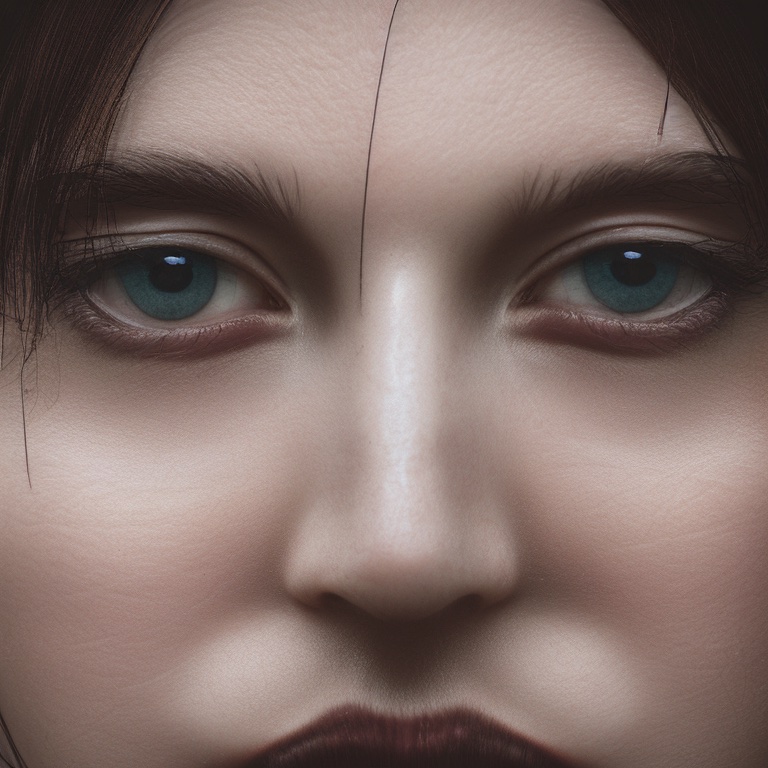}
               {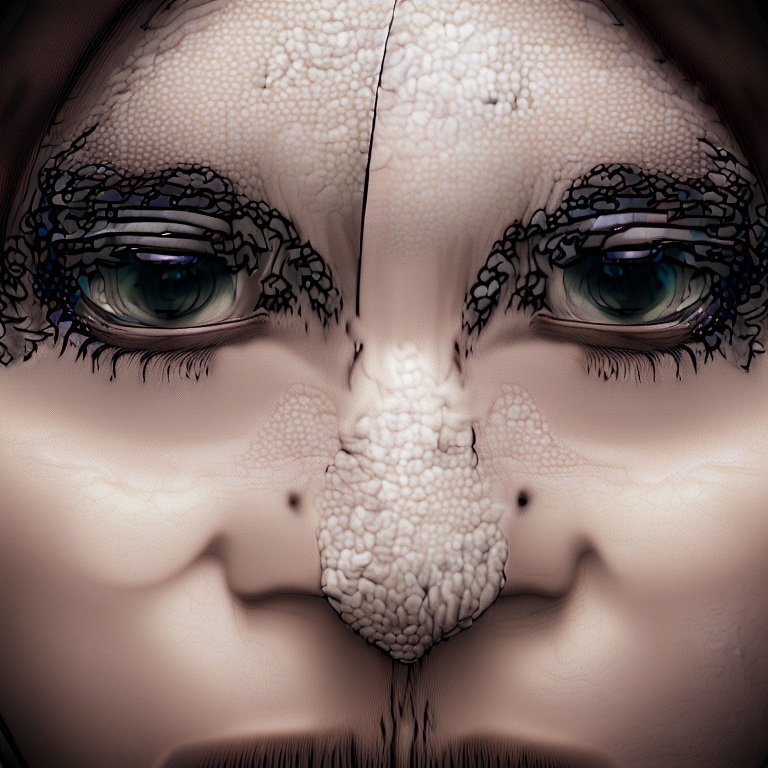}
               {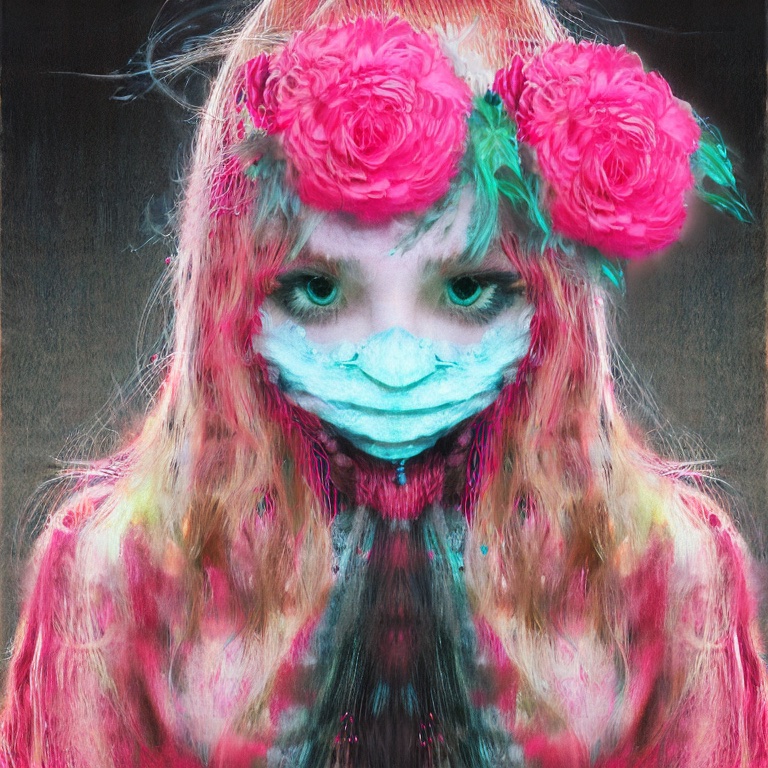}

  \caption{%
    Additional qualitative examples of the main results using the prompt "A beautiful portrait of a mysterious character".
    Each row shows (left~→~right): the reference image, baseline text-only SDv2 output, naive fusion, and our proposed VCF.%
  }
  \label{fig:appendix-main-results}
\end{figure}

\setlength{\appimgheight}{1.6in}   

\renewcommand{\appendixrow}[4]{%
  \begin{minipage}[t]{0.23\linewidth}\centering
    \includegraphics[width=\linewidth,height=\appimgheight,keepaspectratio]{#1}\par
    \scriptsize Reference image
  \end{minipage}\hspace{0.04\linewidth}%
  \begin{minipage}[t]{0.23\linewidth}\centering
    \includegraphics[width=\linewidth,height=\appimgheight,keepaspectratio]{#2}\par
    \scriptsize SDv2 (text-only)
  \end{minipage}\hspace{0.01\linewidth}%
  \begin{minipage}[t]{0.23\linewidth}\centering
    \includegraphics[width=\linewidth,height=\appimgheight,keepaspectratio]{#3}\par
    \scriptsize Naive fusion
  \end{minipage}\hspace{0.01\linewidth}%
  \begin{minipage}[t]{0.23\linewidth}\centering
    \includegraphics[width=\linewidth,height=\appimgheight,keepaspectratio]{#4}\par
    \scriptsize VCF (Ours)
  \end{minipage}\par\vspace{0.9ex}%
}

\begin{figure}[H]  
  \centering
  \appendixrow{figures/main_results/1_1.jpg}
               {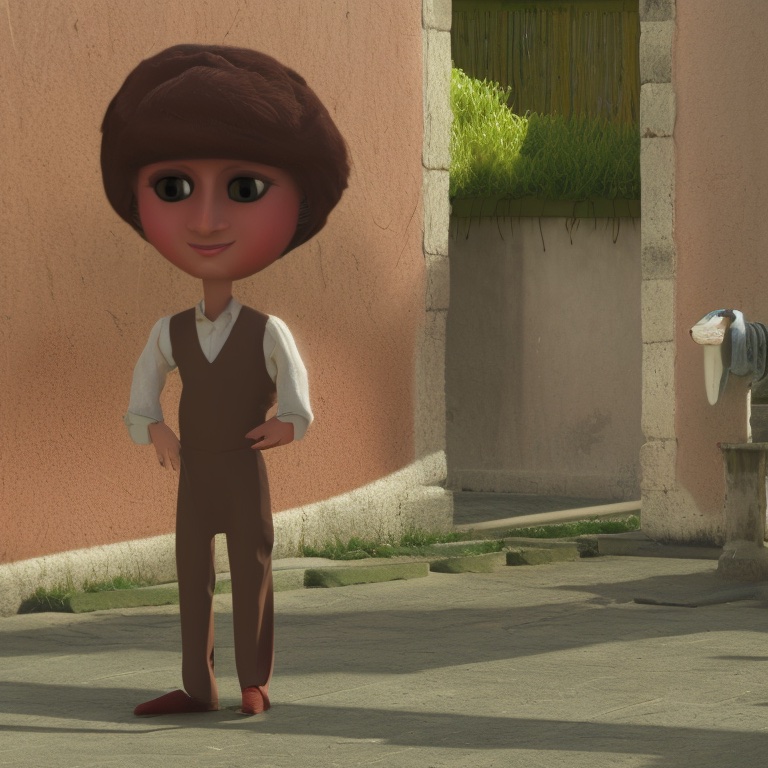}
               {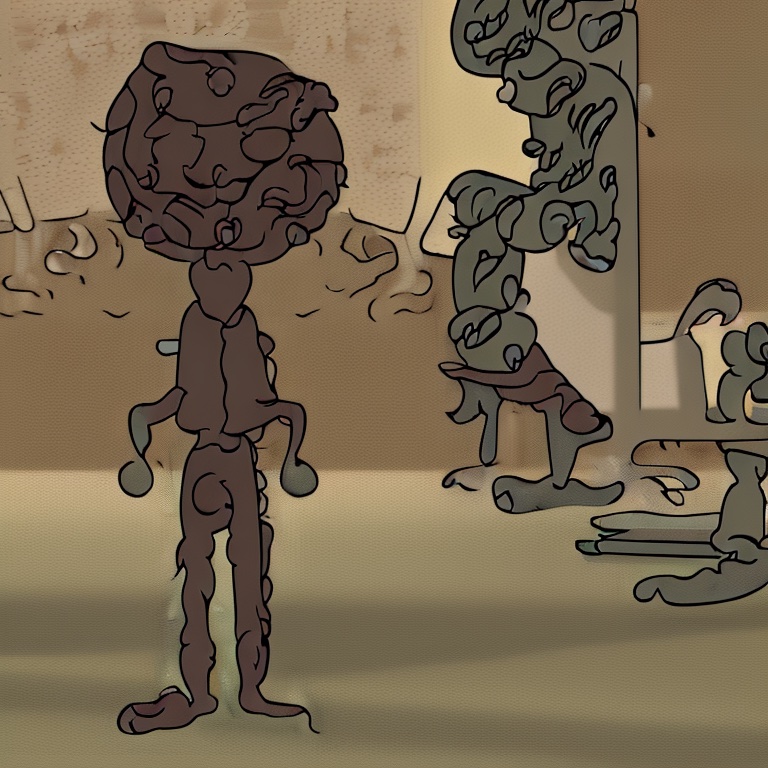}
               {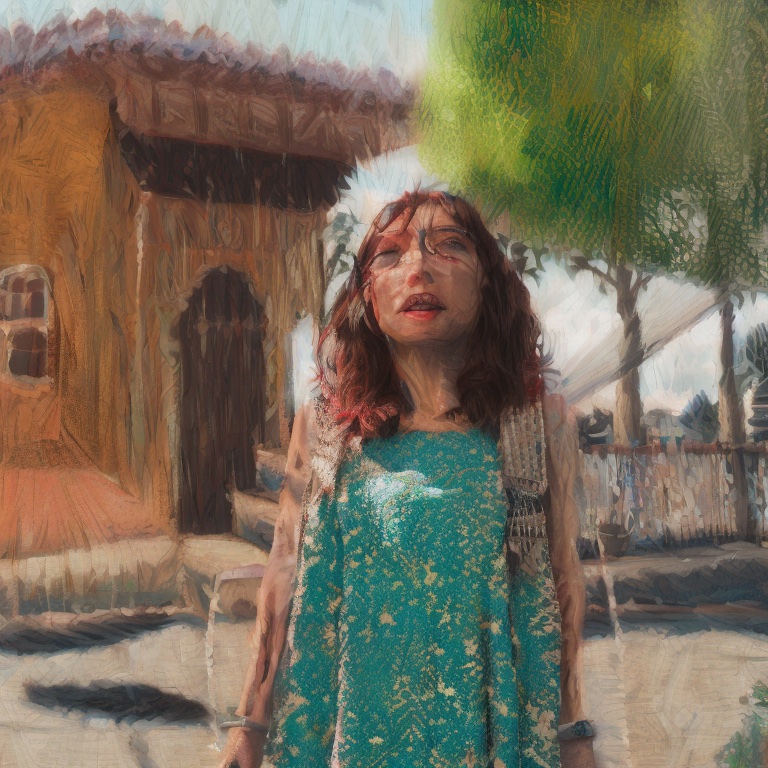}

  \appendixrow{figures/main_results/2_1.jpg}
               {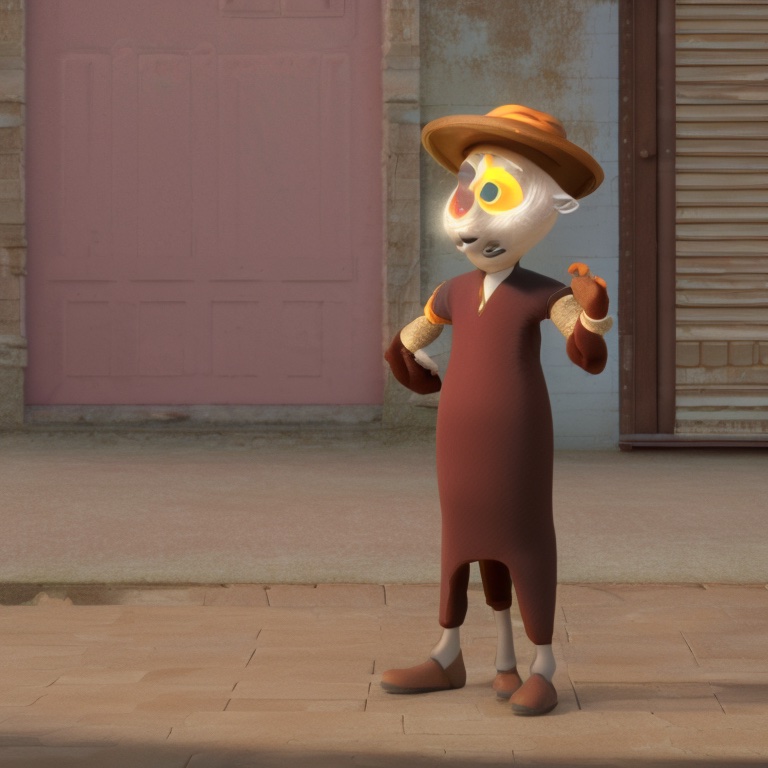}
               {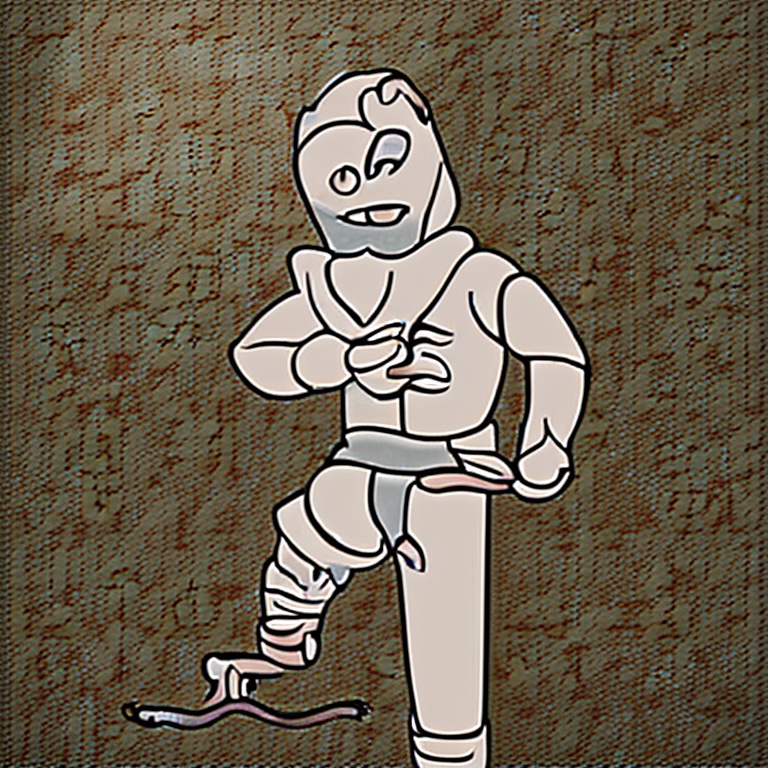}
               {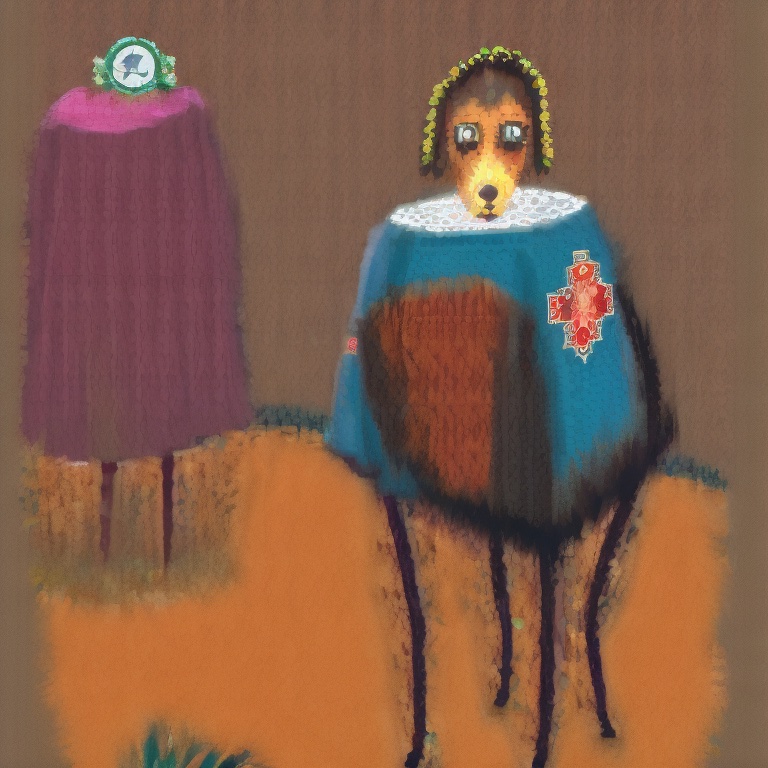}

  \appendixrow{figures/main_results/3_1.jpg}
               {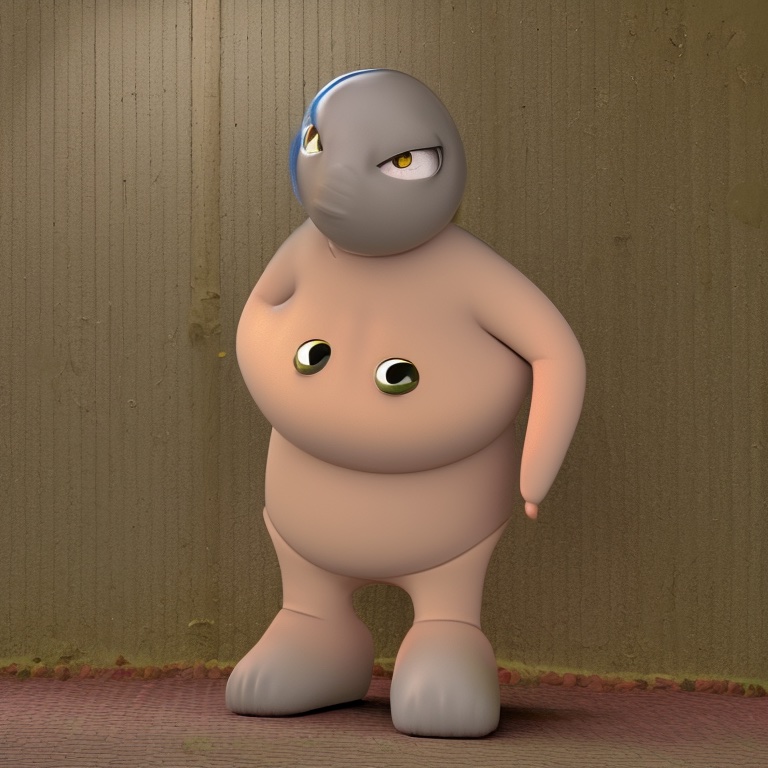}
               {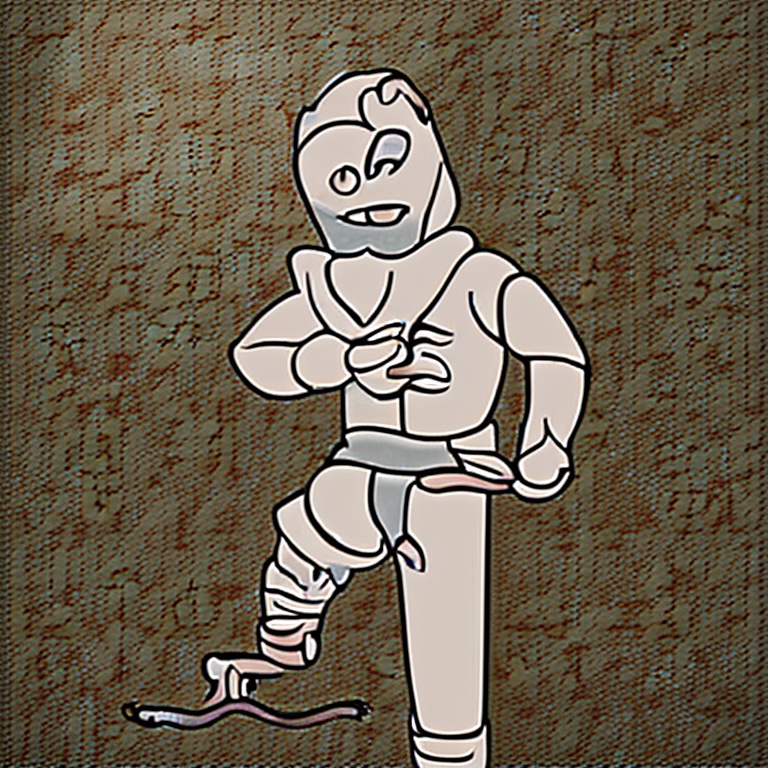}
               {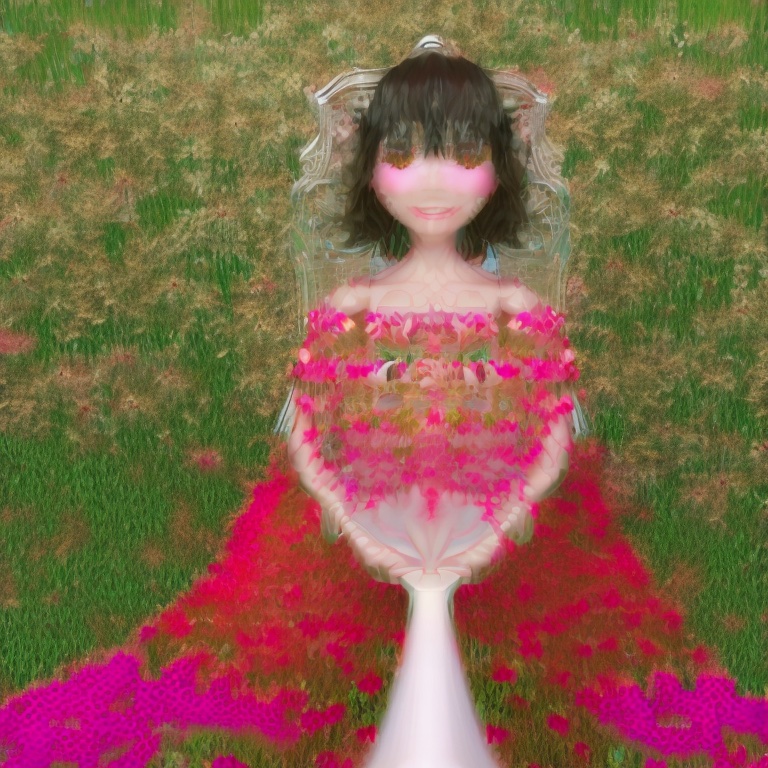}

  \caption{%
    Additional qualitative examples of the main results using the prompt "A charming character emerging from the scene".
    Each row shows (left~→~right): the reference image, baseline text-only SDv2 output, naive fusion, and our proposed VCF.%
  }
  \label{fig:appendix2-main-results}
\end{figure}

\setlength{\appimgheight}{1.6in}   

\renewcommand{\appendixrow}[4]{%
  \begin{minipage}[t]{0.23\linewidth}\centering
    \includegraphics[width=\linewidth,height=\appimgheight,keepaspectratio]{#1}\par
    \scriptsize Reference image
  \end{minipage}\hspace{0.04\linewidth}%
  \begin{minipage}[t]{0.23\linewidth}\centering
    \includegraphics[width=\linewidth,height=\appimgheight,keepaspectratio]{#2}\par
    \scriptsize SDv2 (text-only)
  \end{minipage}\hspace{0.01\linewidth}%
  \begin{minipage}[t]{0.23\linewidth}\centering
    \includegraphics[width=\linewidth,height=\appimgheight,keepaspectratio]{#3}\par
    \scriptsize naive fusion
  \end{minipage}\hspace{0.01\linewidth}%
  \begin{minipage}[t]{0.23\linewidth}\centering
    \includegraphics[width=\linewidth,height=\appimgheight,keepaspectratio]{#4}\par
    \scriptsize VCF (Ours)
  \end{minipage}\par\vspace{0.9ex}%
}

\begin{figure}[H]  
  \centering
  \appendixrow{figures/main_results/1_1.jpg}
               {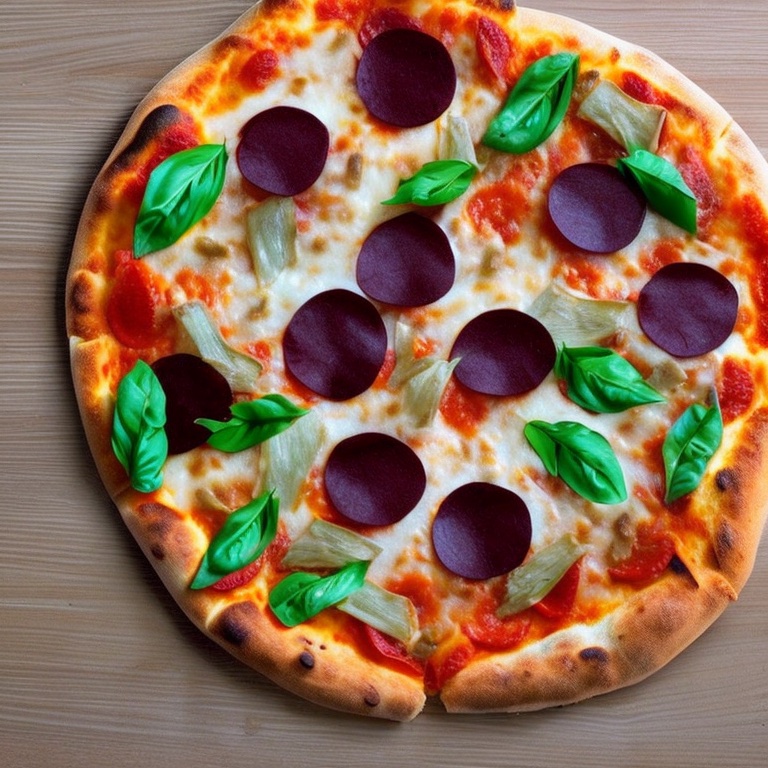}
               {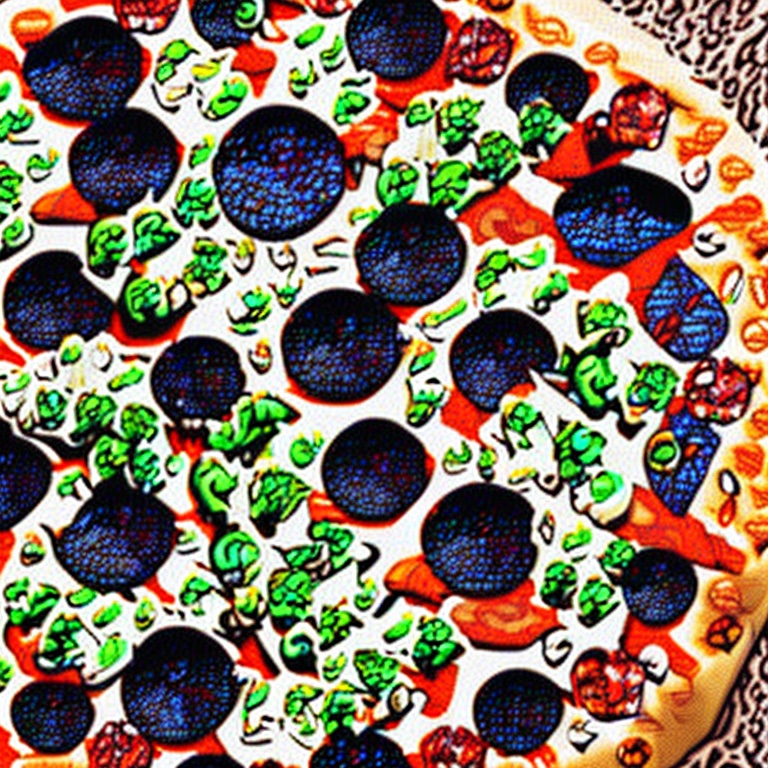}
               {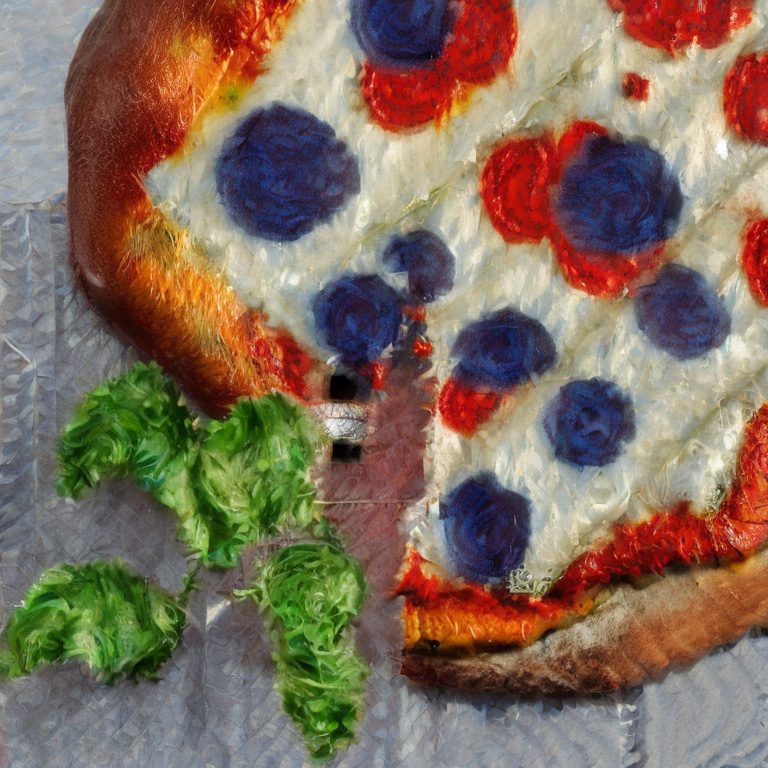}

  \appendixrow{figures/main_results/2_1.jpg}
               {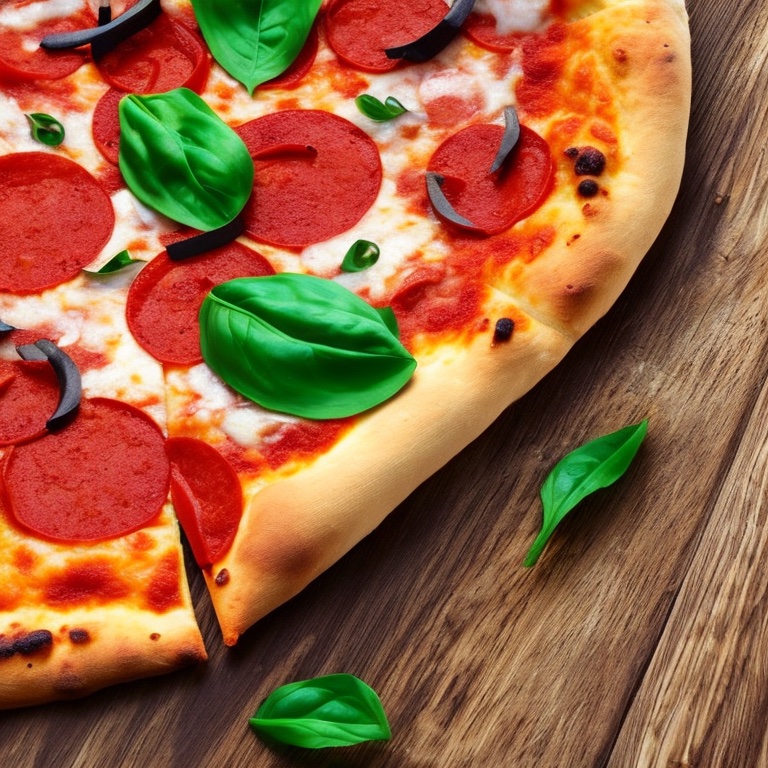}
               {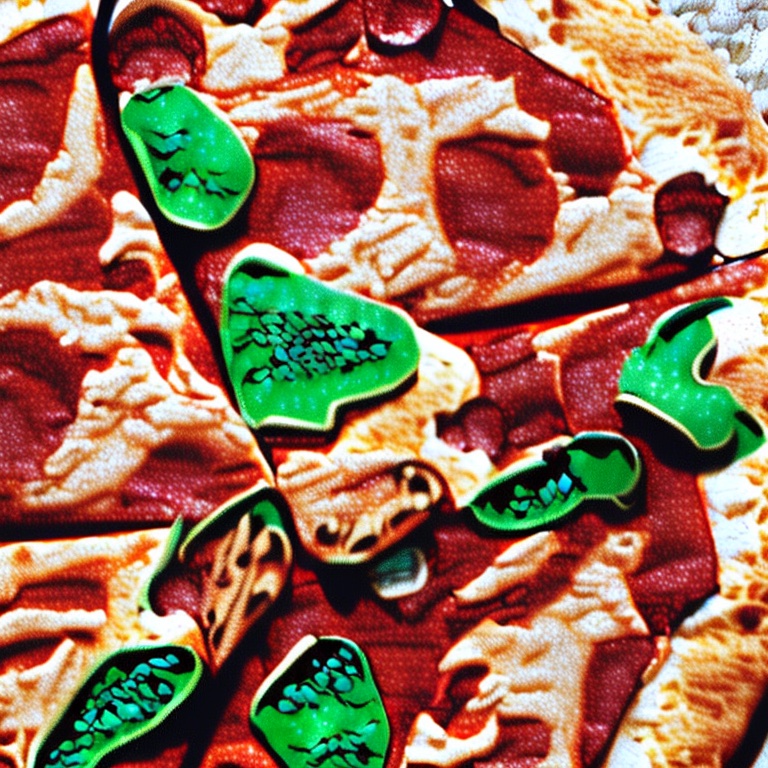}
               {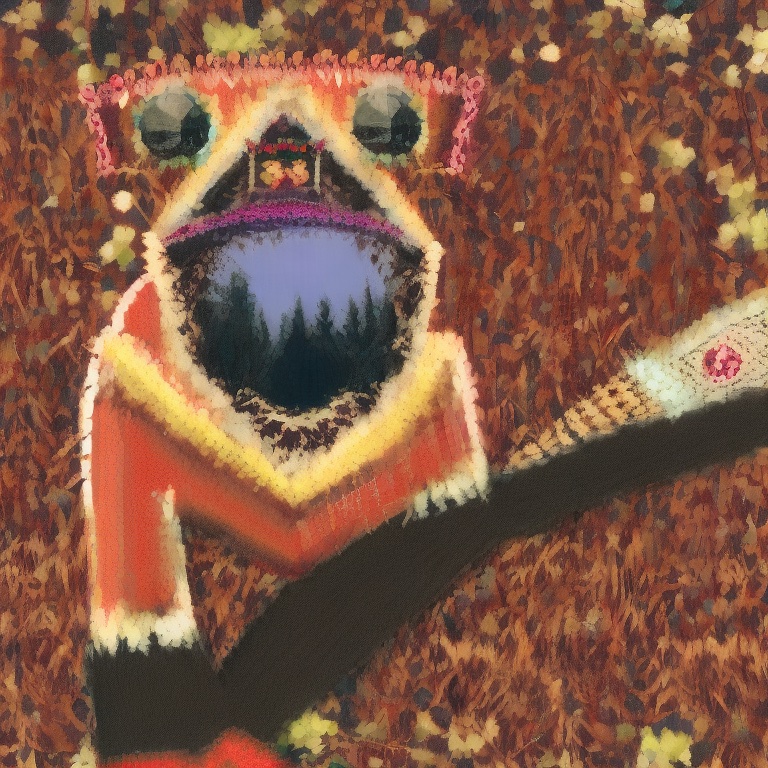}

  \appendixrow{figures/main_results/3_1.jpg}
               {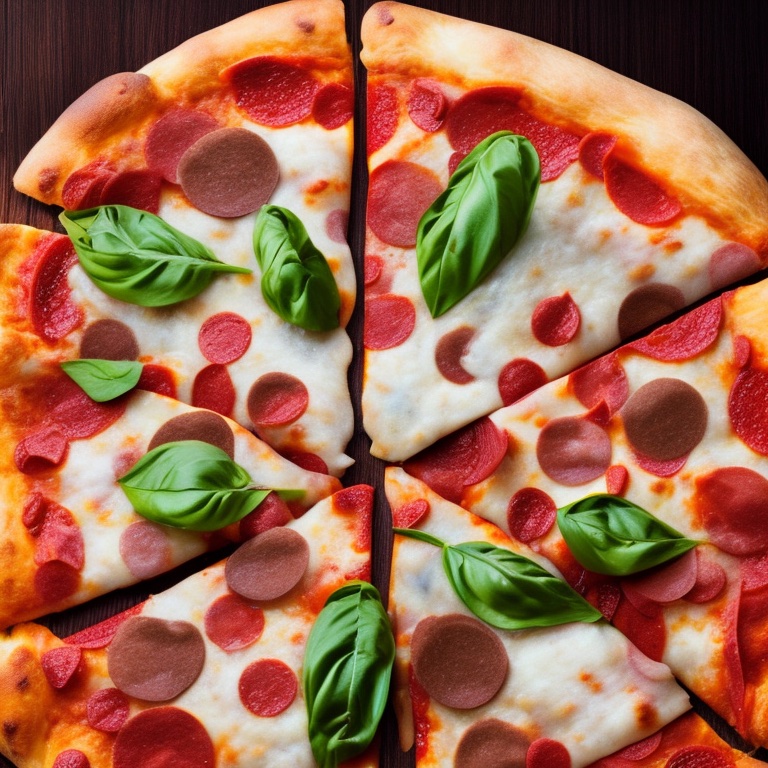}
               {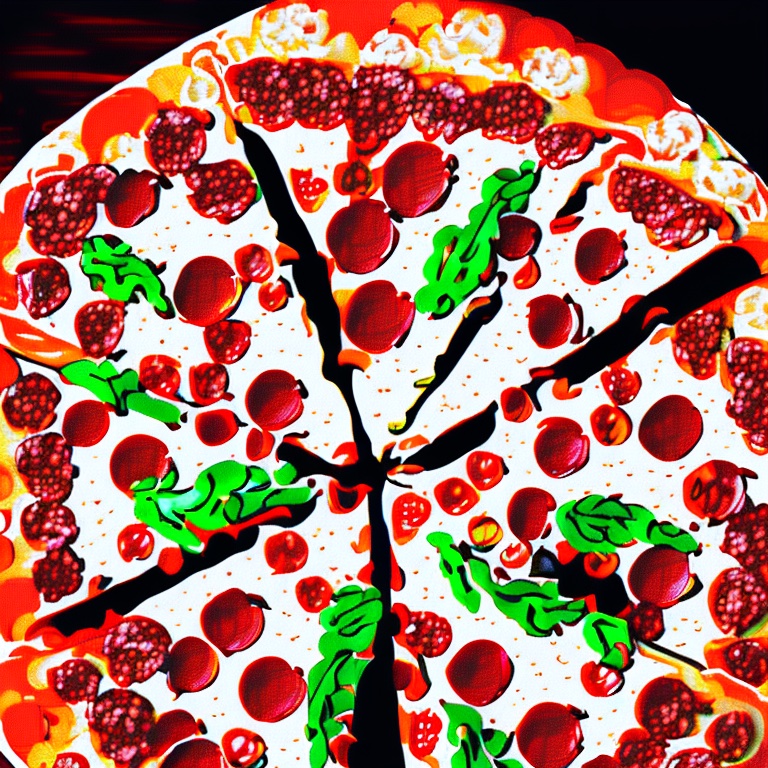}
               {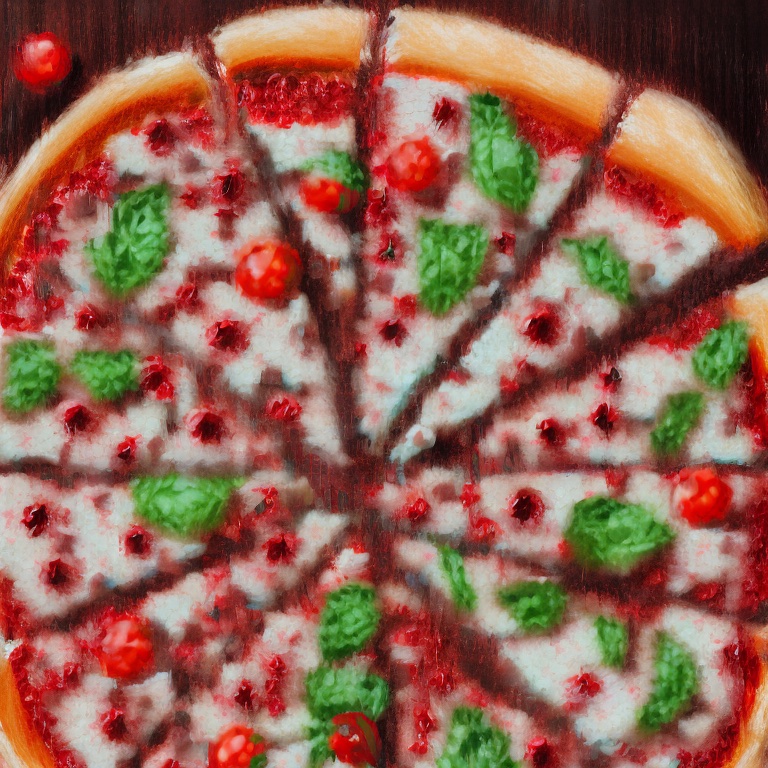}

  \caption{%
    Additional qualitative examples of the main results using the prompt "a delicious pizza".
    Each row shows (left~→~right): the reference image, baseline text-only SDv2 output, naive fusion, and our proposed VCF.%
  }
  \label{fig:appendix3-main-results}
\end{figure}

\section{Cross-Attention Fusion}
\label{app:cross_attention_fusion}
As an alternative to concatenation, we experimented with a cross-attention fusion scheme.  
The idea is to let the text tokens query the aligned image tokens, thereby injecting fine-grained visual cues into the conditioning stream.

\paragraph{Fusion mechanism.}
Given text tokens \(T\) and aligned image tokens \(\hat{I}\), we compute
\[
T_{\text{fused}}
   = \operatorname{Attn}(Q=T,\;K=\hat{I},\;V=\hat{I}),
\]
and blend the result with the original text tokens,
\[
T_{\text{final}}
   = (1-\alpha)\,T
   + \alpha\,\gamma\,T_{\text{fused}},
\]
where \(\alpha\in[0,1]\) sets the overall weight of the image signal.  
The factor \(\gamma\) rescales \(T_{\text{fused}}\) at every denoising step so that its norm remains comparable to that of \(T\).  

\paragraph{Qualitative observations.}
Representative outputs are shown in \autoref{fig:fusion-ablation-fullpage}.  
Cross-attention fusion does transfer some reference features, but the resulting images are noticeably noisier and less coherent than those produced by concatenation fusion, and in several cases introduce artefacts not present in either the prompt or the reference.  
Hence we retain this variant only for completeness and defer to concatenation fusion in the main paper.

\newlength{\fusionimgheight}
\setlength{\fusionimgheight}{1.6in}  

\newcommand{\fusionrowwide}[4]{%
  \begin{minipage}[t]{0.235\linewidth}\centering
    \includegraphics[width=\linewidth,height=\fusionimgheight,keepaspectratio]{#1}\par
    \scriptsize Reference image
  \end{minipage}\hspace{0.02\linewidth}
  \begin{minipage}[t]{0.235\linewidth}\centering
    \includegraphics[width=\linewidth,height=\fusionimgheight,keepaspectratio]{#2}\par
    \scriptsize SDv2 (text-only)
  \end{minipage}\hspace{0.015\linewidth}%
  \begin{minipage}[t]{0.235\linewidth}\centering
    \includegraphics[width=\linewidth,height=\fusionimgheight,keepaspectratio]{#3}\par
    \scriptsize Naive fusion
  \end{minipage}\hspace{0.015\linewidth}%
  \begin{minipage}[t]{0.235\linewidth}\centering
    \includegraphics[width=\linewidth,height=\fusionimgheight,keepaspectratio]{#4}\par
    \scriptsize Cross-attention fusion
  \end{minipage}\par\vspace{0.5ex}%
}

\begin{figure}[H]   
  \centering
  \fusionrowwide{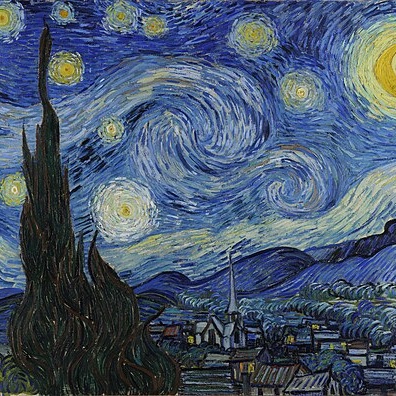}
                {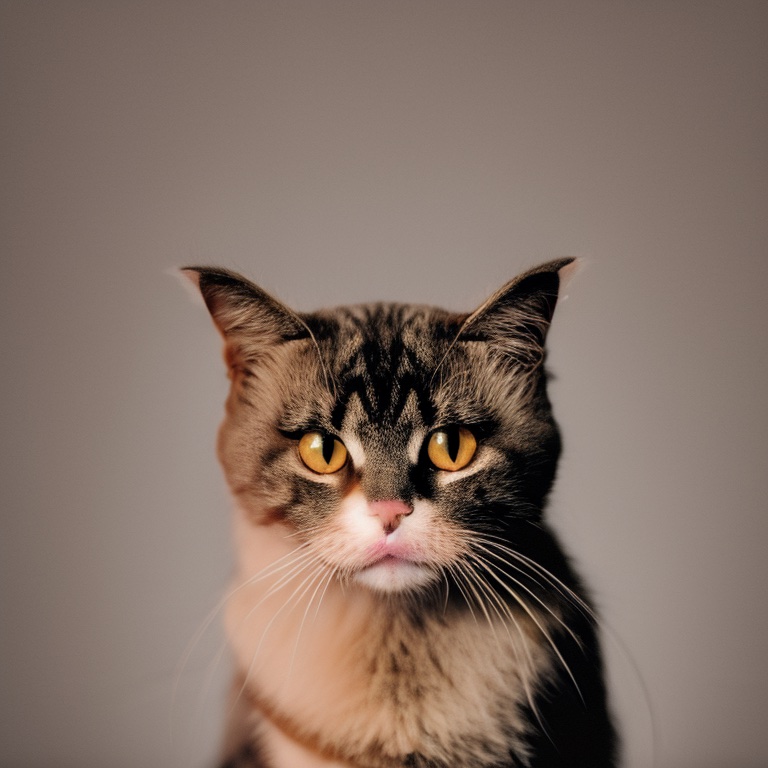}
                {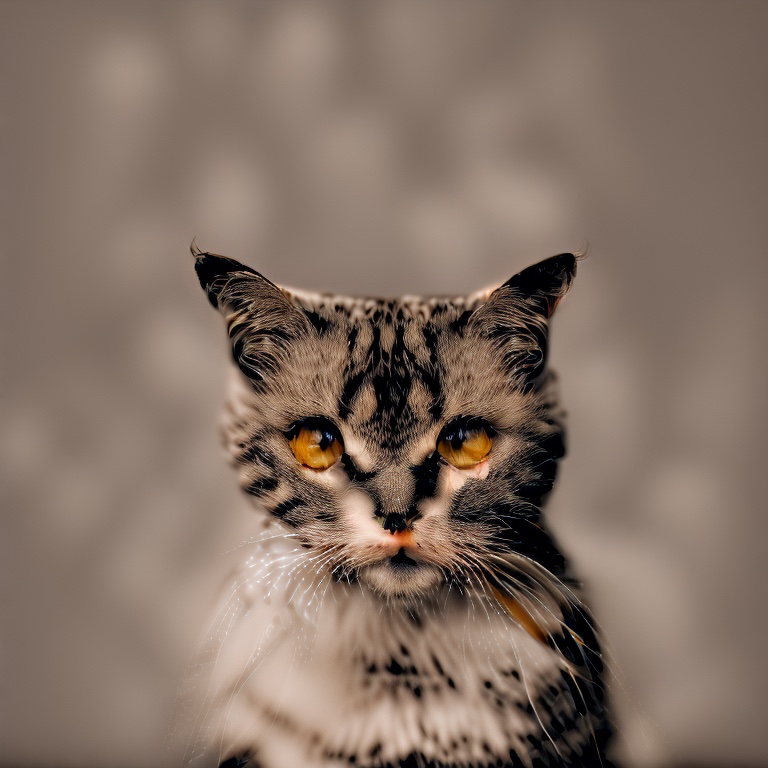}
                {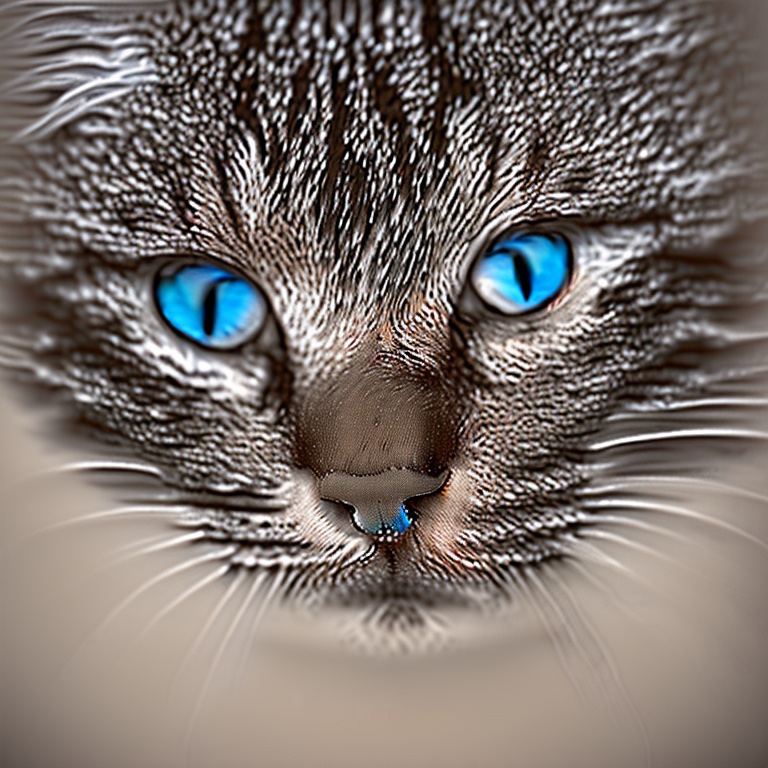}

  \fusionrowwide{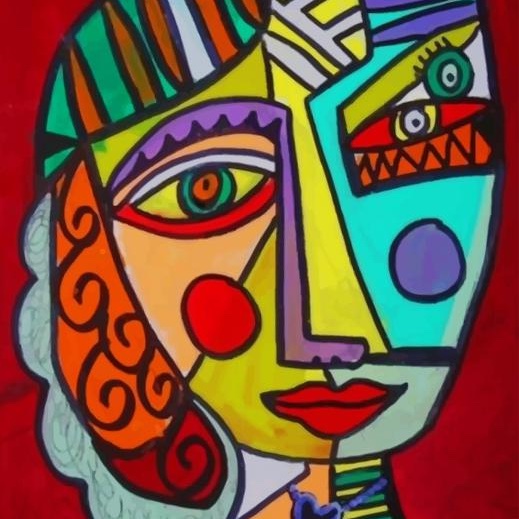}
                {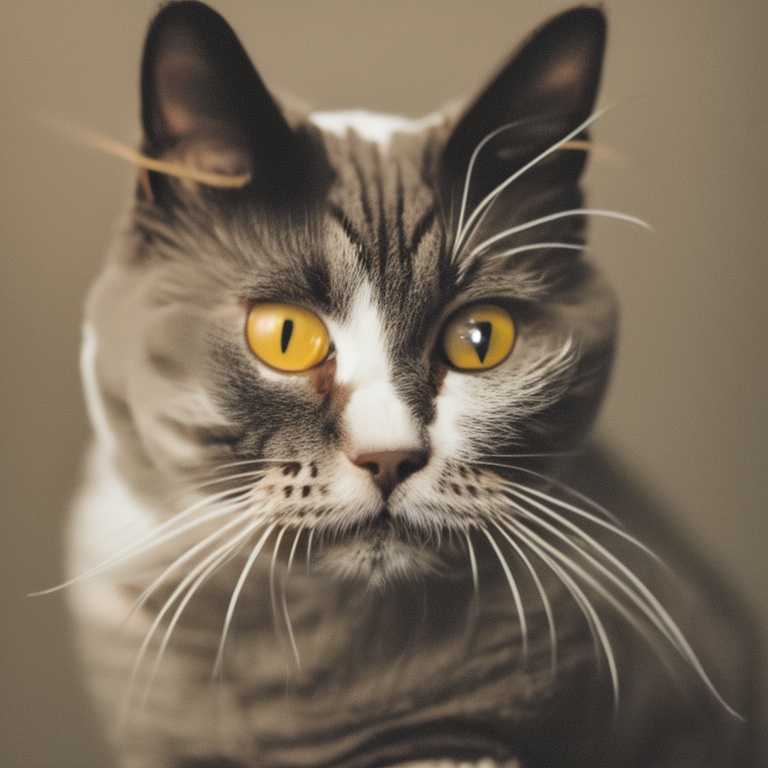}
                {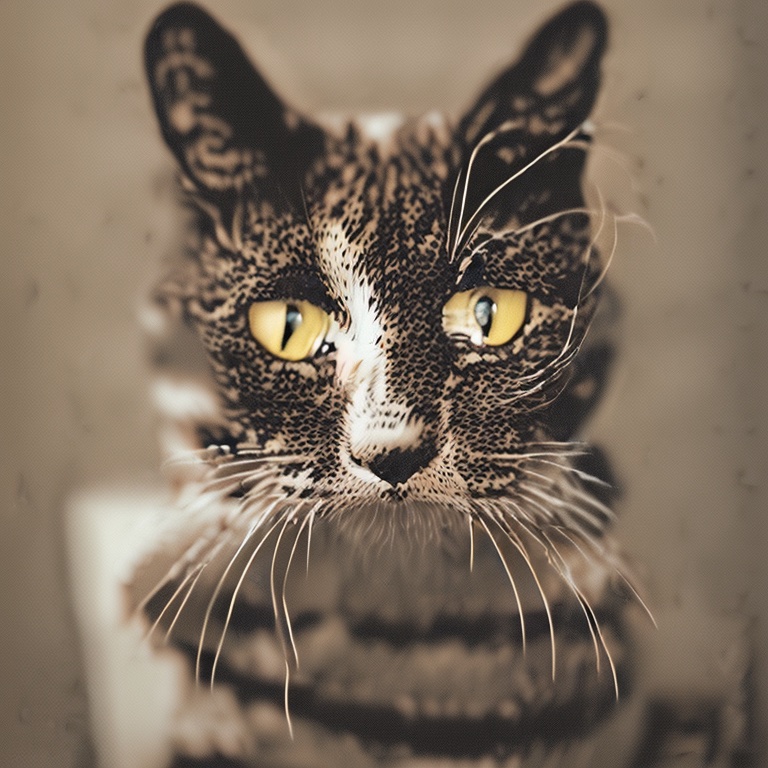}
                {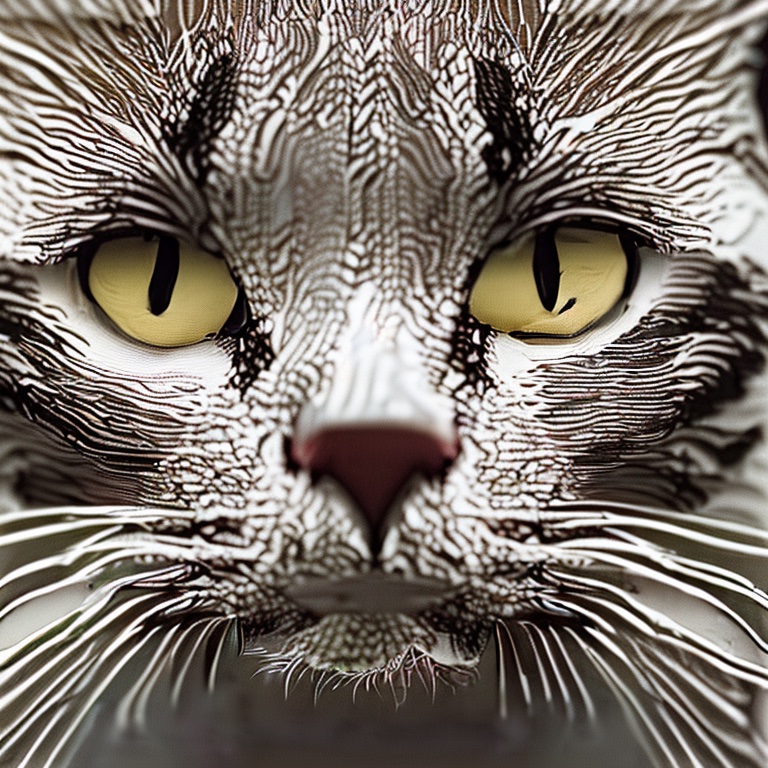}

  \fusionrowwide{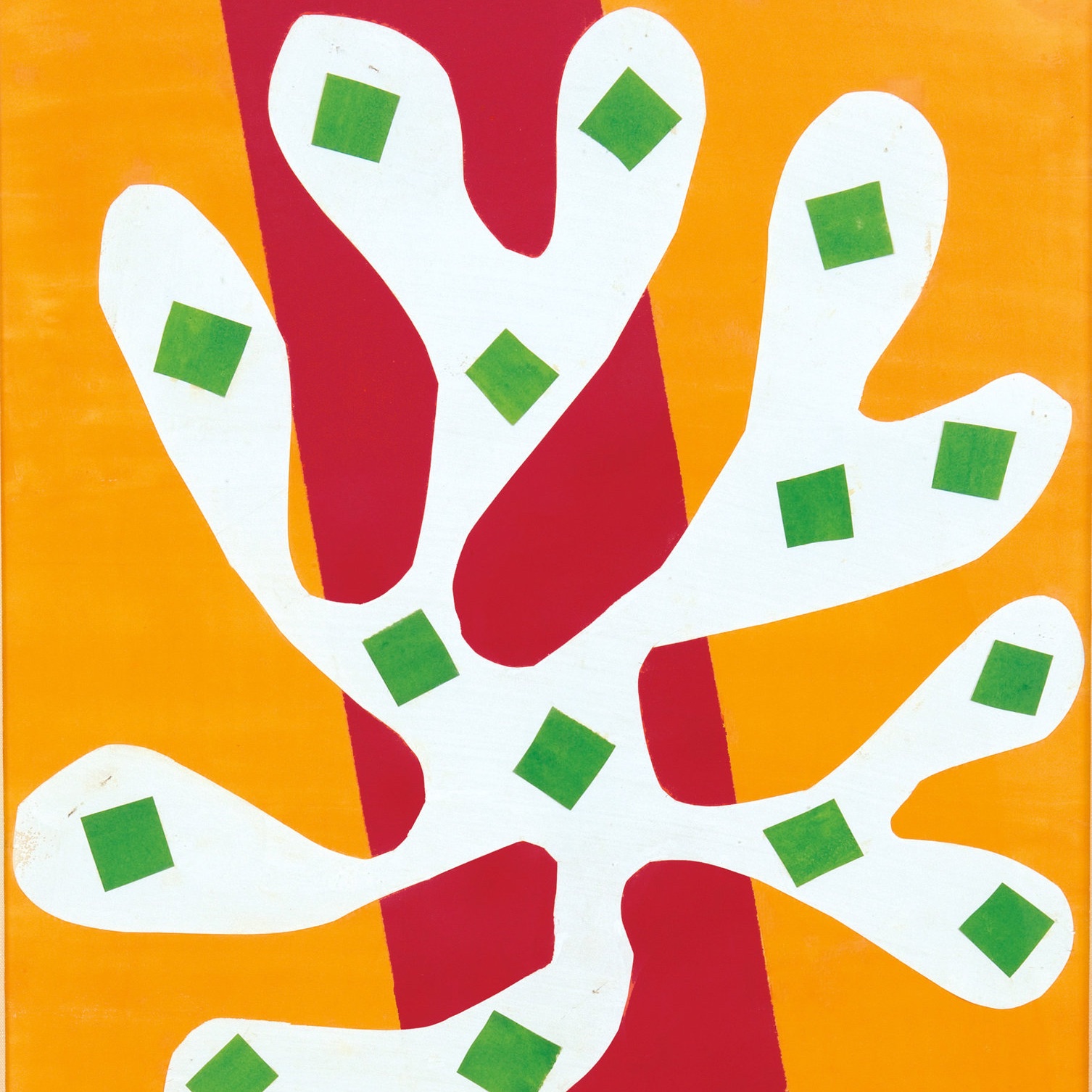}
                {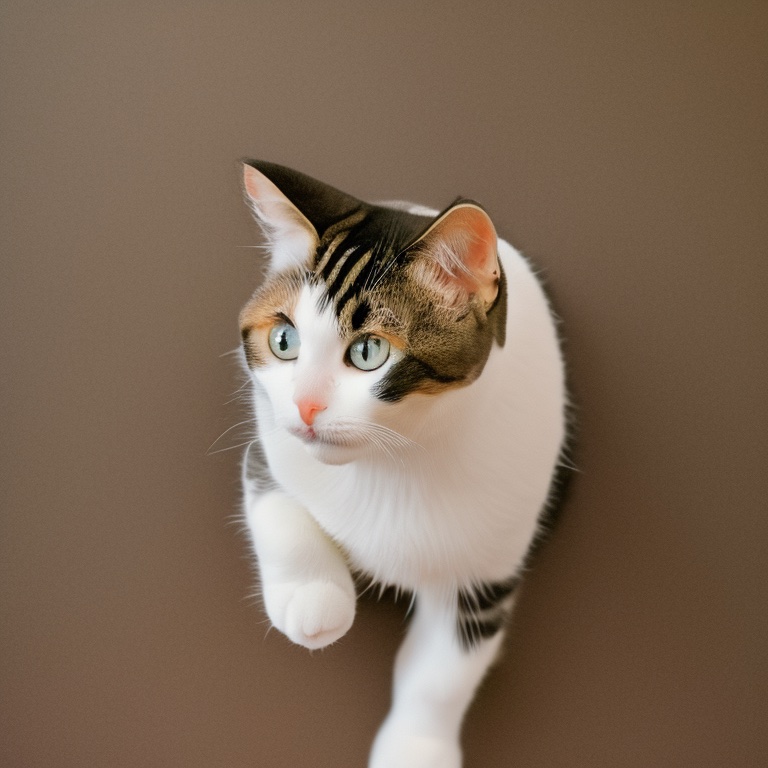}
                {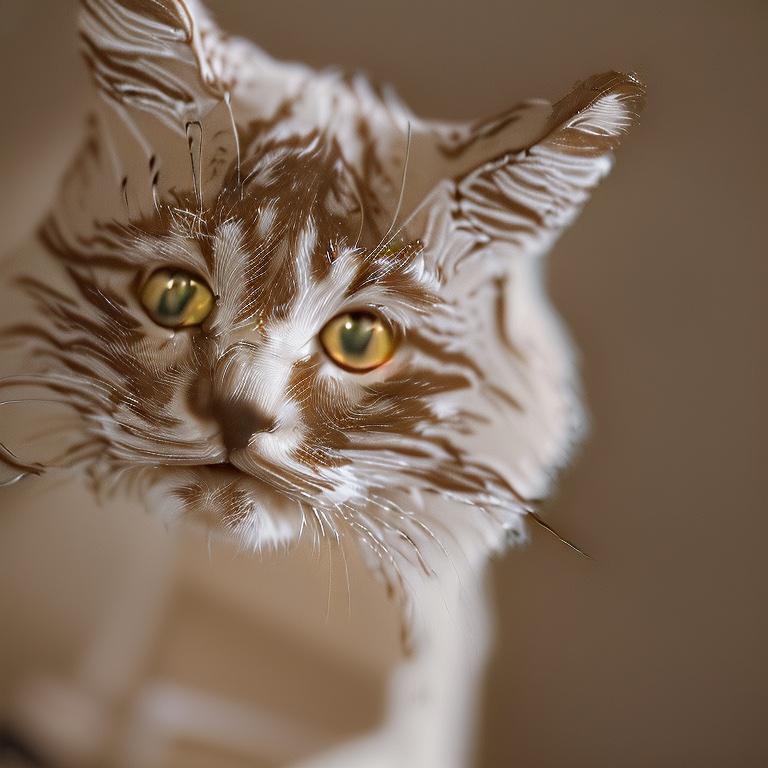}
                {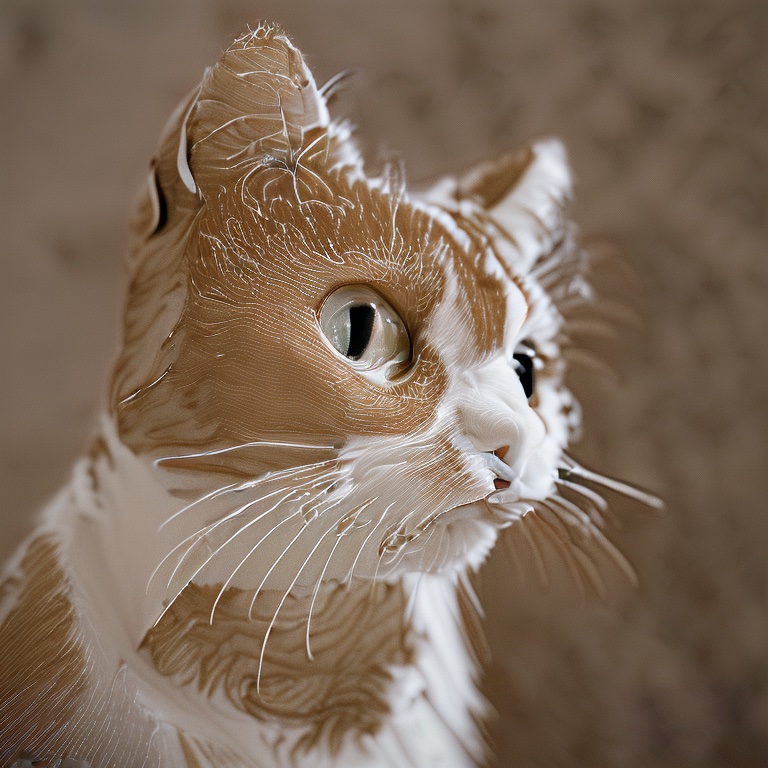}

  \caption{%
    Qualitative ablation on fusion strategy.  
    Each row shows (left → right): the reference image, baseline text-only SDv2 output, naive token fusion, and cross-attention fusion.%
  }
  \label{fig:fusion-ablation-fullpage}
\end{figure}

\end{document}